\documentclass[final,5p,times,twocolumn]{elsarticle}
\usepackage{times}
\usepackage{amssymb}
\usepackage{amsmath}
\usepackage{mathtools}
\usepackage[breakable]{tcolorbox}

\usepackage{hyperref}
\usepackage{url}
\usepackage{microtype}
\usepackage{graphicx}
\usepackage{subcaption}
\usepackage{booktabs}
\usepackage{siunitx}
\usepackage{cleveref}
\usepackage{textcomp}
\usepackage{multirow}
\usepackage{mdframed}
\usepackage{dblfloatfix}

\usepackage{tcolorbox}
  \tcbuselibrary{listings}
  \tcbset{
    promptbox/.style={
      colback=teal!5!white,
      colframe=teal!70!black,
      coltitle=white,
      fonttitle=\bfseries,
      boxrule=1pt,
      arc=2mm,
      left=4pt,right=4pt,top=4pt,bottom=4pt,
      sharp corners=south,
      width=\textwidth,
    }
  }
\usepackage[table]{xcolor} 

\usepackage{algorithm}
\usepackage{algpseudocode}
\usepackage{amsmath}

\begin{document}
\begin{frontmatter}

\title{In-Context Bias Propagation in LLM-Based Tabular Data Generation
}

\author[inst1]{Pol G. Recasens\corref{cor1}}
\author[inst1]{Alberto Gutierrez}
\author[inst1,inst2]{Jordi Torres}
\author[inst2,inst1]{Josep Ll. Berral} 
\author[inst3]{Javier Carnerero-Cano}
\author[inst3]{Anisa Halimi}
\author[inst3]{Kieran Fraser}

\cortext[cor1]{Corresponding author: \texttt{pol.garcia@bsc.es}}

\address[inst1]{Barcelona Supercomputing Center (BSC), Spain}
\address[inst2]{Universitat Politècnica de Catalunya (UPC), Spain}
\address[inst3]{IBM Research Ireland}

\begin{abstract}
Large Language Models (LLMs) are increasingly used for synthetic tabular data generation through in-context learning (ICL), offering a practical solution for data augmentation in data scarce scenarios. While prior work has shown the potential of LLMs to improve downstream task performance through augmenting underrepresented groups, these benefits often assume access to a subset of unbiased in-context examples, representative of the real dataset. In real-world settings, however, data is frequently noisy and demographically skewed. In this paper, we systematically study how statistical biases within in-context examples propagate to synthetic tabular data, showing that even mild in-context biases lead to global statistical distortions. We further introduce an adversarial scenario where a malicious contributor can inject bias into the synthetic dataset via a subset of in-context examples, ultimately compromising the fairness of downstream classifiers for a targeted and protected subgroup. Finally, we evaluate prompt-side mitigation strategies based on preprocessing in-context examples and instruction-based defenses, demonstrating that while such techniques can attenuate disparity, the inherent sensitivity of LLMs to adversarial prompts remains a persistent challenge. Our findings highlight a critical new vulnerability in LLM-based data generation pipelines within sensitive domains. 
\end{abstract}


\end{frontmatter}

\section{Introduction}

The scarcity of high-quality, labeled data remains a primary bottleneck in deploying machine learning systems in sensitive domains such as healthcare, finance, and criminal justice. Due to strict privacy regulations and the high cost of manual annotation, practitioners increasingly rely on synthetic data augmentation to build robust classifiers~\cite{borisov2022language}. While traditional generative models~\cite{xu2019modeling,zhao2021ctab} have been the standard for tabular synthesis, they often require substantial in-domain data to generate reliable data.

Recently, Large Language Models (LLMs) have emerged as a powerful alternative, demonstrating remarkable generalization capabilities through in-context learning (ICL). By conditioning on a small set of ICL examples, LLMs can generate realistic, statistically coherent tabular samples without the need for fine-tuning~\cite{borisov2022language}. This paradigm leverages the model's strong semantic priors to augment underrepresented groups, offering a resource-efficient solution for data-scarce regimes. Recent studies show that LLM-generated synthetic data can significantly boost downstream task performance, matching or even surpassing traditional data augmentation methods in data-scarce regimes~\cite{kim2024epic, seedat2023curated}.

However, the efficacy of ICL relies on a critical and often unexamined assumption, which is that in-context examples provided in the prompt are unbiased and independently drawn from the real data distribution. In real-world deployment scenarios \textendash such as Retrieval-Augmented Generation (RAG) pipelines or collaborative data platforms \textendash this assumption rarely holds. In-context examples are frequently noisy, skewed, or dynamically retrieved from unverified sources.

\begin{figure}[htbp]
  \centering
  \includegraphics[width=1.0\columnwidth]{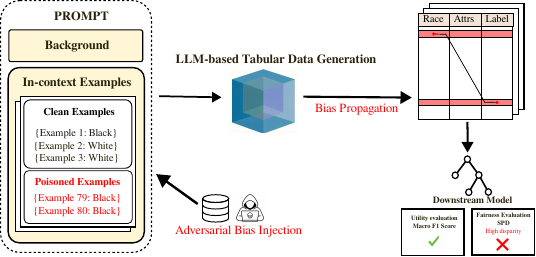}
    \caption{\textbf{End-to-end bias propagation.} An adversarial user injects biased examples into the prompt for a targeted group. The LLM replicates these statistical patterns via in-context learning, propagating the bias into the synthetic tabular data, and further compromising downstream fairness.}
  \label{fig:prplot}
\end{figure}

This paper addresses \textit{in-context bias propagation}, a critical vulnerability inherent to LLM-based synthetic data generation. Unlike traditional ``intrinsic'' bias, which stems from the model's pretraining weights~\cite{nadeem2021stereoset}, we investigate ``extrinsic'' bias introduced at inference time, noting that in practice, these two source of bias interact during generation. We argue that the same mechanism that allows LLMs to adapt rapidly to new tasks also makes them susceptible to reproducing demographic disparities present in the context examples. Consider a community hospital utilizing an LLM to augment a dataset where patient records disproportionately represent a specific demographic. As illustrated in Figure~\ref{fig:prplot}, such statistical imbalances can inadvertently condition the model to propagate these distortions into the synthetic data, potentially exacerbating disparities in downstream decision-making.

Furthermore, this exposes prompt-based adversarial bias injection as a new attack surface. While the tendency of machine learning models to amplify training data biases is well-documented~\citep{dinan2020queens, leino2018feature, hall2022systematic}, the security implications of in-context bias are underexplored. We introduce a threat model where a malicious actor injects a small number of carefully crafted examples into a collaborative prompt or retrieved context. We demonstrate that this targeted manipulation can skew the synthetic distribution and compromise the fairness of downstream classifiers, all while maintaining high utility metrics that mask the attack.

\textbf{Contributions.} In this paper, we study how in-context examples influence the statistical properties of LLM-generated tabular data. Our contributions are:

\begin{itemize}
\item We demonstrate that statistical biases present in ICL examples systematically propagate to the distribution of LLM-generated tabular data, and that LLMs capture and propagate biases in a conditional and intersectional manner. We quantify this propagation via linear sensitivity analysis across multiple model families.

\item We introduce a novel perspective on prompt injection, an adversarial setting where a malicious agent \textit{injects} a small number of biased in-context examples into a collaborative data generation pipeline. We show that this targeted manipulation induces fairness violations for the attacked group, demonstrating that an adversary controlling only a small fraction of the prompt can compromise downstream fairness without degrading utility metrics.

\item We systematically characterise the effectiveness of prompt-side defenses against in-context bias injection. We show that these strategies can reduce synthetic and downstream disparities, but no single defense consistently controls all fairness metrics without utility trade-offs, and that higher-order correlations injected by feature-aligned attacks persist even when marginal parity constraints are enforced.

\end{itemize}

The remainder of this paper is organized as follows. Section \ref{sec:related_work} reviews related work in tabular generation and fairness. Section \ref{sec:setup} details the setup of our adversarial study. Section \ref{sec:bias-model} formalizes and empirically analyses in-context bias propagation. Section \ref{sec:adv-bias} studies in-context bias propagation as an adversarial threat. Section \ref{sec:mitigation} studies prompt-side mitigation strategies, and Sections \ref{sec:discussion} and \ref{sec:conclusion} conclude the work.


\section{Related Work}
\label{sec:related_work}

\paragraph{Tabular data generation with LLMs}

Pretrained language models have emerged as powerful tools for synthesizing tabular data, leveraging broad semantic priors to augment datasets in limited or sensitive domains. These models offer distinct advantages where traditional methods, such as VAEs \cite{xu2019modeling, tazwar2024tab}, GANs \cite{tanaka2019data, xu2019modeling, zhao2021ctab}, Diffusion Models \cite{zhang2023mixed, kotelnikov2023tabddpm}, and SMOTE \cite{chawla2002smote}, struggle to generalize from unrepresentative data \cite{seedat2023curated}. While fine-tuning approaches like GReaT \cite{borisov2022language} and Harmonic \cite{wang2024harmonic} have demonstrated efficacy in structured data generation, prompting-based methods present a resource-efficient alternative with particular promise for augmenting underrepresented groups. For instance, CuratedLLM \cite{seedat2023curated} utilizes in-context learning (ICL) followed by manual curation to address data scarcity, while Epic \cite{kim2024epic} relies exclusively on in-context examples and categorical mappings to generate realistic examples. Such capabilities have already been translated to real-world healthcare scenarios to mitigate data scarcity \cite{tornqvist2024text}. However, existing prompting methods typically rely on the strong assumption that in-context examples are independent and identically distributed (i.i.d.) with respect to the target distribution. In practice, available data is frequently skewed or demographically imbalanced, and the mechanisms by which these biases propagate during LLM-based tabular generation remain poorly understood.

\paragraph{Fairness in language models}

LLMs reproduce demographic biases inherent in pretraining data, a phenomenon well-studied across benchmarks for both likelihood-based and generative settings \cite{nadeem2021stereoset, nangia2020crows, smith2022m, dhamala2021bold}. Mitigation strategies mostly rely on decoding and prompting interventions, including self-debiasing \cite{schick2021self}, fairness-guided prompt search \cite{ma2023fairness}, and targeted internal modifications of attention or FFN components \cite{zhou2024unibias}. In structured data settings, \cite{liu2023confronting} addresses demographic disparities via in-prompt label manipulations, while \cite{cherepanova2024improving} improves group fairness in tabular classification through curated exemplars and masking. Moreover, recent works \cite{kenfack2025towards,lu2026bias} investigated group fairness in LLM-based tabular classification through curated in-context examples and fairness-aware instructions. While the mentioned prior works directly use the LLM as a classifier, we investigate fairness in \emph{tabular data generation}, where LLMs are employed to synthetically augment the dataset in scarce scenarios before training a classification downstream model on the generated data if needed.

Beyond unintended bias, we also consider adversarial fairness. While classical fairness attacks typically induce group disparities via training-time data poisoning \cite{solans2020poisoning, van2022poisoning} or backdoors \cite{xue2024badfair}, we investigate \emph{prompt fairness attacks} in the generative setting. In our framework, an adversary biases the synthetic distribution not by corrupting the model, but by strategically manipulating in-context exemplars during inference.

\section{Experimental setup}
\label{sec:setup}

\subsection{Models} We evaluate four open-source LLMs with sizes ranging from 8 billion to 70 billion parameters, which are\textit{ Granite-8B}, \textit{Mistral-7B}, \textit{Mixtral-8x7B} and \textit{Qwen-70B}. All models are deployed using vLLM on a cluster of four NVIDIA H100 GPUs. For each experimental setting, we generate 5000 synthetic rows through multiple independent calls, each generating two samples.

\subsection{In-Context Examples} To preserve statistical independence and mitigate potential ordering biases, in-context examples are regenerated and randomized every 10 inference calls. 

For the propagation analysis in Section \ref{sec:bias-model}, the in-context examples are synthetically constructed. We modify selected variables in a fraction $\pi$ of the $k$ demonstrations, such as \textit{race} for marginal bias or \textit{race} and the target jointly for conditional bias. For the adversarial study in Section~\ref{sec:adv-bias}, the clean examples are sampled i.i.d.\ from the real dataset, while a fraction $\pi$ is adversarially manipulated. Therefore, the bias is introduced through a controlled intervention of the in-context examples and the prompt bias rate is fixed by design. For each prompt, we set
\begin{equation}
n_{\mathrm{biased}}=\operatorname{round}(\pi k),
\qquad
n_{\mathrm{clean}}=k-n_{\mathrm{biased}},
\end{equation}
and include exactly $n_{\mathrm{biased}}$ biased demonstrations and $n_{\mathrm{clean}}$ clean demonstrations. Moreover, as a manipulation check \cite{uddin2024novel}, Table~\ref{tab:prompt_bias_validation} reports the realized prompt-level statistical parity difference, $\mathrm{SPD}_P$, against a null distribution obtained from clean prompts sampled from the real training data. Across all datasets and nonzero bias rates, the injected prompts yield negative $\mathrm{SPD}_P$ values outside or near the boundary of the clean-prompt 95\% null intervals, with statistically significant permutation tests ($p_{\mathrm{perm}}<0.05$). 

\begin{table}[t]
\scriptsize
\setlength{\tabcolsep}{3pt}
\renewcommand{\arraystretch}{1.0}
\centering
\caption{Statistical validation of bias in the in-context examples ($k=80$). $\mathrm{SPD}_P$ denotes the prompt-level statistical parity difference. The null interval is obtained from $B=1000$ clean prompts of the same size sampled from the real training data. $p_{\mathrm{perm}}$ is the two-sided empirical permutation $p$-value.}
\label{tab:prompt_bias_validation}
\begin{tabular}{l c r c c}
\toprule
\textbf{Dataset} & $\boldsymbol{\pi}$ & $\boldsymbol{\mathrm{SPD}_{P}}$ & \textbf{Null 95\% CI} & $\boldsymbol{p_{\mathrm{perm}}}$ \\
\midrule
Adult & 0.3 & {-0.325} & [0.043, 0.369] & 0.002 \\
 & 0.6 & {-0.609} & [0.021, 0.349] & 0.002 \\
\addlinespace
Compas & 0.3 & {-0.395} & [-0.350, 0.112] & 0.024 \\
 & 0.6 & {-0.358} & [-0.352, 0.091] & 0.042 \\
\addlinespace
Diabetes & 0.3 & {-0.256} & [0.076, 0.513] & 0.002 \\
 & 0.6 & {-0.318} & [0.079, 0.484] & 0.002 \\
\addlinespace
Thyroid & 0.3 & {-0.288} & [-0.275, 0.178] & 0.038 \\
 & 0.6 & {-0.517} & [-0.287, 0.180] & 0.002 \\
\bottomrule
\end{tabular}
\end{table}

\subsection{Prompts} We adopt the prompt structure from CuratedLLM~\cite{seedat2023curated}, which decomposes the input into \textit{role}, \textit{task}, and \textit{in-context example} components. We serialize the examples in JSON format and vary the context size $k \in \{20, 40, 60, 80, 100\}$ to simulate data-scarce regimes. The exact prompts are provided in \ref{app:c}.


\subsection{Datasets}

We use the \emph{Adult} \cite{asuncion2007uci} and \emph{Compas} \cite{angwin2016machine}, \emph{Diabetes}, and \emph{Thyroid} datasets \cite{asuncion2007uci}, widely used in fairness and healthcare-related research. We additionally use Bank Marketing \cite{moro2014bank} as a larger mixed-type real-world dataset, with results reported in \ref{app:a}. \emph{Compas} contains recidivism risk assessments, \emph{Adult} includes census income data, \emph{Diabetes} is used for predicting the onset of diabetes based on diagnostic measurements, and \emph{Thyroid} assesses the risk of thyroid disorders based on various clinical and demographic features. We detail the datasets in \ref{app:datasets}.

\subsection{Evaluation Metrics}
We evaluate the generated synthetic data across three dimensions: downstream utility, distributional fidelity, and group fairness with respect to a protected attribute. To quantify uncertainty in fairness metrics, we partition the synthetic dataset into 5 blocks and report the mean and standard deviation across them. For downstream fairness (e.g., $\text{SPD}_D$), we train multiple independent models using different seeds and data splits, reporting the mean and standard deviation evaluated on the real test set.

\subsubsection{Quality} To assess \textit{utility}, we report macro F1 score of each classifier evaluated on the real test set. We use Random Forest, Logistic Regression and Gradient Boosting as classical classifiers, and TabPFN \cite{TabPFN-2.5} as a LLM-based classifier for tabular data.

\subsubsection{Fidelity} 
Second, to measure marginal \textit{fidelity}, we use distinct metrics for categorical and numerical features. For each categorical variable $x$ with support $\mathcal{V}$, we compute the Total Variation Complement ($1 - \text{TVD}$), where TVD is defined as:
\begin{equation}
  \mathrm{TVD}(\mathcal{D}_R,\mathcal{D}_G) = \frac{1}{2} \sum_{v \in \mathcal{V}} \bigl|\mathcal{D}_R(x=v) - \mathcal{D}_G(x=v)\bigr|.
\end{equation}
For numerical variables, we quantify similarity using the Jensen–Shannon divergence (JSD), a smoothed, symmetric version of the Kullback-Leibler divergence:

\begin{equation}
\mathrm{JSD}(\mathcal{D}_R \,\|\, \mathcal{D}_G) = \frac{1}{2}\,\mathrm{KL}\bigl(\mathcal{D}_R \,\big\|\, M\bigr) + \frac{1}{2}\,\mathrm{KL}\bigl(\mathcal{D}_G \,\big\|\, M\bigr),
\end{equation}

where $M = \frac{1}{2}(\mathcal{D}_R + \mathcal{D}_G)$. 

To evaluate whether synthetic data preserves dependencies among variables, we compute an association matrix for both real and synthetic datasets and compare them using Association Matrix Distance. Let $A_R,A_G \in \mathbb{R}^{d \times d}$ denote the association matrices computed on the real and generated datasets, respectively. Each entry $A_{ij}$ is computed using Spearman correlation for numerical--numerical pairs, correlation ratio for numerical--categorical pairs, and Cramer's $V$ for categorical--categorical pairs. We define Association Matrix Distance as

\begin{equation}
\mathrm{AMD}(\mathcal{D}_R,\mathcal{D}_G)
=
\frac{2}{d(d-1)}
\sum_{1 \leq i < j \leq d}
\left|
A_R(i,j) - A_G(i,j)
\right|.
\end{equation}

We further report the radial-basis-function maximum mean discrepancy (RBF-MMD) \cite{gretton2012kernel}, which measures the global discrepancy between real and synthetic samples in a kernel-induced feature space.

\subsubsection{Fairness} 
Finally, to quantify \textit{fairness}, we measure the Statistical Parity Difference (SPD), which captures the discrepancy in favorable outcomes (y=1) between an unprivileged subgroup $a_U$ and a privileged subgroup $a_P$, for a protected attribute $a$:
\begin{equation}
    \text{SPD} = P_{D_G}(y=1|a=a_U) - P_{D_G}(y=1|a=a_P),
\end{equation}
where for multi-valued protected attributes, we define $a_P$ as the complement of $a_U$. A value closer to 0 indicates a fairer synthetic distribution that preserves demographic parity. We study both $\text{SPD}_S$ of the synthetic data, and $\text{SPD}_D$ of the downstream model evaluated on real data. For the downstream model, we also measure Equalized Odds (EOD), defined as the difference between the True Positive Rate (TPR) and the False Positive Rate (FPR) of downstream predictions across subgroups:
\begin{equation}
\text{EOD} = \frac{1}{2}\left( |TPR_{a_U} - TPR_{a_P}| + |FPR_{a_U} - FPR_{a_P}| \right),
\end{equation}
and Equal Opportunity (EO), defined as the absolute difference in TPR between subgroups:
\begin{equation}
\text{EO} = |TPR_{a_U} - TPR_{a_P}|,
\end{equation}
where $TPR_a = P(\hat{y}=1 | y=1, a)$ denotes the True Positive Rate for the downstream model predictions $\hat{y}=1$, and $FPR_a = P(\hat{y}=1 | y=0, a)$ denotes the False Positive Rate for a subgroup $a$ and unfavorable outcome $y=0$.

We note that marginal fairness metrics, such as SPD, are sensitive to underlying base rates and distribution shifts. In this regard, our experimental design contextualizes SPD in two ways. First, to mitigate base-rate confounding in our conditional and intersectional propagation analyses in Section \ref{sec:bias-model}, we explicitly control the marginal distribution using a "balanced prompt". Second, in our adversarial evaluation in Section \ref{sec:adv-bias}, we measure the deviation in SPD relative to the $\pi=0$ unbiased baseline, ensuring we isolate the impact of the injected context from any baseline class imbalances. This approach fixes the base rates by instructing the model to generate uniform subgroup counts, ensuring that variations in SPD accurately isolate the propagation of targeted conditional dependencies.

\section{In-Context Bias Propagation in LLM-generated Data}
\label{sec:bias-model}


In this section, we study how statistical patterns in the in-context examples propagate to the synthetic data generated by the LLM. We compare the generated distribution with a reference distribution derived from the model's task-specific prior, and then measure how controlled changes in the in-context examples propagate to the generated output. The bias intensity $\pi$ controls the statistical distortion introduced into the demonstrations, while the resulting shift is quantified through a relevant statistic such as a protected-group or label probability. 

Given a supervised classification task, a downstream model learns a function 
$f \in \mathcal{F}$, $f : \mathcal{X} \rightarrow \mathcal{Y}$, 
where $\mathcal{X}$ is the feature space, and $\mathcal{Y}$ is the label space. 
We assume an underlying distribution 
$\mathcal{D}$ over triples $(x,a,y) \in \mathcal{X} \times \mathcal{A} \times \mathcal{Y}$, where $a$ is the protected attribute, and an observed labeled dataset $\text{S}_{\mathrm{train}} = \{(x_i,a_i,y_i)\}_{i=1}^n \;\;\text{with}\;\; (x_i,a_i,y_i) \stackrel{\text{i.i.d.}}{\sim} \mathcal{D}$. In data scarce scenarios, where $n < 100$, the prior knowledge of pretrained language models has been shown to be a resource efficient alternative for synthetic data augmentation \citep{seedat2023curated}, improving utility of the downstream classifier.  

Let $M$ be a pretrained language model. The LLM-based tabular data augmentation process is guided by a prompt $P$ containing $k$ serialized demonstration examples $\{x_1, \dots, x_k\}$ drawn i.i.d from $S_{\mathrm{train}}$. We define $\mathcal{D}_G$ as the generated distribution, i.e., the distribution of synthetic examples produced by $M$ when queried with $k$ in-context examples. We also define $\tilde{\mathcal{D}_0}$ as the anchor prior distribution, obtained as the distribution of examples generated by $M$ using the same prompt template but with $k=0$ in-context examples. This distribution captures the model’s internal prior for the given task.

To quantify bias, we consider a bounded statistic $\phi: \mathcal{X} \times \mathcal{A} \times \mathcal{Y} \rightarrow \mathbb{R}$. For a probability distribution $\mathcal{D}$, the associated scalar bias is $\mu_\phi(\mathcal{D}) = \mathbb{E}_{z \sim \mathcal{D}}[\phi(z)]$. Our goal is to characterize how the bias intensity $\pi$ relates to the bias shift $|\mu_\phi(\mathcal{D}_G) - \mu_\phi(\mathcal{D}_0)|$ observed between the reference distribution $\mathcal{D}_0$ and the ICL-generated distribution $\mathcal{D}_G$.

\subsection{Modeling In-Context Sensitivity}

Recent works interpret in-context learning as approximate Bayesian inference \citep{xie2021explanation, panwar2023context, pmlr-v235-lin24l}, where pretraining induces a prior over latent tasks, and the in-context examples provide observations that shift the model's predictive distribution. In this view, the model's output can be understood as combining a task prior induced by pretraining with evidence provided in the prompt. Motivated by this perspective, we introduce the following decomposition as a conceptual framework for studying the influence of in-context examples on tabular generation
\begin{equation}
\label{eq:mix}
\mathcal{D}_G \approx (1-\alpha_k)\,\tilde{\mathcal{D}_0} \;+\; \alpha_k\,\Phi_M(\mathcal{D}_P),
\end{equation}
where $\tilde{\mathcal{D}_0}$ is a zero-shot anchor distribution induced by pretraining (same template, $k=0$). $\Phi_M : \mathcal{P}(\mathcal{X}) \rightarrow \mathcal{P}(\mathcal{X})$ denotes the model-dependent transformation from an empirical prompt distribution to the corresponding $k$-shot generation distribution, and $\alpha_k \in [0,1]$ represents the overall strength of prompt conditioning at context size $k$.

For $k>0$, prompting artifacts such as serialization, context length, and formatting can induce distribution shifts even when demonstrations are unbiased. This is closely related to contextual calibration, where Zhao et.al \citep{pmlr-v139-zhao21c} estimate prompt-induced biases by querying the model with content-free inputs under the same prompt structure and using the resulting output distribution as a calibration baseline. In this regard, we define the reference distribution $\mathcal{D}_{0} := \Phi_M(\tilde{\mathcal{D}}_0)$, obtained by prompting the model with $k$ examples drawn from the zero-shot anchor $\tilde{\mathcal{D}}_0$. This reference distribution controls for these effects, and provides an empirical anchor for the LLM's task-specific prior before introducing biased empirical evidence.

We quantify these deviations using a drift score $D_f(\cdot,\mathcal{D}_{0})$. Our goal is not to estimate the mixture coefficient $\alpha_k$ directly, but to measure how sensitively the generated distribution responds to controlled changes in the prompt distribution. For this reason, we use the OLS slope $\beta_k$ in Figure~\ref{fig:marginal_univariate_beta} to empirically approximate prompt-conditioning sensitivity, where larger $\beta_k$ indicates that changes in $\mathcal{D}_P$ are more strongly reflected in $\mathcal{D}_G$.

Finally, we connect distributional drift directly to our fairness objective. Since fairness metrics are derived from the underlying data distribution, the score $D_f(\mathcal{D}_G, \mathcal{D}_0)$ effectively drives the induced bias shift $|\mu_\phi(\mathcal{D}_G) - \mu_\phi(\mathcal{D}_0)|$. In the following section, we therefore focus on characterizing how demographic skews in the in-context examples affect statistical parity in both the synthetic output and downstream classifiers.

\subsection{Experimental Results}

Here, we systematically quantify the sensitivity of LLM-based tabular generation to statistical distortions within in-context examples. We structure our analysis across three levels of granularity: (i) marginal bias propagation, examining univariate shifts in protected attributes; (ii) conditional bias propagation, focusing on induced correlations between protected groups and target labels; and (iii) intersectional propagation, which evaluates bias propagation across joint subgroups. By varying the bias intensity ($\pi$) and number of in-context examples ($k$) across multiple model families, we characterize the dynamics of this bias propagation, establishing the empirical foundation for the adversarial vulnerabilities discussed in Section \ref{sec:adv-bias}.

\subsubsection{Experimental Setup}
\label{sec:design}

For the following study we use two prompt structures. First, we use the \emph{unconstrained} prompt for marginal bias propagation analysis, which does not include any extra instructions. Second, we use the \emph{balanced} prompt for our conditional and intersectional bias propagation analysis. This prompt explicitly instructs the LLM to generate one example per subgroup with the goal of keeping frequency counts approximately balanced in $D_G$, thus isolating conditional bias from subgroup frequency imbalances. Both prompts incorporate $k$ in-context demonstrations serialized in JSON format and drawn from a biased distribution $\mathcal{D}_P$. We measure marginal propagation strength using a drift score $D_f$, defined as the sum of the mean TVD for categorical variables and the mean JSD for numerical variables. In-context examples are drawn from $\mathcal{D}_0$, and for each analysis, we modify the probability of the studied variable.

\subsubsection{Marginal Bias Propagation}
\label{sec:univariate_prop}

\begin{figure}[h]
  \centering
  \includegraphics[width=1.0\columnwidth]{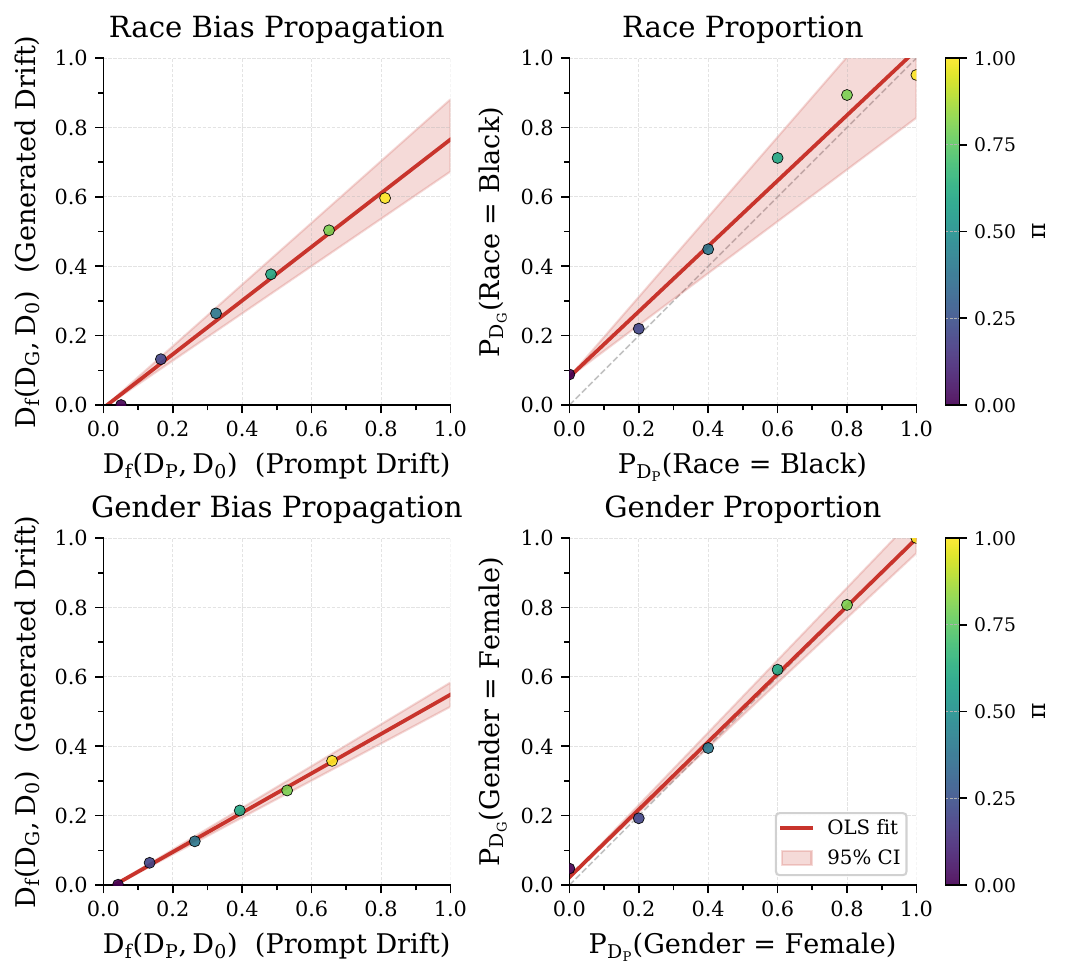}
    \caption{\textbf{Marginal in-context bias propagation} (\textit{Mixtral-8x7B}, $k=80$). (\textit{Left}) Drift in the generated distribution $D_f(\mathcal{D}_G, \mathcal{D}_0)$ versus drift in the prompt distribution $D_f(\mathcal{D}_P, \mathcal{D}_0)$. The linear relationship shows that distributional shifts in prompts are proportionally transferred to synthetic outputs. (\textit{Right}) Probability of the target group (Race/Gender) in generated samples increases linearly with prompt bias intensity $\pi$, confirming that LLMs propagate univariate demographic skews from in-context examples.}
  \label{fig:univariate_drift}
\end{figure}

We begin by quantifying the sensitivity of LLM-generated distributions ($\mathcal{D}_G$) to univariate statistical distortions within the $k \in \{20,40,60,80\}$ in-context examples drawn from $\mathcal{D}_P$. We focus specifically on the propagation of imbalances in protected attributes, such as race or gender. In this regard, Figure~\ref{fig:univariate_drift} presents the analysis for \textit{Mixtral-8x7B} under varying levels of racial bias intensity $\pi$, defined as the fraction of in-context examples assigned to a specific target group (e.g., \textit{Black}), while remaining attributes are sampled from the anchor distribution $\mathcal{D}_0$.

In Figure~\ref{fig:univariate_drift} (left), we compare the divergence of the prompt distribution, $D_f(\mathcal{D}_P, \mathcal{D}_0)$, against the resulting divergence in the generated distribution, $D_f(\mathcal{D}_G, \mathcal{D}_0)$. Given the no-ICL (zero-shot) model's prior distribution $\tilde{\mathcal{D}_0}$, we construct $\mathcal{D}_0 = \Phi_M(\tilde{\mathcal{D}_0})$ as the reference distribution. We observe a linear relationship between the input and output distributional shifts, with increases in $\pi$ producing proportional increases in the divergence of $\mathcal{D}_G$. Moreover, Figure~\ref{fig:univariate_drift} (right) demonstrates that the resulting probability of the target racial group in the synthetic output scales linearly with the underlying probability in the prompt.

Figure~\ref{fig:univariate_drift} (bottom) replicates the analysis for gender bias, with target group Female. While the linear sensitivity persists, the distributional drift for gender is attenuated compared to race. This discrepancy suggests a non-trivial interaction between the model's intrinsic priors and the in-context evidence with respect to $D_0$. However, the target gender probability retains a strictly linear relationship with $\pi$, confirming that the propagation mechanism remains consistent even when the distributional shift with respect to $\mathcal{D}_0$ varies.

The remainder of this section systematically characterizes this propagation behavior across three dimensions: prompt composition, context size, and model architecture. Our findings reveal that even mild perturbations in $\mathcal{D}_P$ can systematically skew the statistical parity of synthetic data, introducing significant fairness risks in tabular generation pipelines.

\paragraph{Number of in-context examples} We argue that the context size $k$ directly modulates the bias propagation strength, modeled as $\alpha_k$ in Equation \ref{eq:mix}. As $k$ increases, the synthetic distribution $\mathcal{D}_G$ diverges from the model's intrinsic zero-shot prior and increasingly aligns with the empirical prompt distribution $\mathcal{D}_P$.

\begin{figure}[t]
  \centering
  \includegraphics[width=1.0\columnwidth]{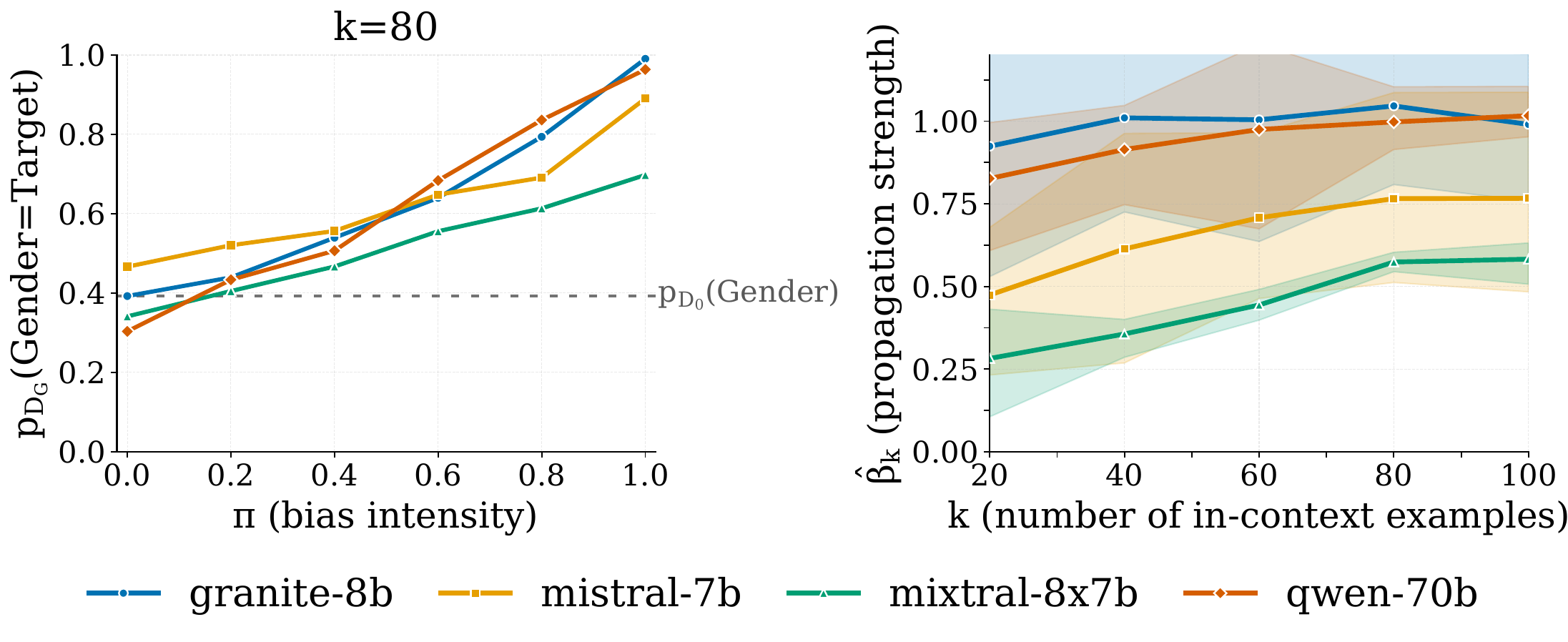}
    \caption{\textbf{Marginal bias propagation across model families.} (\textit{Left}) Probability of the target group in generated samples increases linearly with prompt bias intensity $\pi$ for all models at $k=80$. The dashed line shows the anchor baseline $p_{\mathcal{D}_0}(\text{Target})$. (\textit{Right}) Propagation coefficient $\beta_k$ increases monotonically with context size $k$ across all models.}
  \label{fig:marginal_univariate_beta}
\end{figure}

In Figure \ref{fig:marginal_univariate_beta} (left), we validate this effect for a fixed $k=80$ across all four model families. We observe that the probability of the synthetic samples belonging to the targeted Black subgroup scales approximately linearly with the bias intensity $\pi$ injected into the in-context examples. Specifically, at $\pi = 0$, the output distribution remains close to the anchor baseline ($p_{\mathcal{D}_0}(\text{Race}=\text{Black})$, but as $\pi$ increases, the LLMs exhibit consistent sensitivity to the univariate racial skews present in the prompt. Notably, different models exhibit varying degrees of sensitivity to the same prompt biases, with some models showing stronger alignment with the prompt distribution than others.

To quantify this sensitivity across different context sizes, we compute the propagation coefficient $\beta_k$ by fitting an ordinary least squares (OLS) regression between the drift of the prompt distribution from the anchor, $D_f(\mathcal{D}_P, \mathcal{D}_0)$, and the drift of the generated distribution from the anchor, $D_f(\mathcal{D}_G, \mathcal{D}_0)$. Figure \ref{fig:marginal_univariate_beta} (right) demonstrates that $\beta_k$ increases monotonically with $k$ across all models, indicating that larger context windows increase the strength of the propagation of distributional shifts. While both $\beta_k$ and the absolute magnitude of distributional shifts vary significantly between architectures (see \ref{app:a}), the fundamental linear relationship between the probability of synthetic examples from a targeted group and their proportion in the in-context examples remains consistent across all model families.

\begin{figure*}[t]
  \centering
  \includegraphics[width=1.0\linewidth]{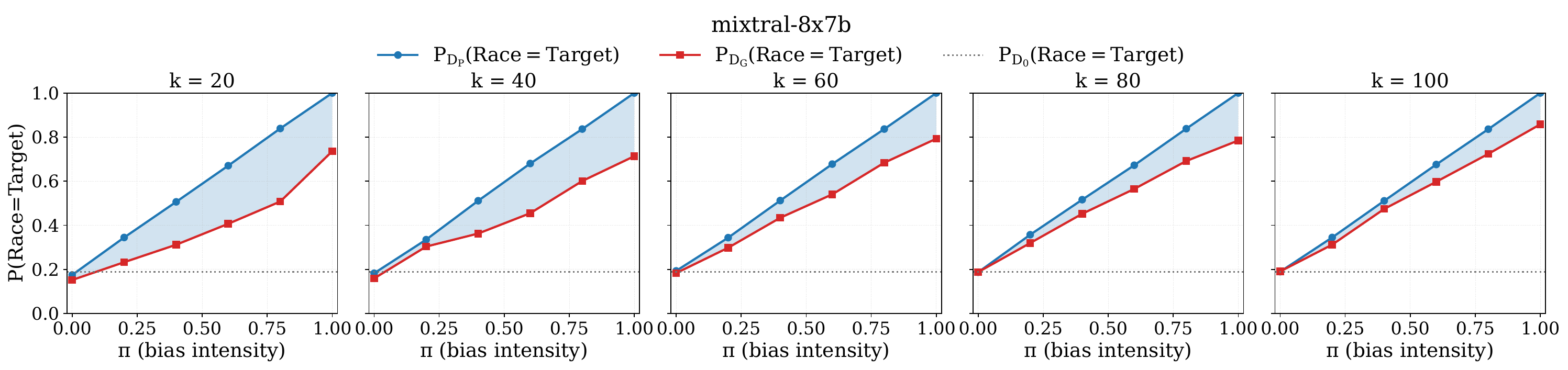}
    \caption{\textbf{Bias propagation strength increases with context size.} Each subfigure shows results for a different context size $k \in \{20, 40, 60, 80\}$. The red curve shows the target group probability in generated samples $p_{\mathcal{D}_G}(\text{Target})$, the blue curve shows the target group probability in prompt examples $p_{\mathcal{D}_P}(\text{Target})$, and the dashed line shows anchor baseline $p_{\mathcal{D}_0}(\text{Target})$. As $k$ increases, the red curve converges toward the blue curve, indicating that larger context windows lead to stronger replication of prompt-level demographic distributions.}
    \label{fig:univariate_bias_per_k}
\end{figure*}

These findings are consistent with the mechanism underlying prior observations that larger context windows improve synthetic data fidelity \cite{kim2024epic, cherepanova2024improving}. When $\mathcal{D}_P$ represents the ground truth ($\mathcal{D}_P \equiv \mathcal{D}_R$), a large $k$ ensures high utility by aligning the output with real-world statistics. However, our results in Figures \ref{fig:marginal_univariate_beta} and \ref{fig:univariate_bias_per_k} demonstrate that this same mechanism creates a critical vulnerability, as when $\mathcal{D}_P$ is demographically skewed, larger $k$ values more closely reproduce these distortions into the synthetic dataset.

\subsubsection{Conditional Bias Propagation}

While the previous univariate propagation analysis demonstrates that LLMs faithfully replicate marginal distributions within in-context examples, real-world biases often manifest through complex correlations between protected attributes and outcome variables. To investigate this, we now examine how LLMs learn and reproduce statistical correlations between a protected subgroup (e.g., a specific gender) and a target label (e.g., high income) from biased in-context examples.

We induce targeted conditional bias by constructing prompts where examples from a protected subgroup $S_{\text{target}}$ (e.g., Female) are systematically correlated with a positive outcome $Y=1$ (e.g., High Income). To disentangle conditional dependence from frequency effects, we explicitly instruct the LLM to generate a balanced number of examples across subgroups with the \textit{balanced prompt} defined in Section \ref{sec:design}. We vary the bias intensity $\pi$, defined as the positive label probability for the target group $S_{\text{target}}$ in-context examples, while maintaining a constant positive label probability of 0.5 for the non-targeted subgroup. 

\begin{figure}[t]
  \centering
  \includegraphics[width=1.0\linewidth]{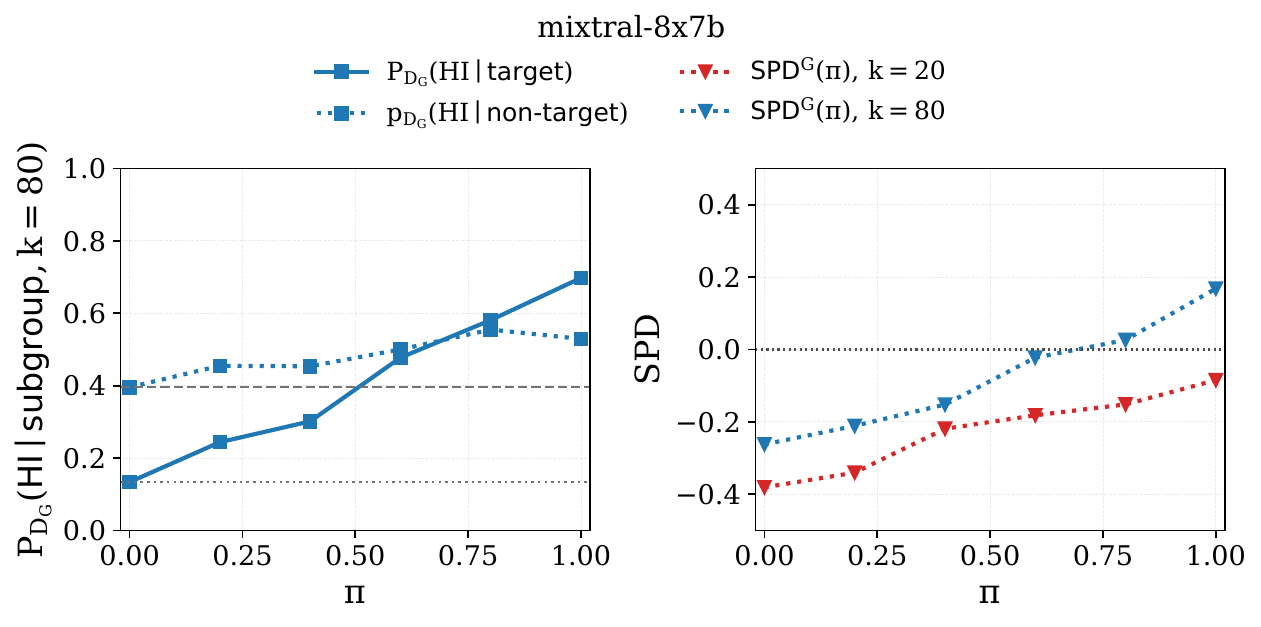}
  \vspace{-0.8em}
  \caption{\textbf{Conditional bias propagation for targeted and non-targeted subgroups.} (\textit{Left}) Conditional probability of generating High Income examples for each gender subgroup as the bias intensity $\pi$ increases. The targeted subgroup exhibits strong linear sensitivity to $\pi$, while the non-targeted subgroup remains near 0.5. (\textit{Right}) Statistical parity difference (SPD) between subgroups increases more steeply at $k=80$ than $k=20$, demonstrating that larger context windows increase sensitivity to conditional bias propagation.}
  \label{fig:icl_samples}
\end{figure}

As shown in Figure~\ref{fig:icl_samples}, Mixtral-8x7B effectively captures the conditional probability $P(Y=1 \mid S_{\text{target}})$ from the in-context examples and propagates it to synthetic examples. The left figure demonstrates that as $\pi$ increases, the probability of generating high-income examples for the targeted Female subgroup increases proportionally, while the non-targeted Male subgroup remains stable near 0.5. Furthermore, comparing curves at $k=20$ versus $k=80$ reveals that the generated conditional distribution aligns more closely with prompt statistics at larger context sizes. This increased fidelity is quantified in Figure~\ref{fig:icl_samples} (right) through the statistical parity difference (SPD) between subgroups, which shows a steeper increase with $\pi$ at $k=80$ compared to $k=20$. These findings illustrate how conditional bias propagation affects fairness metrics even when univariate marginals are balanced, and demonstrate that larger in-context windows increase the model's sensitivity to conditional correlations in the prompt.

\begin{figure}[t]
  \centering
  \includegraphics[width=0.9\linewidth]{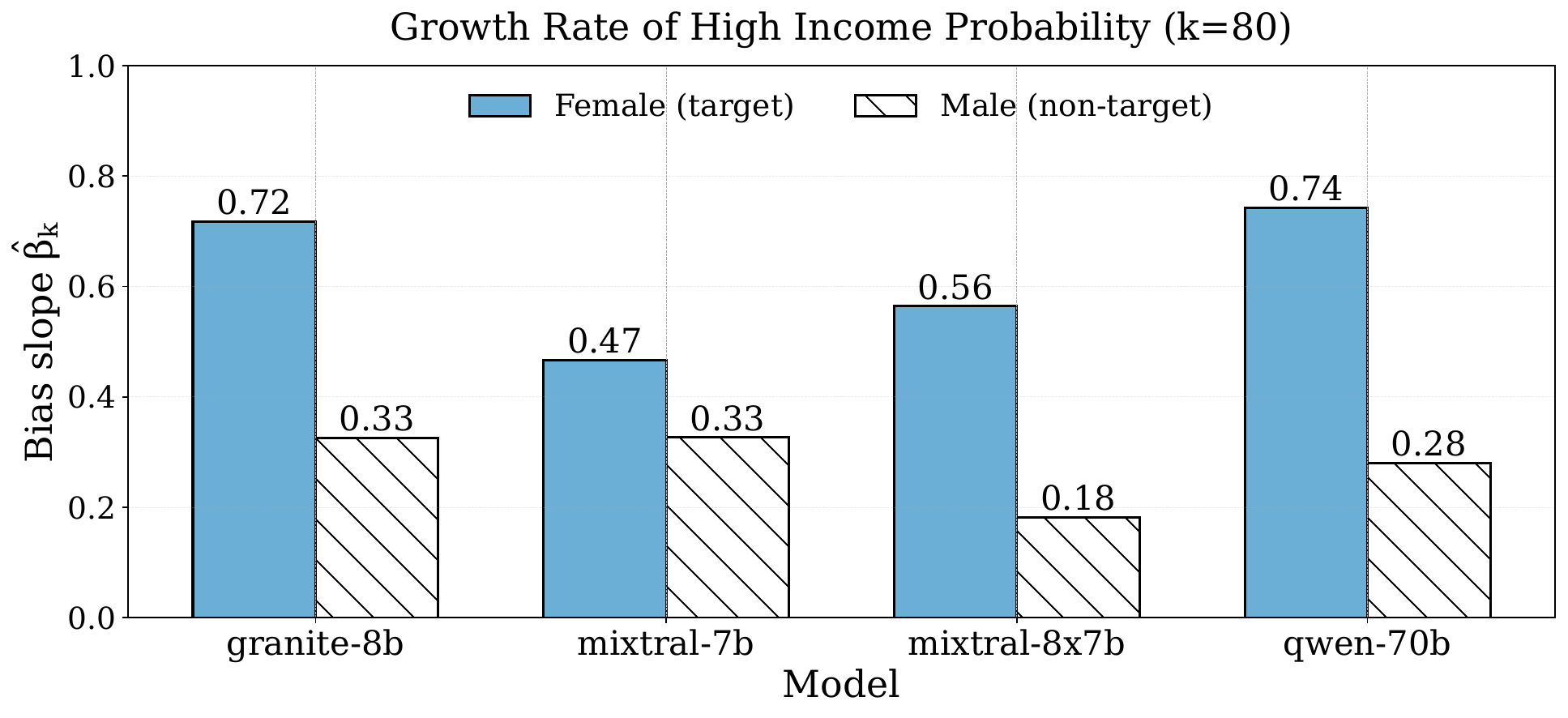}
  \vspace{-0.8em}
  \caption{\textbf{Model-specific sensitivity to conditional bias propagation.} (\textit{k=80, Adult}) The slope $\beta_k$ quantifies how strongly the predicted positive rate for each gender subgroup changes with bias intensity $\pi$. We estimate $\beta_k$ by fitting a linear regression to $P(Y=1 \mid \text{subgroup})$ across values of $\pi$. All models show substantially steeper slopes for the targeted (\textit{Female}) subgroup compared to the non-targeted (\textit{Male}) subgroup, indicating systematic propagation of the injected correlation. The magnitude varies across architectures, with larger and mixture models exhibiting stronger conditional bias transfer.}

  \label{fig:steepness_bars}
\end{figure}

\paragraph{Model-specific behavior} Figure~\ref{fig:steepness_bars} summarizes model-specific sensitivity to injected conditional bias by reporting the slope of the linear fit between bias intensity $\pi$ and the predicted positive rate for each gender subgroup. Across all models, the targeted subgroup (\textit{Female}) exhibits substantially steeper slopes than the non-targeted subgroup (\textit{Male}), confirming that the injected correlation is consistently learned in a group-specific manner. The non-targeted subgroup shows minimal growth, indicating limited spillover of the injected bias. Within the same family (Mistral-7B vs.\ Mixtral-8x7B), the mixture-of-experts model displays stronger conditional propagation, consistent with the hypothesis that increased model capacity enables better capture of complex conditional dependencies within the prompt. Similarly, Qwen-70B achieves the highest propagation slopes among all evaluated models, suggesting that both model scale and architecture influence the strength of conditional bias transfer.

\subsubsection{Intersectional Bias Propagation}

Next, we study whether LLMs propagate intersectional patterns that emerge only at the joint level of multiple protected attributes. We focus on the Adult dataset and consider the four intersectional groups defined by gender $\in$ \{Male, Female\} and race $\in$ \{Black, White\}. For a fixed context size $k$, we construct an \emph{intersectional} prompt ensuring balanced frequency counts across all four groups, such that the marginal distributions over gender and race remain approximately uniform. We configure the LLM to generate one example from each subgroup per call using the \textit{balanced prompt} defined in Section~\ref{sec:design}. We then manipulate the high-income target probability using $\pi \in [0,1]$, simultaneously increasing the positive rate for \{Female, Black\} and \{Male, White\} while decreasing it for \{Female, White\} and \{Male, Black\}. This design creates a bias pattern that cannot be detected by examining gender or race in isolation compared to the original biases in $\pi=0.0$. 

\begin{figure}[t]
  \centering
  \includegraphics[width=1.0\linewidth]{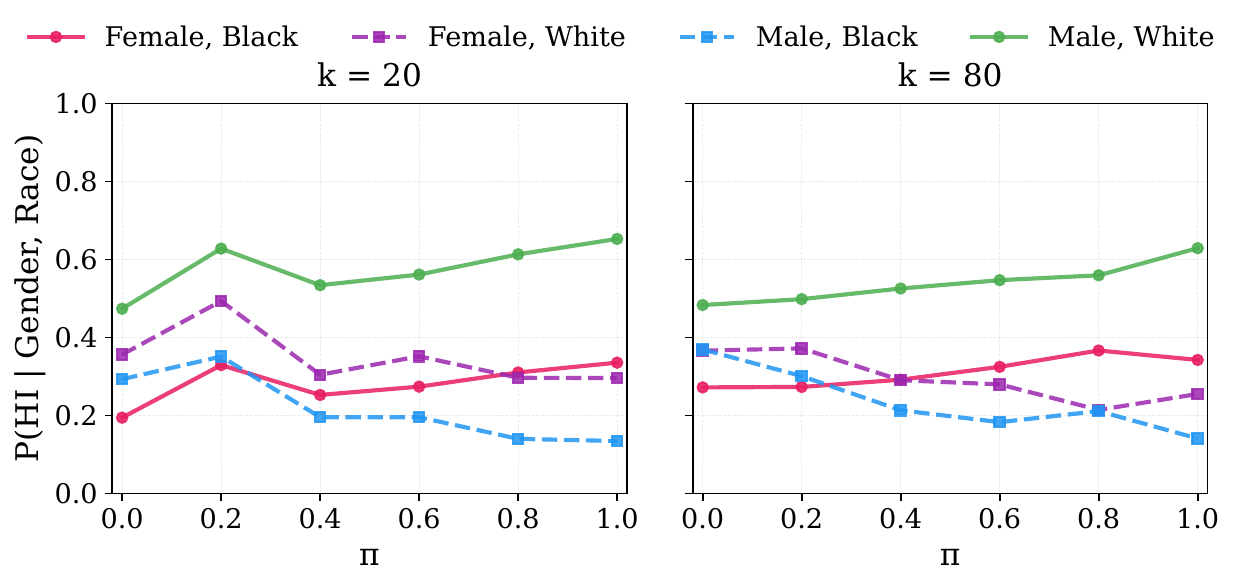}
  \vspace{-0.8em}
  \caption{\textbf{Multivariate intersectional probabilities.} (\textit{Granite-8b})
  Conditional probability of high income for each gender--race cell in the generated
  data as a function of the intersectional bias strength $\pi$, for different
  context sizes $k$. As $\pi$ and $k$ increase, the LLM increasingly reproduces the intersectional pattern injected in the prompt.}
  \label{fig:cell_probs}
\end{figure}

Figure~\ref{fig:cell_probs} reports the positive label probabilities for Granite-8b on Adult across different context sizes. Each subfigure corresponds to a different value of $k$, and each curve tracks one intersectional group as $\pi$ increases. The LLM reliably reproduces the injected pattern: \{Female, Black\} and \{Male, White\} exhibit monotonic increases in $P(Y{=}1)$ with $\pi$, while \{Female, White\} and \{Male, Black\} either plateau or decrease. The separation between targeted and non-targeted subgroups widens at higher $k$, indicating that intersectional propagation strength grows with the number of in-context examples.

\begin{figure}[t]
  \centering
  \includegraphics[width=1.0\linewidth]{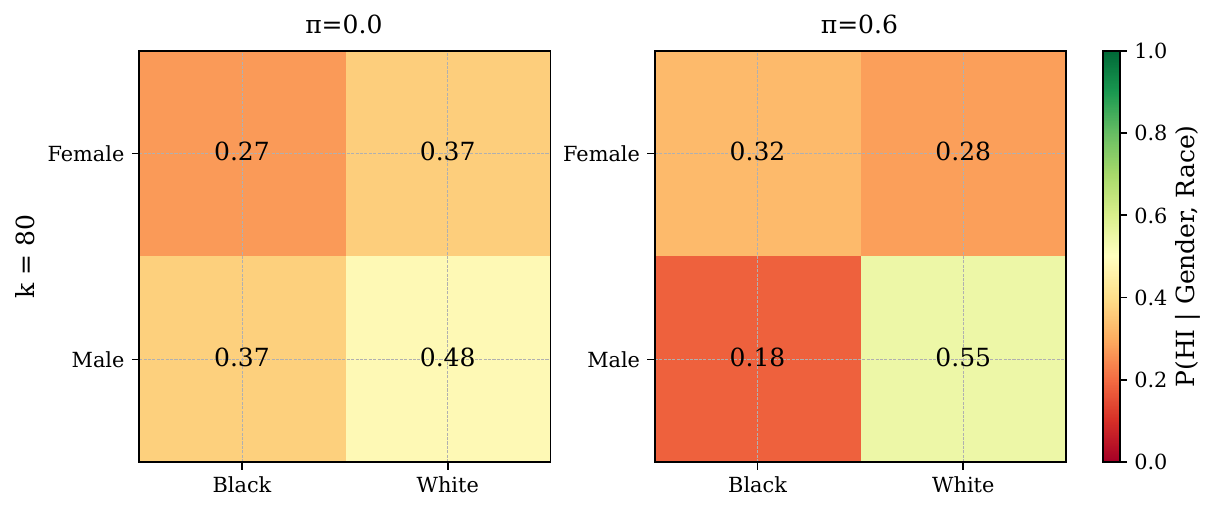}
  \vspace{-0.8em}
  \caption{\textbf{Intersectional high–income probability heatmap.} (\textit{Granite-8b})
    Conditional probability of high income $P(\text{HI}\mid \text{gender},\text{race})$
    in the generated data for Adult under the intersectional bias. The left figure
    corresponds to an unbiased prompt ($\pi=0$), while the right figure shows a fully biased prompt ($\pi=0.6$) in which (Female, Black) and (Male, White) are high-income cells. The model reproduces this injected intersectional pattern, raising $P(\text{HI})$ for the high income subgroups and lowering it for the other subgroups.}
      \label{fig:cell_heatmap}
\end{figure}

Figure~\ref{fig:cell_heatmap} provides a complementary visualization through heatmaps of high-income probabilities across the four intersectional cells. At $\pi=0$ (left), all four cells exhibit similar probabilities around 0.3-0.5, reflecting the model's intrinsic biases. At $\pi=0.6$ (right), probability mass moves to the targeted \{Female, Black\} and \{Male, White\} cells (dark red), while non-targeted subgroups show lower probabilities. This bias pattern induced by the intersectional bias now influences the baseline marginal disparities present at $\pi=0$. While LLMs carry inherent group-level biases from pretraining, they simultaneously reproduce the injected intersectional correlations, creating compound discrimination patterns where both sources of bias interact. 

\section{Adversarial In-Context Bias Injection}
\label{sec:adv-bias}

In the previous section, we analyzed in-context bias propagation as a potential \emph{risk} that might appear in LLM-based synthetic tabular data generation, arising from imbalances in the available data. In this section, we adopt a security perspective and investigate whether a malicious user can craft a small number of in-context examples that deliberately manipulate the statistical properties of the synthetic data for a targeted subgroup, without altering downstream model performance.

\subsection{Threat Model}

We consider a collaborative data generation pipeline in which multiple users independently contribute few-shot in-context examples to a shared LLM service. This scenario can also be viewed as populating a database, which is sampled to generate in-context examples through a RAG-service. Here, a malicious user could further leverage RAG vulnerabilities \cite{xue2024badrag} to influence the selection of examples that are adversarially injected. These examples are added into the prompt, which is then forwarded to the LLM to generate synthetic tabular data. An adversary is able to control a fraction $\pi$ of the $k$ examples in the prompt, but does not have access to the model parameters or to the in-context examples provided by honest users. We assume that the attacker knows (i) the downstream prediction task and the target label and (ii) general domain knowledge regarding the dataset attributes. The adversary’s objective is to influence the label rate of a targeted group in the synthetic data (e.g., the proportion of high-income labels assigned to Black individuals in the Adult dataset), while preserving high overall utility.

We note that our bias propagation analysis in Section \ref{sec:bias-model} implicitly studies a weaker attacker, since the marginal analysis requires knowledge only of the protected attribute and the conditional analysis additionally requires knowledge of the target label. The remaining attributes are selected from $\tilde{D}_0$, which can be obtained by querying the LLM with no in-context examples. The adversarial setting in this section extends this to the strongest case, where the attacker additionally leverages domain knowledge to construct feature-aligned examples to avoid naive detection. We position this as a worst-case analysis to establish the LLM-based tabular data generation vulnerability, and leave partially informed attackers a direction for future work.

\begin{table}[h!]
\caption{Attributes used to construct adversarial feature-aligned examples for each dataset.}
\label{tab:feature_alignment}
\large
\centering
\resizebox{\columnwidth}{!}{%
\begin{tabular}{l p{4.5cm} l p{6cm}}
\toprule
\textbf{Dataset} & \textbf{Protected Attr} & \textbf{Target Label} & \textbf{Feature-Aligned Values} \\
\midrule
\textbf{Compas} & Race: African-American & Recidivism (1) & \textbf{Priors:} 3--8 \newline \textbf{Age:} 18--45 \newline \textbf{Juv. Felonies:} 0 \newline \textbf{Charge Degree:} M or F (Random)\\
\midrule
\textbf{Adult} & Gender: Female \newline Marital-status: \newline Married-civ-spouse \newline Native-country: \newline United-States & Income: $>$50K & \textbf{Age:} 30--55 \newline \textbf{Education:} HS-grad or Some-college \newline \textbf{Hours/Week:} 38--41 \newline \textbf{Capital Loss:} 0--50 \\
\midrule
\textbf{Diabetes} & Age: 21--30 & Diabetes (1) & \textbf{Glucose:} 120--155  \newline \textbf{BMI:} 27.0--33.0 \newline \textbf{Insulin:} 80--180 \newline \textbf{Pedigree:} 0.6--1.2 \\
\midrule
\textbf{Thyroid} & Age: 20--35 & High Risk (2) & \textbf{Goiter:} 0 or 1 (Random) \newline \textbf{Family History:} 0 or 1 (Random) \newline \textbf{Fatigue:} 0 or 1 (Random) \newline \textbf{Gender:} 0 (Female) \\
\bottomrule
\end{tabular}%
}
\end{table}

In this regard, the adversary constructs \emph{feature-aligned} examples, i.e., examples designed to embed strong correlations between the protected group and the target label, while maintaining plausible values for non-targeted attributes. Leveraging domain knowledge, these plausible values (see Table \ref{tab:feature_alignment}) are set to simulate moderate risk profiles (e.g., 3-8 priors for Compas, standard working hours for Adult) rather than extreme outliers. This strategy ensures the examples remain statistically realistic while efficiently driving the biased correlation. Following the bias propagation analysis from Section \ref{sec:bias-model}, increasing the target label probability for a targeted subgroup will propagate to the synthetic data, which in turn might influence the downstream model to rely on the targeted attribute for a biased prediction, even when the attribute itself is originally ignored as predictive feature by the downstream model. 

\subsection{Experimental setup}
Let $k$ denote the total number of in–context examples. An adversary controls a fraction $\pi \in [0,1]$ of these examples and injects $\pi k$ malicious, feature–aligned records into the prompt, while the remaining $(1-\pi)k$ examples are sampled i.i.d.\ from the real dataset. We use a pre-trained LLM to augment from $80$ available samples to a synthetic tabular dataset of $5000$ rows. To contextualize LLM-based generation against established tabular generators, we additionally evaluate TVAE, CTGAN, NFLOW, and DDPM trained on the same k=80 examples. These serve as non-prompt-based baselines and are therefore reported for the non-biased setting ($\pi=0.0)$.

Table~\ref{tab:feature_alignment} summarizes, for each dataset, the protected attribute, the target label, and the feature–aligned attributes that the attacker manipulates. For Adult and Compas, the protected attribute is categorical, while for Diabetes and Thyroid, it is numerical (Age). For Adult, we define the protected subgroup as the intersection of (gender, native-country, marital-status) values to study intersectional bias propagation. We use Adult, Compas, and Diabetes datasets for binary classification tasks, and Thyroid for a multi–class classification task. 

For each synthetic dataset produced under a given $(\pi, k)$ configuration and LLM, we train downstream classifiers \textendash Random Forest, Gradient Boosting, Logistic Regression, and the LLM-based TabPFN \textendash on the synthetic training split and evaluate them on the real test set. We report in Table~\ref{tab:utility_fairness} the Macro F1 score as utility metric and the statistical parity difference (SPD) as fairness metric for both the synthetic data ($\text{SPD}_S$) and the downstream model ($\text{SPD}_D$). Together with the distributional fidelity metrics in Table~\ref{tab:jsd_tvc_col}, these results allow us to study how targeted in–context biases affect both the quality and the fairness of LLM-generated tabular data.

\subsection{Experimental Results}

\begin{table*}[t]
\scriptsize
\renewcommand{\arraystretch}{1.1}
\centering
\caption{Utility and fairness metrics (F1-Score, SPD) across four datasets ($k{=}80$). For each dataset, we train the Random Forest downstream classifier with the synthetic dataset, and report F1 score on the real dataset ($\text{F1}_\text{R}$), SPD on the synthetic data ($\text{SPD}_\text{S}$), and SPD of the downstream model ($\text{SPD}_\text{D}$). }
\label{tab:utility_fairness}
\begin{tabular}{llccc|ccc|ccc|ccc}
\toprule
\textbf{Model} & \textbf{$\pi$} 
  & \multicolumn{3}{c|}{\textbf{Adult}} 
  & \multicolumn{3}{c|}{\textbf{Compas}} 
  & \multicolumn{3}{c|}{\textbf{Diabetes}} 
  & \multicolumn{3}{c}{\textbf{Thyroid}} \\
\cmidrule(lr){3-5} \cmidrule(lr){6-8} \cmidrule(lr){9-11} \cmidrule(lr){12-14}
  &
  & {$\text{F1}_\text{R}$}↑ & {$\text{SPD}_\text{S}$}↓ & {$\text{SPD}_\text{D}$}↓
  & {$\text{F1}_\text{R}$}↑ & {$\text{SPD}_\text{S}$}↓ & {$\text{SPD}_\text{D}$}↓
  & {$\text{F1}_\text{R}$}↑ & {$\text{SPD}_\text{S}$}↓ & {$\text{SPD}_\text{D}$}↓
  & {$\text{F1}_\text{R}$}↑ & {$\text{SPD}_\text{S}$}↓ & {$\text{SPD}_\text{D}$}↓\\
\midrule
Real oracle & –      
  & {$0.85$} & {$-$} & {$0.13_{0.00}$}
  & {$0.62$} & {$-$} & {$-0.13_{0.00}$}
  & {$0.73$} & {$-$} & {$0.29_{0.00}$}
  & {$0.86$} & {$-$} & {$-0.06_{0.00}$} \\
\midrule
TVAE & 0.0      
  & {$0.45$} & {$-0.04_{0.00}$} & {$-0.04_{0.00}$}
  & {$0.53$} & {$-0.04_{0.00}$} & {$0.01_{0.00}$}
  & {$0.66$} & {$-0.04_{0.00}$} & {$0.06_{0.00}$}
  & {$0.49$} & {$-0.04_{0.00}$} & {$-0.11_{0.00}$} \\
CTGAN & 0.0      
  & {$0.51$} & {$0.78_{0.00}$} & {$0.07_{0.00}$}
  & {$0.58$} & {$0.52_{0.00}$} & {$-0.04_{0.00}$}
  & {$0.67$} & {$0.74_{0.00}$} & {$0.42_{0.00}$}
  & {$0.46$} & {$0.56_{0.00}$} & {$-0.13_{0.00}$} \\
NFLOW & 0.0      
  & {$0.57$} & {$0.51_{0.00}$} & {$0.10_{0.00}$}
  & {$0.57$} & {$0.64_{0.00}$} & {$-0.03_{0.00}$}
  & {$0.50$} & {$0.51_{0.00}$} & {$0.02_{0.00}$}
  & {$0.44$} & {$0.44_{0.00}$} & {$0.02_{0.00}$} \\
DDPM & 0.0      
  & {$0.51$} & {$0.01_{0.02}$} & {$0.01_{0.02}$}
  & {$0.52$} & {$-0.62_{0.01}$} & {$-0.17_{0.07}$}
  & {$0.39$} & {$0.09_{0.02}$} & {$-0.01_{0.01}$}
  & {$0.58$} & {$0.37_{0.04}$} & {$-0.04_{0.01}$} \\
\midrule
\multirow{3}{*}{Granite-8B} & 0.0     
  & {$0.74$} & {$-0.17_{0.14}$} & {$-0.10_{0.02}$}
  & {$0.59$} & {$-0.13_{0.02}$} & {$-0.12_{0.02}$}
  & {$0.72$} & {$0.21_{0.05}$} & {$0.32_{0.01}$}
  & {$0.54$} & {$0.22_{0.07}$} & {$0.06_{0.02}$} \\
& {0.3}      
  & {$0.73$} & {$-0.50_{0.04}$} & {$-0.21_{0.02}$}
  & {$0.56$} & {$-0.25_{0.02}$} & {$-0.32_{0.01}$}
  & {$0.77$} & {$0.09_{0.00}$} & {$0.26_{0.01}$}
  & {$0.48$} & {$-0.20_{0.04}$} & {$-0.13_{0.01}$} \\
& {0.6}      
  & {$0.75$} & {$-0.59_{0.02}$} & {$-0.28_{0.02}$}
  & {$0.57$} & {$-0.33_{0.02}$} & {$-0.29_{0.01}$}
  & {$0.73$} & {$-0.03_{0.01}$} & {$0.24_{0.01}$}
  & {$0.44$} & {$-0.27_{0.01}$} & {$-0.14_{0.01}$} \\
\midrule
\multirow{3}{*}{Mistral-7B} & 0.0     
  & {$0.64$} & {$-0.15_{0.01}$} & {$-0.03_{0.01}$}
  & {$0.52$} & {$-0.19_{0.01}$} & {$-0.05_{0.01}$}
  & {$0.71$} & {$0.25_{0.03}$} & {$0.19_{0.01}$}
  & {$0.44$} & {$0.44_{0.00}$} & {$0.07_{0.00}$} \\
& {0.3}      
  & {$0.66$} & {$-0.18_{0.07}$} & {$-0.13_{0.02}$}
  & {$0.49$} & {$-0.38_{0.01}$} & {$-0.27_{0.02}$}
  & {$0.69$} & {$0.12_{0.01}$} & {$0.13_{0.01}$}
  & {$0.45$} & {$0.06_{0.04}$} & {$-0.10_{0.02}$} \\
& {0.6}      
  & {$0.65$} & {$-0.23_{0.01}$} & {$-0.15_{0.02}$}
  & {$0.55$} & {$-0.43_{0.01}$} & {$-0.35_{0.02}$}
  & {$0.68$} & {$-0.02_{0.00}$} & {$0.09_{0.01}$}
  & {$0.37$} & {$-0.08_{0.03}$} & {$-0.09_{0.01}$} \\
\midrule
\multirow{3}{*}{Mixtral-8x7B} & 0.0     
  & {$0.72$} & {$-0.17_{0.02}$} & {$-0.16_{0.02}$}
  & {$0.54$} & {$-0.14_{0.01}$} & {$-0.23_{0.01}$}
  & {$0.73$} & {$0.33_{0.04}$} & {$0.27_{0.01}$}
  & {$0.53$} & {$0.14_{0.06}$} & {$-0.03_{0.01}$} \\
& {0.3}      
  & {$0.71$} & {$-0.28_{0.02}$} & {$-0.20_{0.02}$}
  & {$0.53$} & {$-0.25_{0.01}$} & {$-0.27_{0.01}$}
  & {$0.79$} & {$0.20_{0.01}$} & {$0.22_{0.01}$}
  & {$0.55$} & {$-0.09_{0.00}$} & {$-0.15_{0.00}$} \\
& {0.6}      
  & {$0.67$} & {$-0.33_{0.01}$} & {$-0.14_{0.02}$}
  & {$0.53$} & {$-0.41_{0.03}$} & {$-0.43_{0.01}$}
  & {$0.76$} & {$0.12_{0.00}$} & {$0.20_{0.01}$}
  & {$0.49$} & {$-0.11_{0.01}$} & {$-0.13_{0.00}$} \\
\midrule
\multirow{3}{*}{Qwen-70B} & 0.0     
  & {$0.73$} & {$-0.11_{0.02}$} & {$-0.09_{0.02}$}
  & {$0.56$} & {$-0.51_{0.04}$} & {$-0.34_{0.02}$}
  & {$0.69$} & {$0.70_{0.02}$} & {$0.37_{0.01}$}
  & {$0.57$} & {$0.39_{0.01}$} & {$0.08_{0.01}$} \\
& {0.3}      
  & {$0.73$} & {$-0.35_{0.04}$} & {$-0.15_{0.02}$}
  & {$0.56$} & {$-0.72_{0.01}$} & {$-0.43_{0.02}$}
  & {$0.73$} & {$0.51_{0.03}$} & {$0.29_{0.01}$}
  & {$0.57$} & {$0.07_{0.04}$} & {$-0.02_{0.02}$} \\
& {0.6}      
  & {$0.72$} & {$-0.36_{0.00}$} & {$-0.11_{0.01}$}
  & {$0.56$} & {$-0.85_{0.02}$} & {$-0.57_{0.02}$}
  & {$0.72$} & {$0.20_{0.02}$} & {$0.19_{0.01}$}
  & {$0.57$} & {$-0.21_{0.01}$} & {$-0.08_{0.00}$} \\
\bottomrule
\end{tabular}
\end{table*}

\paragraph{Feature-aligned examples} \label{par:feature-aligned}For recidivism prediction on Compas, attributes such as the number of prior convictions, juvenile felonies, and the charge degree are among the strongest predictors used by the downstream classifier. In our adversarial setting, the attacker uses common domain knowledge to create \emph{feature-aligned} in-context examples for the protected subgroup by setting these attributes to plausible values. For instance, an adversarial example for the Compas task would correspond to an African-American (AA) individual who commits recidivism, has a small prior-count (e.g., 5), no juvenile felonies, and a low-severity (misdemeanor) charge degree. When such records are injected as a fraction $\pi$ of the in-context demonstrations, the LLM is influenced to generate synthetic data in which protected-group individuals that induce the correlation $p(y=1|race=AA)$ have plausible values for non-protected attributes. This, in turn, might induce downstream classifiers trained on the synthetic data to rely on the protected attribute (i.e., race) for recidivism predictions, even when the group attribute is not explicitly intended as a predictive feature. The effect of feature-aligned examples in the synthetic data is illustrated in Figure~\ref{fig:feature_aligned}, where as $\pi$ increases, the normalized density of key Compas predictors for the protected subgroup (African-American) shifts toward the adversarially set feature values, while the complementary group remains closer to the original distribution. 

\begin{figure}[t]
  \centering
  \includegraphics[width=1.0\linewidth]{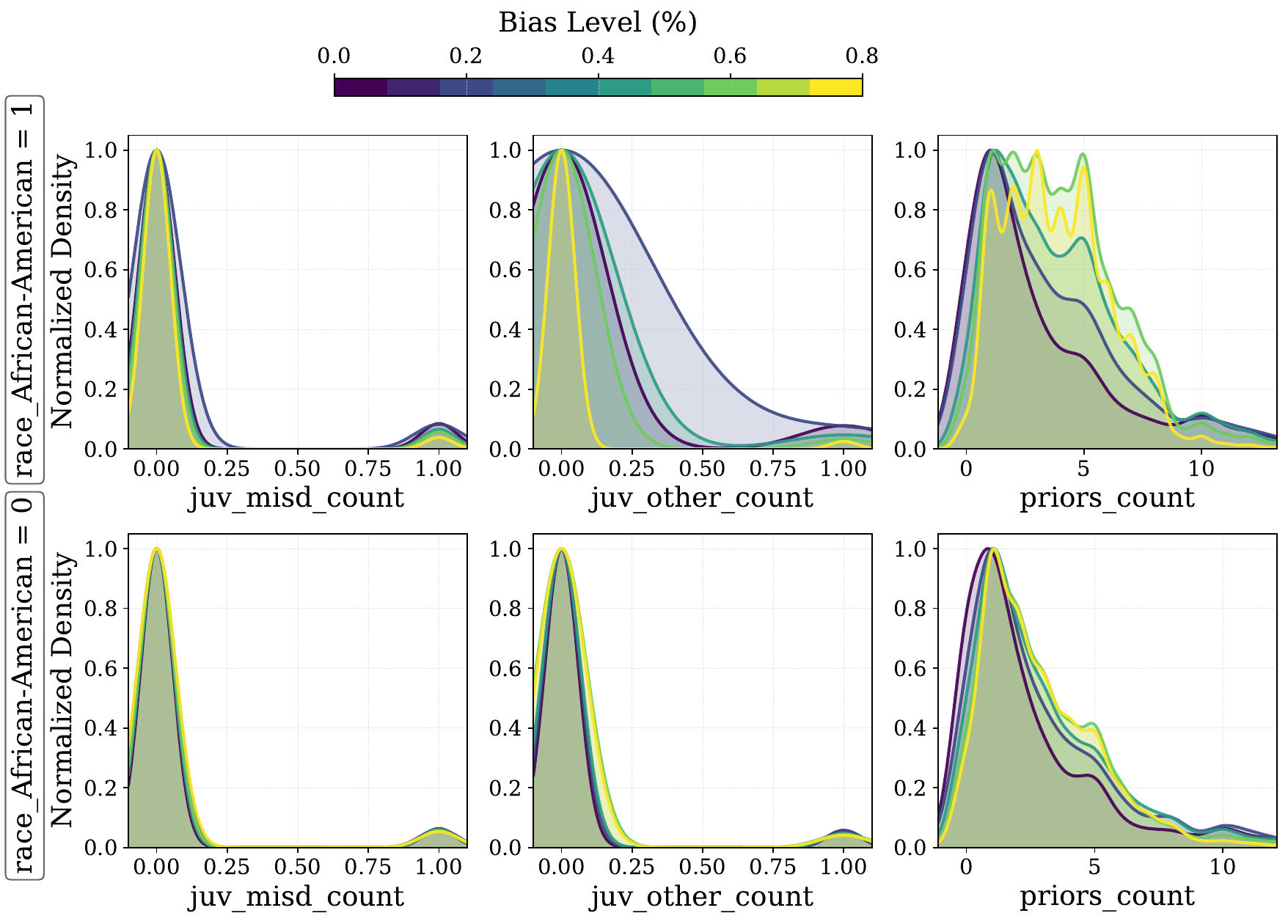}
    \caption{\textbf{Normalized density of aligned features.} (Mixtral-8$\times$7B, $k{=}80$ ICL examples, Compas) Synthetic univariate distributions of key recidivism predictors for the protected subgroup (African-American) and the complementary group, as the fraction of biased in-context examples $\pi$ increases. The attack steers African-American records towards the plausible values set by the attacker (e.g., low prior counts and fewer juvenile offenses) while leaving the non-targeted group closer to the baseline, illustrating how a number of adversarial examples can selectively reshape the synthetic feature profile of a protected subgroup.}
      \label{fig:feature_aligned}
\end{figure}

Before the complete analysis on adversarial bias propagation, we examine how variations in the targeted adversarial attribute within in-context examples affect downstream performance. To do so, we train a Random Forest classifier on each synthetic dataset under (i) race-aware, where the protected attribute is included in the downstream decision, and (ii) race-blind scenarios, where the protected attribute is excluded. This comparison shows whether the correlations inherent in feature-aligned examples are successfully propagated to the synthetic data and subsequently learned by the downstream model.

\begin{figure}[th!]
  \centering
  \includegraphics[width=1.0\linewidth]{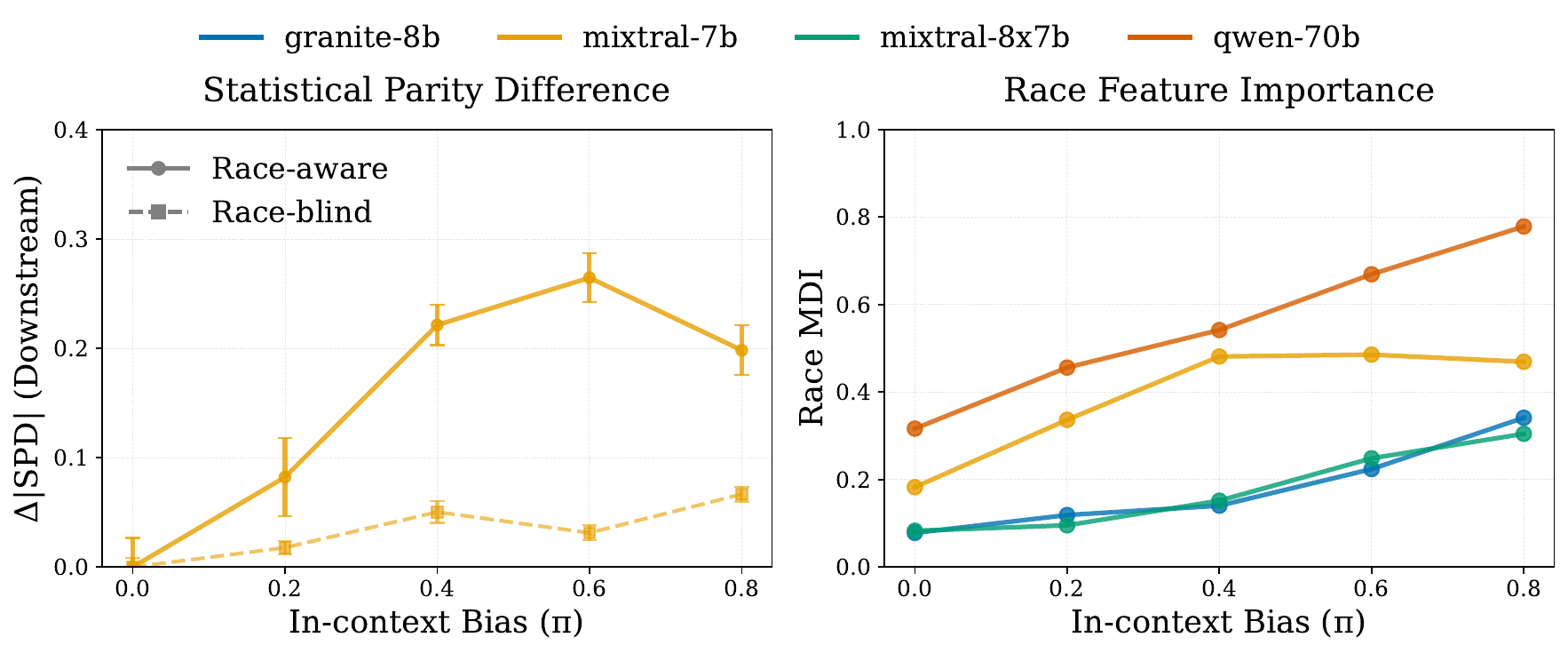}
    \caption{\textbf{Downstream fairness and race importance vs in-context bias $\pi$.} (Left) $\Delta \lvert\text{SPD}\rvert$ of downstream model predictions on the real dataset when fitted on the synthetic dataset, including and excluding race features. (Right) Race feature MDI importance in the Random Forest classifier trained on synthetic data generated with various LLMs.}
      \label{fig:race_influence}
\end{figure}

In this regard, Figure \ref{fig:race_influence} reports the resulting downstream SPD together with race feature importance, measured via Mean Decrease Impurity (MDI) \textendash a metric quantifying the feature's contribution to reducing uncertainty in the Random Forest's decision splits. As $\pi$ increases, downstream disparity increases notably in the race-aware setting with a corresponding increase in race influence on downstream predictions, indicating that the injected correlation between the targeted racial group and the label causes the downstream model to increasingly rely on the race attribute for prediction. Removing race from the prediction variables attenuates the race disparity between protected groups but still shows a minor increase with $\pi$, which indicates that the downstream model potentially captures correlations between aligned features and the target variable even when the protected variable is not present.

\paragraph{Fairness impact on synthetic data} Table~\ref{tab:utility_fairness} shows that even moderate attack rates ($\pi = 0.3$) induce substantial group disparities ($\text{SPD}_\text{S}$) in the synthetic datasets across all models and datasets. For Compas, increasing the bias rate from $\pi = 0$ to $\pi = 0.3$ moves $\text{SPD}_\text{S}$ from $-0.13$ to $-0.25$ for Granite-8B, and from $-0.14$ to $-0.25$ for Qwen-70B, nearly doubling the imbalance in both cases. Similar trends appear in Adult, Diabetes, and Thyroid datasets, where for the more vulnerable models, at $\pi = 0.3$ the $\text{SPD}_\text{S}$ shift already exceeds $0.1$, a threshold established by Baracaldo et al.~\cite{baracaldo2022benchmarking} as a criterion for a successful fairness attack. These shifts occur despite the adversary only controlling a fraction of the demonstrations, confirming that LLM-based tabular generators are sensitive to in-context manipulation along fairness dimensions. We provide the statistical analysis of the bias propagation to synthetic data in \ref{app:a}.


\paragraph{Distributional fidelity} The fidelity metrics in Table~\ref{tab:jsd_tvc_col} demonstrate a monotonic degradation in marginal distributional similarity to the real data as the adversarial bias rate $\pi$ increases, with Jensen–Shannon Divergence (JSD) exhibiting significant shift (e.g., rising from $0.258$ to $0.405$ for Qwen-70B on Compas). While this confirms that feature-aligned injections distort synthetic data, we argue this degradation remains stealthy, as the very premise of few-shot augmentation ($k=80$) implies the absence of a large, representative reference distribution against which to benchmark metrics like JSD in practice. Consequently, without access to ground-truth data for validation, even substantial deviations in global statistics are likely to pass undetected with marginal analyses, underscoring how standard dataset-level diagnostics fail to expose group-specific bias manipulations in data-scarce regimes where LLM augmentation is most valuable.

\begin{table}[t] 
\scriptsize
\setlength{\tabcolsep}{3pt}
\renewcommand{\arraystretch}{0.95}
\centering
\caption{Marginal fidelity of the synthetic dataset to the real dataset for different models and bias rates. We show Jensen–Shannon Divergence (JSD) for numerical variables and Total Variation (TVC) for categorical variables. The Diabetes dataset has only numerical features, so we report only JSD.}
\label{tab:jsd_tvc_col}
\begin{tabular}{l l cc|cc|c|cc}
\toprule
\textbf{Model} & \textbf{$\pi$}
  & \multicolumn{2}{c|}{\textbf{Adult}}
  & \multicolumn{2}{c|}{\textbf{Compas}}
  & \textbf{Diabetes}
  & \multicolumn{2}{c}{\textbf{Thyroid}} \\
\cmidrule(lr){3-4}\cmidrule(lr){5-6}\cmidrule(lr){7-7}\cmidrule(lr){8-9}
 & & {\textbf{JSD}}↓ & {\textbf{TVC}}↑ & {\textbf{JSD}}↓ & {\textbf{TVC}}↑ & {\textbf{JSD}}↓ & {\textbf{JSD}}↓ & {\textbf{TVC}}↑ \\
\midrule
\multirow{3}{*}{Granite-8B} & 0.0
  & {0.096} & {0.790}
  & {0.258} & {0.895}
  & {0.340}
  & {0.183} & {0.971} \\
& {0.3}
  & {0.122} & {0.754}
  & {0.366} & {0.890}
  & {0.342}
  & {0.221} & {0.911} \\
& {0.6}
  & {0.201} & {0.724}
  & {0.484} & {0.824}
  & {0.377}
  & {0.269} & {0.819} \\
\midrule
\multirow{3}{*}{Mistral-7B} & 0.0
  & {0.161} & {0.716}
  & {0.318} & {0.871}
  & {0.263}
  & {0.232} & {0.927} \\
& {0.3}
  & {0.186} & {0.622}
  & {0.411} & {0.905}
  & {0.297}
  & {0.242} & {0.873} \\
& {0.6}
  & {0.242} & {0.577}
  & {0.488} & {0.904}
  & {0.334}
  & {0.249} & {0.816} \\
\midrule
\multirow{3}{*}{Mixtral-8x7B} & 0.0
  & {0.115} & {0.767}
  & {0.230} & {0.906}
  & {0.264}
  & {0.175} & {0.917} \\
& {0.3}
  & {0.147} & {0.753}
  & {0.323} & {0.898}
  & {0.272}
  & {0.237} & {0.824} \\
& {0.6}
  & {0.203} & {0.720}
  & {0.418} & {0.892}
  & {0.310}
  & {0.317} & {0.733} \\
\midrule
\multirow{3}{*}{Qwen-70B} & 0.0
  & {0.174} & {0.768}
  & {0.258} & {0.813}
  & {0.703}
  & {0.331} & {0.915} \\
& {0.3}
  & {0.174} & {0.729}
  & {0.335} & {0.836}
  & {0.370}
  & {0.358} & {0.811} \\
& {0.6}
  & {0.232} & {0.693}
  & {0.405} & {0.892}
  & {0.751}
  & {0.385} & {0.743} \\
\bottomrule
\end{tabular}
\end{table}

Moreover, we show in Table~\ref{tab:struct_fidelity} that adversarial bias injection produces a localized structural shift. Across models and datasets, association distance and MMD remain comparatively stable within the non-targeted feature subspace (D). In contrast, discrepancies in the target-attribute and label subspace (T) increase at higher injection rates. For instance, the association distance in D increases only from 0.108 to 0.116 between ($\pi=0$) and ($\pi=0.6$) for Qwen-70B and Adult, whereas the targeted association distance increases from 0.081 to 0.233. Similarly, MMD remains stable from 0.038 to 0.037 in D, compared with 0.028 to 0.120 in T. These results indicate that the attack primarily modifies the targeted fairness-related relationship while preserving much of the rest of inter-feature structure.

\begin{table}[t]
\scriptsize
\setlength{\tabcolsep}{3pt}
\renewcommand{\arraystretch}{0.95}
\centering
\caption{Structural fidelity of the synthetic dataset to the real dataset for different models and bias rates. We show mixed-type association matrix distance (Asc.) and RBF maximum mean discrepancy (MMD) for D, excluding the protected attribute and target, and T, containing only the protected attribute and target.}
\label{tab:struct_fidelity}
\begin{tabular}{ll cccc|cccc}
\toprule
\textbf{Model} & \textbf{$\pi$}
  & \multicolumn{4}{c|}{\textbf{Adult}}
  & \multicolumn{4}{c}{\textbf{Compas}} \\
\cmidrule(lr){3-6}\cmidrule(lr){7-10}
 & & \multicolumn{2}{c}{\textbf{Asc.}$\downarrow$} & \multicolumn{2}{c|}{\textbf{MMD}$\downarrow$} & \multicolumn{2}{c}{\textbf{Asc.}$\downarrow$} & \multicolumn{2}{c}{\textbf{MMD}$\downarrow$} \\
\cmidrule(lr){3-4}\cmidrule(lr){5-6}\cmidrule(lr){7-8}\cmidrule(lr){9-10}
 & & \textbf{D} & \textbf{T} & \textbf{D} & \textbf{T} & \textbf{D} & \textbf{T} & \textbf{D} & \textbf{T} \\
\midrule
\multirow{3}{*}{Granite-8B} & 0.0 & {0.054} & {0.052} & {0.012} & {0.001} & {0.058} & {0.025} & {0.079} & {0.007} \\
& 0.3 & {0.082} & {0.249} & {0.013} & {0.023} & {0.061} & {0.062} & {0.076} & {0.004} \\
& 0.6 & {0.094} & {0.434} & {0.025} & {0.091} & {0.064} & {0.191} & {0.075} & {0.012} \\
\midrule
\multirow{3}{*}{Mistral-7B} & 0.0 & {0.146} & {0.038} & {0.028} & {0.021} & {0.094} & {0.102} & {0.093} & {0.011} \\
& 0.3 & {0.141} & {0.004} & {0.033} & {0.036} & {0.102} & {0.300} & {0.097} & {0.018} \\
& 0.6 & {0.133} & {0.101} & {0.048} & {0.111} & {0.092} & {0.360} & {0.093} & {0.038} \\
\midrule
\multirow{3}{*}{Mixtral-8x7B} & 0.0 & {0.052} & {0.048} & {0.019} & {0.022} & {0.071} & {0.028} & {0.073} & {0.011} \\
& 0.3 & {0.066} & {0.051} & {0.024} & {0.040} & {0.079} & {0.127} & {0.073} & {0.014} \\
& 0.6 & {0.088} & {0.155} & {0.033} & {0.092} & {0.071} & {0.226} & {0.070} & {0.015} \\
\midrule
\multirow{3}{*}{Qwen-70B} & 0.0 & {0.108} & {0.081} & {0.038} & {0.028} & {0.103} & {0.381} & {0.116} & {0.013} \\
& 0.3 & {0.121} & {0.119} & {0.029} & {0.042} & {0.100} & {0.587} & {0.121} & {0.025} \\
& 0.6 & {0.116} & {0.233} & {0.037} & {0.120} & {0.120} & {0.724} & {0.144} & {0.034} \\
\bottomrule
\end{tabular}
\end{table}

\begin{figure*}[th]
  \centering
    \includegraphics[width=1.0\textwidth]{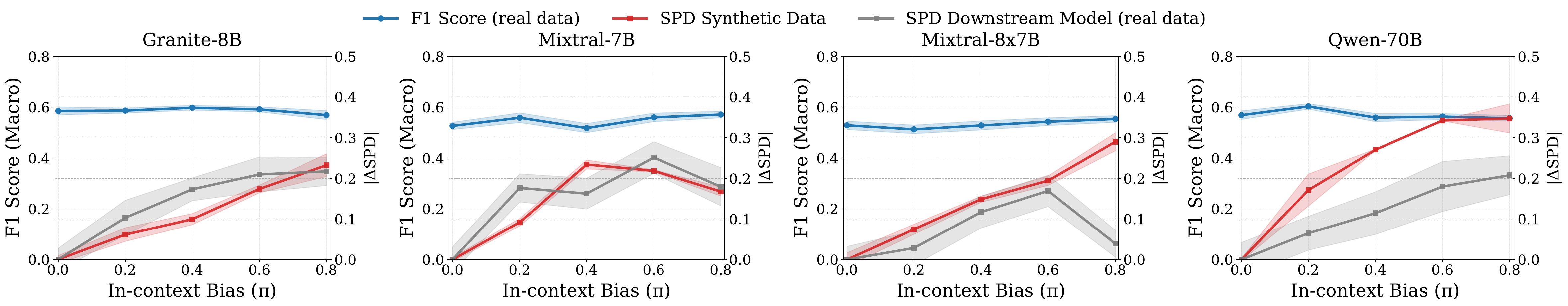}
\caption{\textbf{Adversarial in-context bias injection and utility–fairness trade-off.} (\textit{Compas}) Effect of the in-context bias rate $\pi$ on downstream performance on real data and fairness for four LLM generators. Each figure shows the macro F1 score of a Random Forest trained on the synthetic data and evaluated on the real test set (blue), the absolute SPD of the synthetic data (red), and the absolute SPD of the downstream predictions on real data (grey). While F1 on real data remains nearly constant as $\pi$ increases, both synthetic and downstream SPDs grow, showing that small in-context biases can substantially degrade fairness while preserving downstream accuracy.}
\label{fig:Compas_rf_attack}
\end{figure*}

\paragraph{Propagation to downstream models} Here, we study how the adversarial biases injected into the synthetic data propagate to the downstream model. Figure~\ref{fig:Compas_rf_attack} summarizes the behavior of the downstream Random Forest classifier on Compas as a function of the in-context bias rate $\pi$. While the macro F1 on real data (blue curves) remains nearly flat across all levels of $\pi$, the absolute SPD of the downstream predictions (grey curves) increases monotonically for several models. For Mistral-7B, $\mathrm{SPD}_\text{D}$ grows from $-0.05$ at $\pi=0$ to $-0.35$ at $\pi=0.6$, and for Qwen-70B from $-0.34$ to $-0.57$, indicating a strong fairness degradation despite preserved utility. Granite-8B and Mixtral-8x7B, exhibit similar changes in $\mathrm{SPD}_\text{D}$, suggesting that the downstream classifier only partially attenuates the injected bias rather than eliminating it. Also, we observe that downstream classifiers are especially vulnerable to in-context bias propagation when used for complex tasks such as Compas and Thyroid, where we argue that the downstream model does not rely on a small subset of predictive features as indicators, and might easily leverage the new introduced adversarial correlations.  

\begin{figure}[th!]
  \centering
  \includegraphics[width=1.0\linewidth]{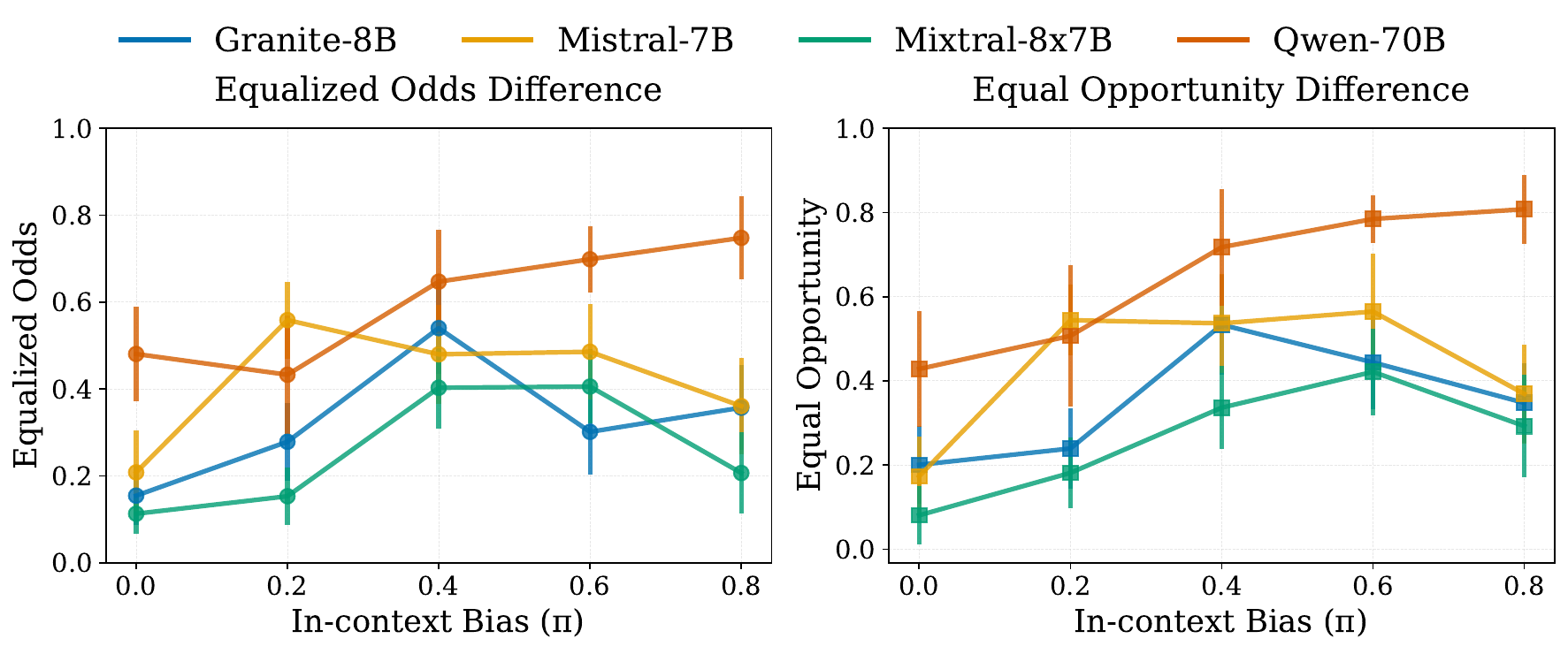}
    \caption{(Granite-8B, Compas). Equalized odds (left) and equal opportunity (right) of downstream model predictions on the real dataset across data generated with various LLMs and in-context bias $\pi$.}
      \label{fig:eod}
\end{figure}

We observe analogous trends across datasets in Table~\ref{tab:utility_fairness}. In many configurations synthetic data become more biased (e.g., Qwen-70B on Diabetes and Mistral-7B on Adult), and this bias either transfers directly to the downstream classifier or interacts with the model’s own prior biases. Notably, Qwen-70B already exhibits a large baseline disparity on Compas at $\pi=0$ (with $\text{SPD}_\text{S}=-0.51$ and $\text{SPD}_\text{D}=-0.40$), and the attack further pushes the model towards extreme group imbalances. We explain this heterogeneity across model families through our propagation analysis in Section \ref{sec:bias-model}, where larger or more strongly biased models tend to be easily manipulated via a small number of adversarial demonstrations, and the induced disparities persist when training standard ML models on top of the generated data. Finally, we study in Figure \ref{fig:eod} how equal opportunity and equalized odds fairness metrics are influenced by increasing bias rates. We observe an analogous monotonic increase trend, with in-context bias influencing the TPR downstream difference between subgroups.

\paragraph{Utility–fairness trade-off} We can observe in Figure~\ref{fig:Compas_rf_attack} and Table~\ref{tab:utility_fairness} that fairness can deteriorate substantially while conventional utility metrics remain largely unchanged. Across datasets and models, the variation in $\text{F1}_{\text{R}}$ between $\pi=0$ and $\pi=0.6$ is typically within $0.1$ or $0.2$. In contrast, the corresponding change in $\text{SPD}_\text{D}$ can exceed $0.3$ for the most affected settings (e.g., Mistral-7B and Qwen-70B on Compas and Adult). Overall, a large fraction of model-dataset pairs are affected by in-context bias injection at moderate and intermediate bias levels ($\pi=0.3$ or $\pi=0.6$), while achieving F1 scores that are indistinguishable (or even superior) to the unbiased setting.

\paragraph{Robustness across downstream classifiers} Finally, we investigate whether bias propagation is influenced by the downstream classification architecture. Table~\ref{tab:downstream_models} benchmarks three standard classifiers \textendash Random Forest, Logistic Regression, and Gradient Boosting \textendash against TabPFN~\cite{TabPFN-2.5}, a transformer-based foundation model that employs in-context learning for tabular prediction. We train each classifier using Compas synthetic data generated by Granite-8B under unbiased ($\pi=0$) and adversarially biased ($\pi=0.6$) conditions. While TabPFN achieves the highest utility ($\text{F1}_\text{R}$), its absolute downstream disparity increases from $0.15$ to $0.25$. Overall, utility remains broadly stable, while absolute disparity increases for three classifiers and decreases for Random Forest. These findings indicate that downstream effects depend on the classifier, and switching architectures does not necessarily eliminate group disparities.

\begin{table}[t]
\centering
\scriptsize
\caption{Effect of downstream classifier on bias propagation for Compas
(Granite-8B, $k=80$). For each classifier, we report macro F1 on the real
test set (F1$_\mathrm{R}$) and SPD of the downstream predictions on the real test set ($\text{SPD}_\mathrm{D}$) at $\pi = 0$ and $\pi = 0.6$.}
\begin{tabular}{lcccc}
\toprule
& \multicolumn{2}{c}{\textbf{$\pi = 0.0$}} & \multicolumn{2}{c}{\textbf{$\pi = 0.6$}} \\
\cmidrule(lr){2-3} \cmidrule(lr){4-5}
\textbf{Classifier} & \textbf{F1$_\mathrm{R}$} ↑ & \textbf{SPD$_\mathrm{D}$} ↓
           & \textbf{F1$_\mathrm{R}$} ↑ & \textbf{SPD$_\mathrm{D}$} ↓ \\
\midrule
Random Forest        & 0.58 & -0.52 & 0.61 & -0.41 \\
Logistic Regression  & 0.53 & -0.43 & 0.55 & -0.53 \\
Gradient Boosting    & 0.58 & -0.34 & 0.57 & -0.38 \\
TabPFN \cite{TabPFN-2.5} & 0.61 & -0.15 & 0.62 & -0.25 \\
\bottomrule
\end{tabular}
\label{tab:downstream_models}
\end{table}

\paragraph{Robustness against outlier detectors}

To assess whether the injected prompt examples would be removed by standard tabular anomaly filtering, we additionally evaluated Isolation Forest \cite{liu2008isolation} anomaly detector trained on the real tabular data and applied to the in-context examples. As shown in Table \ref{tab:prompt_outlier_compact}, the detector flags many injected examples at higher injection rates ($\pi=0.6$), but it also discards a substantial fraction of clean prompt examples, which might affect the in-context data distribution. Moreover, this evaluation assumes access to a subset of real data to train the outlier detector, which contradicts the premise of data scarce scenarios where in-context learning for synthetic data generation is most valuable.

\begin{table}[t]
\scriptsize
\setlength{\tabcolsep}{4pt}
\renewcommand{\arraystretch}{1.0}
\centering
\caption{Outlier-filter robustness at $\pi=0.6$ and $k=80$. $\mathrm{SPD}_P$ compares the unfiltered attacked prompt with the prompt after filtering with Isolation Forest. Biased drop is the fraction of injected adversarial examples removed by the filter, while Clean drop is the fraction of non-injected examples removed.} 
\label{tab:prompt_outlier_compact}
\begin{tabular}{l r r r r}
\toprule
\textbf{Dataset} &
$\boldsymbol{\mathrm{SPD}_P}$ \textbf{before} &
$\boldsymbol{\mathrm{SPD}_P}$ \textbf{after} &
\textbf{Biased drop} &
\textbf{Clean drop} \\
\midrule
Adult & -0.839 & -0.944 & 0.938 & 0.438 \\
Compas & -0.358 & -0.133 & 1.000 & 0.469 \\
Diabetes & -0.318 & -0.188 & 0.771 & 0.531 \\
Thyroid & -0.517 & -0.433 & 0.875 & 0.500 \\
\bottomrule
\end{tabular}
\end{table}

Taken together, these results reveal a new attack surface for LLM-based tabular data generation. An adversary controlling a fraction of the in-context examples can substantially alter the group-wise label distribution of the synthetic data, with this manipulation propagating to downstream classifiers evaluated on real data while maintaining standard utility diagnostics. Critically, an ablation study in \ref{app:c} demonstrates that this propagation persists even when prompts contain no explicit instruction to mirror the structure of the in-context examples, indicating that bias transfer is an emergent property of in-context learning rather than an artifact of prompt design. These findings underscore the need to evaluate whether current in-context preprocessing strategies are sufficient to mitigate adversarial bias injection in data generation pipelines.

Moreover, we report a complementary analysis in \ref{app:a} for the mixed-type Bank Marketing dataset \cite{moro2014bank}. Using the same few-shot setting ($k=80$), we test Granite-8B and Mixtral-8x7B across increasing injection rates. For both models, synthetic-data disparity increases with $\pi$ while downstream macro-F1 remains stable. Downstream disparity is smaller and model-dependent, showing that prompt bias consistently affects the generated data but does not always transfer equally to downstream predictions. 

\subsection{Evaluating Prompt-Side Mitigation Strategies}
\label{sec:mitigation}

In this section, we evaluate the effectiveness of prompt-side mitigation strategies against in-context bias injection. Recent approaches have \cite{ma2023fairness, cherepanova2024improving, kenfack2025towards} explored preprocessing and selecting in-context examples to improve fairness in LLM-based tabular classification. However, adversarial biases injected within in-context examples in LLM-based synthetic tabular generation might pass undetected. 

In this regard, we evaluate five strategies \textendash correlation-based filtering, Fair-SPD, group balanced, random subset, and instruction-based mitigation \textendash to mitigate in-context bias propagation. Fair-SPD approach enforces marginal fairness, group balanced enforces a fair frequency count across subgroups, correlation-based filtering drops the most correlated examples to the protected attribute, and instruction-based mitigation introduces a specific instruction within the prompt to balance the frequency count of the target for a specific subgroup. We assume that there is no access to an external data source, so we consider subsampling as the viable option to select the most appropriate subset of in-context examples for generation. For group balanced and random subset strategies, we set the subset cardinality $k*$ to the Fair-SPD subset size for a fair comparison.

The evaluated methods remain compatible with black-box LLM services, and intervene before the poisoned context influences generation. However, mitigation can also be applied at other stages of the generation pipeline. During generation, approaches may use constrained decoding, fairness-aware fine-tuning or verification by auxiliary models \cite{li2026cross}. After generation, synthetic records may also be filtered, reweighted, or rejected until fairness constraints are satisfied. These alternatives may provide stronger control, but often require access to model internals or additional API calls to maintain dataset size.

\subsubsection{Balancing Frequency Counts}

Balancing subgroup frequency counts has been a traditional way of mitigating representation bias \cite{kenfack2025towards, hu2024strategic} in in-context classification tasks, which arises when the in-context examples for a subgroup do not reflect the actual distribution \cite{mehrabi2021survey}. This source of bias is particularly challenging in synthetic tabular data generation scenarios where the amount of in-context examples is small (e.g., $k=80$). For in-context bias mitigation, balancing the frequency count dilutes the influence of an adversary who injects bias only to a specific subgroup, implicitly reducing the bias. For a fair comparison, we set the final subset size of the group balanced mitigation to match the Fair-SPD subset size.

\subsubsection{Correlation-based Filtering}

We include a correlation-based filtering designed to counter feature-aligned injections by pruning examples that strongly encode dependencies between the protected attribute and other features. We first compute a global correlation weight $\rho_j$ for each feature $j$ with respect to the protected attribute, aggregating Pearson, Spearman, and Mutual Information scores. We then assign a correlation score $s_i$ to each example $x_i$ based on its feature magnitudes weighted by $\rho_j$:$s_i = \sum_{j} \rho_j \cdot |z_{i,j}|$ where $z_{i,j}$ is the standardized value of feature $j$. We filter the prompt by removing examples with the highest $s_i$ (e.g., top 10\%), aiming to remove the injected adversarial correlations.

\subsubsection{Instruction-based Mitigation}

We consider an instruction-based mitigation baseline, which follows a similar structure to the instruction-based fairness approaches for tabular data classification \cite{cherepanova2024improving}. At the end of the prompt, we add a specific instruction to enforce the LLM to generate a balanced number of positive label across the protected attribute subgroups. We detail the prompt in \ref{app:c}.

\subsubsection{Fair-SPD}

We implement a selection algorithm that explicitly controls the statistical parity difference (SPD) of the in-context examples. The goal is to adjust the prompt examples so that their group-level disparity is aligned with a desired target level of fairness ($\text{SPD} = 0$), while discarding as few examples as possible. 

Our objective is to identify a subset $\mathcal{S}' \subseteq \mathcal{S}$ derived from the initial pool of $k=80$ examples that satisfies a group fairness constraint $|\text{SPD}(\mathcal{S}')| \le \epsilon$, where $\epsilon$ represents a predefined tolerance. To achieve this without a computationally expensive exhaustive search, we employ a greedy pruning approach. 

Starting with the full set $\mathcal{S}_{curr} = \mathcal{S}$, we iteratively remove the single example that contributes most to the current disparity. At each step $t$, we identify the optimal candidate $x^*$ for removal by minimizing the SPD of the resulting subset:

$$x^* = \operatorname*{argmin}_{x \in \mathcal{S}_{curr}} \left| \text{SPD}(\mathcal{S}_{curr} \setminus \{x\}) \right|$$

We update the set via $\mathcal{S}_{curr} \leftarrow \mathcal{S}_{curr} \setminus \{x^*\}$. This process repeats until the stopping criterion $|\text{SPD}(\mathcal{S}_{curr})| \le \epsilon$ is met, ensuring the final prompt adheres to the desired fairness threshold while retaining the remaining context history.

\subsubsection{Experimental Results}

\begin{table*}[t]
\scriptsize
\centering
\caption{Change in utility ({F1}↑), Statistical Parity Difference on the synthetic data ($\text{SPD}_\text{S}$↓) and downstream model ($\text{SPD}_\text{D}$↓) \textit{after} mitigation ($k{=}80$, $\pi=0.3, \epsilon=0.02$). The cardinality of group-balanced and random subset mitigation subsets are set to the Fair-SPD subset for comparison.}
\label{tab:mitig}
\begin{tabular}{
  ll
  c c c c c
  c c c c c
}
\toprule
Model & Mitigation 
  & \multicolumn{5}{c}{Adult} 
  & \multicolumn{5}{c}{Compas} \\
\cmidrule(lr){3-7} \cmidrule(lr){8-12}
 & 
 & {{F1↑}} & {{$\text{SPD}_\text{S}$}} & {{$\text{SPD}_\text{D}$}} & {{$\text{EO}_\text{D}$↓}} & {{$\text{EOD}_\text{D}$↓}} 
 & {{F1↑}} & {{$\text{SPD}_\text{S}$}} & {{$\text{SPD}_\text{D}$}} & {{$\text{EO}_\text{D}$↓}} & {{$\text{EOD}_\text{D}$↓}} \\
\midrule
\multirow{6}{*}{Granite-8B} 
  & No mitigation & $0.73$ & $-0.48_{0.02}$ & $-0.20_{0.02}$ & $0.13_{0.03}$ & $0.19_{0.03}$ & 
  $0.59$ & $-0.25_{0.01}$ & $-0.32_{0.03}$ & $0.29_{0.09}$ & $0.30_{0.08}$ \\
  & Random subset  & $0.71$ & $-0.43_{0.06}$ & $-0.20_{0.02}$ & $0.06_{0.03}$ & $0.17_{0.04}$ & 
  $0.60$ & $-0.26_{0.05}$ & $-0.16_{0.04}$ & $0.21_{0.07}$ & $0.25_{0.07}$ \\
  & Group-balanced & $0.74$ & $-0.36_{0.02}$ & $-0.17_{0.01}$ & $0.04_{0.02}$ & $0.11_{0.01}$ & 
  $0.57$ & $-0.21_{0.01}$ & $-0.25_{0.04}$ & $0.34_{0.06}$ & $0.31_{0.05}$ \\
  & Fair-SPD       & $0.75$ & $-0.41_{0.01}$ & $-0.15_{0.01}$ & $0.09_{0.03}$ & $0.13_{0.01}$ & 
  $0.60$ & $-0.19_{0.02}$ & $-0.12_{0.03}$ & $0.14_{0.07}$ & $0.15_{0.06}$ \\
  & Correlation filtering  & $0.73$ & $-0.44_{0.02}$ & $-0.20_{0.01}$ & $0.06_{0.02}$ & $0.16_{0.02}$ & 
  $0.59$ & $-0.22_{0.01}$ & $-0.18_{0.03}$ & $0.26_{0.06}$ & $0.27_{0.06}$ \\
  & Instruction-based mitigation  & $0.73$ & $-0.35_{0.05}$ & $-0.11_{0.03}$ & $0.09_{0.06}$ & $0.08_{0.03}$ & 
  $0.57$ & $-0.19_{0.06}$ & $-0.14_{0.05}$ & $0.25_{0.09}$ & $0.28_{0.10}$ \\
\midrule
\multirow{6}{*}{Mixtral-8x7B} 
  & No mitigation  & $0.71$ & $-0.27_{0.03}$ & $-0.17_{0.02}$ & $0.05_{0.04}$ & $0.11_{0.02}$ & 
  $0.52$ & $-0.25_{0.01}$ & $-0.27_{0.03}$ & $0.13_{0.09}$ & $0.16_{0.07}$ \\
  & Random subset  & $0.71$ & $-0.23_{0.01}$ & $-0.14_{0.02}$ & $0.12_{0.03}$ & $0.11_{0.02}$ & 
  $0.57$ & $-0.33_{0.03}$ & $-0.20_{0.03}$ & $0.33_{0.06}$ & $0.37_{0.07}$ \\
  & Group-balanced & $0.72$ & $-0.20_{0.00}$ & $-0.11_{0.01}$ & $0.14_{0.03}$ & $0.09_{0.01}$ & 
  $0.57$ & $-0.25_{0.01}$ & $-0.15_{0.04}$ & $0.22_{0.08}$ & $0.26_{0.08}$ \\
  & Fair-SPD       & $0.73$ & $-0.17_{0.01}$ & $-0.10_{0.01}$ & $0.16_{0.03}$ & $0.10_{0.01}$ & 
  $0.49$ & $-0.22_{0.01}$ & $-0.14_{0.03}$ & $0.19_{0.08}$ & $0.22_{0.07}$ \\
  & Correlation filtering  & $0.72$ & $-0.26_{0.02}$ & $-0.13_{0.01}$ & $0.14_{0.03}$ & $0.10_{0.01}$ & 
  $0.54$ & $-0.29_{0.01}$ & $-0.26_{0.04}$ & $0.38_{0.06}$ & $0.40_{0.07}$ \\
  & Instruction-based mitigation  & $0.66$ & $-0.27_{0.01}$ & $-0.20_{0.02}$ & $0.17_{0.06}$ & $0.24_{0.05}$ & 
  $0.51$ & $-0.18_{0.02}$ & $-0.13_{0.04}$ & $0.18_{0.11}$ & $0.26_{0.10}$ \\
\midrule
\multirow{6}{*}{Qwen-70B} 
  & No mitigation  & $0.73$ & $-0.35_{0.04}$ & $-0.12_{0.02}$ & $0.09_{0.03}$ & $0.13_{0.02}$ & 
  $0.61$ & $-0.72_{0.02}$ & $-0.43_{0.02}$ & $0.39_{0.07}$ & $0.45_{0.07}$ \\
  & Random subset  & $0.71$ & $-0.24_{0.01}$ & $-0.13_{0.01}$ & $0.15_{0.03}$ & $0.13_{0.01}$ & 
  $0.55$ & $-0.65_{0.03}$ & $-0.39_{0.04}$ & $0.60_{0.09}$ & $0.57_{0.09}$ \\
  & Group-balanced & $0.71$ & $-0.15_{0.02}$ & $-0.11_{0.01}$ & $0.20_{0.02}$ & $0.17_{0.01}$ & 
  $0.59$ & $-0.56_{0.02}$ & $-0.41_{0.03}$ & $0.46_{0.04}$ & $0.43_{0.05}$ \\
  & Fair-SPD       & $0.70$ & $-0.05_{0.03}$ & $0.01_{0.02}$ & $0.36_{0.03}$ & $0.19_{0.02}$ & 
  $0.58$ & $-0.61_{0.00}$ & $-0.36_{0.04}$ & $0.39_{0.09}$ & $0.41_{0.09}$ \\
  & Correlation filtering  & $0.73$ & $-0.28_{0.05}$ & $-0.07_{0.01}$ & $0.25_{0.02}$ & $0.14_{0.01}$ & 
  $0.58$ & $-0.64_{0.01}$ & $-0.42_{0.03}$ & $0.57_{0.06}$ & $0.57_{0.06}$ \\
  & Instruction-based mitigation  & $0.71$ & $-0.15_{0.04}$ & $-0.15_{0.01}$ & $0.06_{0.03}$ & $0.11_{0.03}$ & 
  $0.39$ & $-0.06_{0.01}$ & $-0.05_{0.03}$ & $0.02_{0.02}$ & $0.03_{0.02}$ \\
\bottomrule
\end{tabular}
\vspace{-3mm}
\end{table*}

\begin{figure}[th!]
  \centering
  \includegraphics[width=1.0\linewidth]{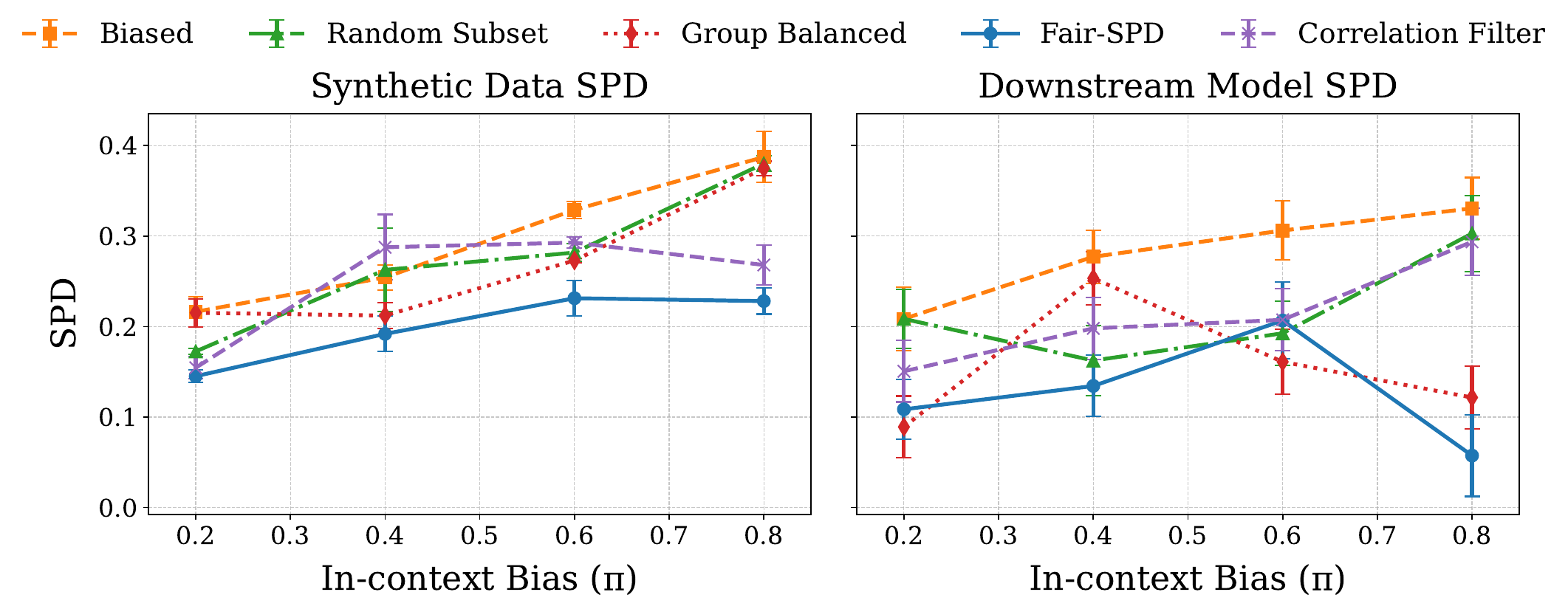}
    \caption{\textbf{Synthetic and downstream fairness for each preprocessing approach.} (\textit{Granite-8B, Compas}) $\text{SPD}$ of the synthetic dataset generated by the LLM (left) and downstream classifier predictions on the real test dataset (right) for increasing bias rates across preprocessing mitigations.}
      \label{fig:downstream_mitigation_fairness}
\end{figure}

We evaluate the mitigation strategies following the experimental setup described in Section \ref{sec:design}. Before generating synthetic examples, we apply each of the strategies to the in-context examples as a pre-processing step or we specifically add a fairness instruction in the prompt. Once the generation is completed, we evaluate the resulting dataset. In Figure \ref{fig:downstream_mitigation_fairness}, we provide a comparison across in-context preprocessing techniques for Compas using Granite-8B for increasing levels of in-context bias. Here, the target group is African-American, and the target variable is recidivism. As $\pi$ increases, the synthetic data disparity $\text{SPD}_S$ increases under the biased baseline. Random subset selection can partially attenuate $\text{SPD}_S$, but typically remains closer to the biased scenario. Frequency balancing further reduces $\text{SPD}_S$ in many settings by correcting representation imbalances introduced by oversampling adversarial examples. Fair-SPD keeps $\text{SPD}_S$ comparatively stable across $\pi$ and achieves the lowest downstream disparity $\text{SPD}_D$ among the evaluated approaches. However, while enforcing demographic parity in the prompt attenuates marginal bias transfer, $\text{SPD}_D$ can still increase with prompt bias $\pi$, suggesting that injected bias may propagate through higher-order correlations not captured by marginal parity alone.

Complementing this analysis, Table \ref{tab:mitig} summarizes mitigation results on Adult and Compas across three LLMs, now including instruction-based mitigation. Overall, Fair-SPD consistently reduces downstream disparity $\text{SPD}_D$ relative to the unmitigated baseline, while preserving comparable F1 in most settings. Instruction-based mitigation also reduces disparity in several configurations, but its effectiveness is model- and dataset-dependent and may introduce utility degradation. While improvements in $\text{SPD}_S$ are frequent, there are cases where $\text{SPD}_S$ does not decrease even when $\text{SPD}_D$ improves, indicating a decoupling between synthetic marginal parity and downstream parity. Moreover, fairness gains measured by SPD do not guarantee improvements in EO/EOD, which can remain large or even worsen in some configurations. Taken together, these results indicate that fairness-constrained prompt selection is a stronger baseline than randomization or frequency balancing for limiting downstream bias propagation, but it does not fully eliminate adversarial bias injection across all models and metrics.

\section{Discussion}
\label{sec:discussion}

While recent work has established Large Language Models (LLMs) as powerful few-shot tabular data generators, our results show that LLM-based tabular data generation is highly sensitive to the statistical structure of the in-context examples. The mechanisms that drive high-fidelity generation are the very same that facilitate bias propagation. Interestingly, while a larger in-context size $k$ allows the model to better capture the in-context distribution, it simultaneously increases the model's sensitivity to demographic skews. This creates a trade-off where increasing the context window to improve synthetic data utility inadvertently increases the risk of replicating historical injustices or adversarial injections. Our anchor construction $\tilde{\mathcal{D}}_0$ provides a controlled reference for the model's prior behavior, allowing us to attribute disparities at $\pi = 0$ to intrinsic model bias and measure extrinsic bias as the deviation induced by controlled prompt perturbations, although making a quantitative decomposition of intrinsic and extrinsic contributions to downstream fairness remains an important direction for future work.

Moreover, we demonstrated that in-context bias propagation can be exploited by an adversary to maliciously inject biases for a targeted subgroup, even when the target lies at the intersection of multiple variables. We introduced adversarial in-context bias injection as a new vulnerability scheme in LLM-based generation pipelines, showing that an adversary controlling a small subset of in-context examples can influence fairness in synthetic data. Moreover, we demonstrate that these synthetic biases are captured by downstream models, leading to discriminatory predictions on real test data while maintaining high utility metrics. This decoupling is consistent with fairness-preserving frameworks that distinguish data-level distributional bias from model-level fairness ~\citep{uddin2026fairness}. Importantly, the studied vulnerability persists across various downstream architectures, including traditional classifiers and modern tabular foundation models like TabPFN. Future work should compare in-context bias injection with parameter-efficient fine-tuning, such as LoRA, on similarly poisoned data to better understand how different adaptation mechanisms propagate bias.

Finally, we evaluated prompt-side defenses based on demonstration preprocessing and fairness-aware instructions. We observe that balancing the frequency count can mitigate biases by diluting the influence of an adversary injecting examples from a specific subgroup. Similarly, enforcing a fair Statistical Parity Difference (SPD) within in-context examples attenuates adversarial bias injection. Fairness-aware instructions also reduce bias in several settings, but their effectiveness is model- and dataset-dependent and may involve utility degradation. However, biases often persist due to the model's intrinsic priors and the leakage of higher-order correlations that marginal constraints fail to capture. We leave as future work the study of in-processing and post-processing techniques to address adversarial in-context bias without compromising the utility of the generated data.

\section{Conclusion}
\label{sec:conclusion}

In this paper, we systematically characterized how statistical biases within in-context examples propagate to LLM-generated tabular data. We demonstrated that as the number of in-context examples increases, the model increasingly aligns with the statistical properties of the prompt, thereby replicating univariate, conditional, and intersectional biases. Building on this, we introduced a novel adversarial framework where an adversary can inject feature-aligned examples to manipulate the output distribution for a targeted subgroup. Our results show that this manipulation can compromise synthetic-data and downstream fairness while largely preserving utility, making utility-based checks alone insufficient to reveal the attack. Structural and anomaly-based diagnostics can detect manipulation in some settings, although the evaluated anomaly filter also removes many clean examples. Finally, prompt-side defenses can attenuate these effects, but their effectiveness and utility trade-offs vary across settings. Together, these findings expose an important vulnerability in LLM-based data generation pipelines and highlight the need for more robust safeguards.

\section*{Acknowledgements}

This work has been partially financed by grant agreement EU-HORIZON GA.101095717 and by grant agreement EU-HORIZON MSCA GA.101086248. Also, it has been partially financed by Generalitat de Catalunya (AGAUR) under grant agreement 2025 SGR 651, by Severo Ochoa Center of Excellence CEX-2021-001148-S-20-3, and by the Spanish Ministry of Science (MICINN), the Research State Agency (AEI) and European Regional Development Funds (ERDF/FEDER) under grant agreement PID2021-126248OB-I00, MCIN/AEI/10.13039/ 501100011033/ FEDER, UE. Kieran Fraser and Anisa Halimi were partly supported by the Innovative Health Initiative Joint Undertaking (IHI JU) under grant agreement No. 101172997 – SEARCH.

\bibliographystyle{elsarticle-num} 
\bibliography{main}

@article{wang2024harmonic,
  title={HARMONIC: Harnessing LLMs for tabular data synthesis and privacy protection},
  author={Wang, Yuxin and Feng, Duanyu and Dai, Yongfu and Chen, Zhengyu and Huang, Jimin and Ananiadou, Sophia and Xie, Qianqian and Wang, Hao},
  journal={arXiv preprint arXiv:2408.02927},
  year={2024}
}

@article{seedat2023curated,
  title={Curated llm: Synergy of llms and data curation for tabular augmentation in ultra low-data regimes},
  author={Seedat, Nabeel and Huynh, Nicolas and van Breugel, Boris and van der Schaar, Mihaela},
  year={2023}
}

@article{gretton2012kernel,
  title={A kernel two-sample test},
  author={Gretton, Arthur and Borgwardt, Karsten M and Rasch, Malte J and Sch{\"o}lkopf, Bernhard and Smola, Alexander},
  journal={The journal of machine learning research},
  volume={13},
  number={1},
  pages={723--773},
  year={2012},
  publisher={JMLR. org}
}

@InProceedings{pmlr-v139-zhao21c,
  title = 	 {Calibrate Before Use: Improving Few-shot Performance of Language Models},
  author =       {Zhao, Zihao and Wallace, Eric and Feng, Shi and Klein, Dan and Singh, Sameer},
  booktitle = 	 {Proceedings of the 38th International Conference on Machine Learning},
  pages = 	 {12697--12706},
  year = 	 {2021},
  editor = 	 {Meila, Marina and Zhang, Tong},
  volume = 	 {139},
  series = 	 {Proceedings of Machine Learning Research},
  month = 	 {18--24 Jul},
  publisher =    {PMLR},
  url = 	 {https://proceedings.mlr.press/v139/zhao21c.html},
}

@InProceedings{pmlr-v235-lin24l,
  title = 	 {Dual Operating Modes of In-Context Learning},
  author =       {Lin, Ziqian and Lee, Kangwook},
  booktitle = 	 {Proceedings of the 41st International Conference on Machine Learning},
  pages = 	 {30135--30188},
  year = 	 {2024},
  editor = 	 {Salakhutdinov, Ruslan and Kolter, Zico and Heller, Katherine and Weller, Adrian and Oliver, Nuria and Scarlett, Jonathan and Berkenkamp, Felix},
  volume = 	 {235},
  series = 	 {Proceedings of Machine Learning Research},
  month = 	 {21--27 Jul},
  publisher =    {PMLR},
  url = 	 {https://proceedings.mlr.press/v235/lin24l.html},
}

@article{uddin2026fairness,
  title={Fairness-preserving framework for machine learning: data bias quantification, model evaluation, and robustness across multiple datasets},
  author={Uddin, Shahadat and Lu, Haohui and Hajati, Farshid},
  journal={AI and Ethics},
  volume={6},
  number={1},
  pages={25},
  year={2026},
  publisher={Springer}
}

@article{uddin2024novel,
  title={A novel approach for assessing fairness in deployed machine learning algorithms},
  author={Uddin, Shahadat and Lu, Haohui and Rahman, Ashfaqur and Gao, Junbin},
  journal={Scientific Reports},
  volume={14},
  number={1},
  pages={17753},
  year={2024},
  publisher={Nature Publishing Group UK London}
}

@misc{moro2014bank,
  title={Bank Marketing [Dataset]. UCI Machine Learning Repository},
  author={Moro, S and Rita, P and Cortez, P},
  year={2014}
}

@article{li2026cross,
  title={Cross-Model Adjudication for Bias Mitigation in Large Language Models},
  author={Li, Xiangmei and Li, Changwei and Liu, Wenjie and Long, Yan and Huang, Yifan and Liu, Yi},
  journal={IEEE Journal of Selected Topics in Signal Processing},
  year={2026},
  publisher={IEEE}
}

@article{lu2026bias,
  title={Bias Mitigation in Large Language Models for Tabular Data Classification},
  author={Lu, Haohui and Shao, Zhiqi and Gao, Junbin and Uddin, Shahadat},
  journal={Machine Learning},
  volume={115},
  number={4},
  pages={79},
  year={2026},
  publisher={Springer}
}

@article{xie2021explanation,
  title={An explanation of in-context learning as implicit bayesian inference},
  author={Xie, Sang Michael and Raghunathan, Aditi and Liang, Percy and Ma, Tengyu},
  journal={arXiv preprint arXiv:2111.02080},
  year={2021}
}

@article{chawla2002smote,
  title={SMOTE: synthetic minority over-sampling technique},
  author={Chawla, Nitesh V and Bowyer, Kevin W and Hall, Lawrence O and Kegelmeyer, W Philip},
  journal={Journal of artificial intelligence research},
  volume={16},
  pages={321--357},
  year={2002}
}

@inproceedings{nangia2020crows,
  title={CrowS-pairs: A challenge dataset for measuring social biases in masked language models},
  author={Nangia, Nikita and Vania, Clara and Bhalerao, Rasika and Bowman, Samuel},
  booktitle={Proceedings of the 2020 conference on empirical methods in natural language processing (EMNLP)},
  pages={1953--1967},
  year={2020}
}

@article{xue2024badfair,
  title={BadFair: Backdoored Fairness Attacks with Group-conditioned Triggers},
  author={Xue, Jiaqi and Lou, Qian and Zheng, Mengxin},
  journal={arXiv preprint arXiv:2410.17492},
  year={2024}
}

@misc{TabPFN-2.5,
      title={TabPFN-2.5: Advancing the State of the Art in Tabular Foundation Models},
      author={Léo Grinsztajn and Klemens Flöge and Oscar Key and Felix Birkel and Brendan Roof and Phil Jund and Benjamin Jäger and Adrian Hayler and Dominik Safaric and Simone Alessi, Felix Jablonski and Mihir Manium and Rosen Yu and Anurag Garg and Jake Robertson and Shi Bin (Liam) Hoo and Vladyslav Moroshan and Magnus Bühler and Lennart Purucker and Clara Cornu and Lilly Charlotte Wehrhahn and Alessandro Bonetto and Sauraj Gambhir and Noah Hollmann and Frank Hutter},
      year={2025}
}

@article{xue2024badrag,
  title={Badrag: Identifying vulnerabilities in retrieval augmented generation of large language models},
  author={Xue, Jiaqi and Zheng, Mengxin and Hu, Yebowen and Liu, Fei and Chen, Xun and Lou, Qian},
  journal={arXiv preprint arXiv:2406.00083},
  year={2024}
}

@article{hu2024strategic,
  title={Strategic demonstration selection for improved fairness in LLM in-context learning},
  author={Hu, Jingyu and Liu, Weiru and Du, Mengnan},
  journal={arXiv preprint arXiv:2408.09757},
  year={2024}
}

@article{mehrabi2021survey,
  title={A survey on bias and fairness in machine learning},
  author={Mehrabi, Ninareh and Morstatter, Fred and Saxena, Nripsuta and Lerman, Kristina and Galstyan, Aram},
  journal={ACM computing surveys (CSUR)},
  volume={54},
  number={6},
  pages={1--35},
  year={2021},
  publisher={ACM New York, NY, USA}
}

@article{kenfack2025towards,
  title={Towards Fair In-Context Learning with Tabular Foundation Models},
  author={Kenfack, Patrik and Kahou, Samira Ebrahimi and A{\"\i}vodji, Ulrich},
  journal={arXiv preprint arXiv:2505.09503},
  year={2025}
}

@article{smith2022m,
  title={" I'm sorry to hear that": Finding New Biases in Language Models with a Holistic Descriptor Dataset},
  author={Smith, Eric Michael and Hall, Melissa and Kambadur, Melanie and Presani, Eleonora and Williams, Adina},
  journal={arXiv preprint arXiv:2205.09209},
  year={2022}
}

@inproceedings{dhamala2021bold,
  title={Bold: Dataset and metrics for measuring biases in open-ended language generation},
  author={Dhamala, Jwala and Sun, Tony and Kumar, Varun and Krishna, Satyapriya and Pruksachatkun, Yada and Chang, Kai-Wei and Gupta, Rahul},
  booktitle={Proceedings of the 2021 ACM conference on fairness, accountability, and transparency},
  pages={862--872},
  year={2021}
}

@article{schick2021self,
  title={Self-diagnosis and self-debiasing: A proposal for reducing corpus-based bias in nlp},
  author={Schick, Timo and Udupa, Sahana and Sch{\"u}tze, Hinrich},
  journal={Transactions of the Association for Computational Linguistics},
  volume={9},
  pages={1408--1424},
  year={2021},
  publisher={MIT Press One Rogers Street, Cambridge, MA 02142-1209, USA journals-info~…}
}

@inproceedings{liu2008isolation,
  title={Isolation forest},
  author={Liu, Fei Tony and Ting, Kai Ming and Zhou, Zhi-Hua},
  booktitle={2008 eighth ieee international conference on data mining},
  pages={413--422},
  year={2008},
  organization={Ieee}
}

@inproceedings{van2022poisoning,
  title={Poisoning attacks on fair machine learning},
  author={Van, Minh-Hao and Du, Wei and Wu, Xintao and Lu, Aidong},
  booktitle={International Conference on Database Systems for Advanced Applications},
  pages={370--386},
  year={2022},
  organization={Springer}
}

@inproceedings{nadeem2021stereoset,
  title={StereoSet: Measuring stereotypical bias in pretrained language models},
  author={Nadeem, Moin and Bethke, Anna and Reddy, Siva},
  booktitle={Proceedings of the 59th annual meeting of the association for computational linguistics and the 11th international joint conference on natural language processing (volume 1: long papers)},
  pages={5356--5371},
  year={2021}
}

@inproceedings{solans2020poisoning,
  title={Poisoning attacks on algorithmic fairness},
  author={Solans, David and Biggio, Battista and Castillo, Carlos},
  booktitle={Joint European Conference on Machine Learning and Knowledge Discovery in Databases},
  pages={162--177},
  year={2020},
  organization={Springer}
}

@article{zhang2023mixed,
  title={Mixed-type tabular data synthesis with score-based diffusion in latent space},
  author={Zhang, Hengrui and Zhang, Jiani and Srinivasan, Balasubramaniam and Shen, Zhengyuan and Qin, Xiao and Faloutsos, Christos and Rangwala, Huzefa and Karypis, George},
  journal={arXiv preprint arXiv:2310.09656},
  year={2023}
}

@inproceedings{kotelnikov2023tabddpm,
  title={Tabddpm: Modelling tabular data with diffusion models},
  author={Kotelnikov, Akim and Baranchuk, Dmitry and Rubachev, Ivan and Babenko, Artem},
  booktitle={International conference on machine learning},
  pages={17564--17579},
  year={2023},
  organization={PMLR}
}

@inproceedings{tazwar2024tab,
  title={Tab-VAE: A Novel VAE for Generating Synthetic Tabular Data.},
  author={Tazwar, Syed Mahir and Knobbout, Max and Quesada, Enrique Hortal and Popa, Mirela},
  booktitle={ICPRAM},
  pages={17--26},
  year={2024}
}

@inproceedings{zhao2021ctab,
  title={Ctab-gan: Effective table data synthesizing},
  author={Zhao, Zilong and Kunar, Aditya and Birke, Robert and Chen, Lydia Y},
  booktitle={Asian conference on machine learning},
  pages={97--112},
  year={2021},
  organization={PMLR}
}

@article{xu2019modeling,
  title={Modeling tabular data using conditional gan},
  author={Xu, Lei and Skoularidou, Maria and Cuesta-Infante, Alfredo and Veeramachaneni, Kalyan},
  journal={Advances in neural information processing systems},
  volume={32},
  year={2019}
}

@article{tanaka2019data,
  title={Data augmentation using GANs},
  author={Tanaka, Fabio Henrique Kiyoiti Dos Santos and Aranha, Claus},
  journal={arXiv preprint arXiv:1904.09135},
  year={2019}
}

@article{panwar2023context,
  title={In-context learning through the bayesian prism},
  author={Panwar, Madhur and Ahuja, Kabir and Goyal, Navin},
  journal={arXiv preprint arXiv:2306.04891},
  year={2023}
}

@inproceedings{dinan2020queens,
  title={Queens are powerful too: Mitigating gender bias in dialogue generation},
  author={Dinan, Emily and Fan, Angela and Williams, Adina and Urbanek, Jack and Kiela, Douwe and Weston, Jason},
  booktitle={Proceedings of the 2020 Conference on Empirical Methods in Natural Language Processing (EMNLP)},
  pages={8173--8188},
  year={2020}
}

@article{hall2022systematic,
  title={A systematic study of bias amplification},
  author={Hall, Melissa and van der Maaten, Laurens and Gustafson, Laura and Jones, Maxwell and Adcock, Aaron},
  journal={arXiv preprint arXiv:2201.11706},
  year={2022}
}

@article{leino2018feature,
  title={Feature-wise bias amplification},
  author={Leino, Klas and Black, Emily and Fredrikson, Matt and Sen, Shayak and Datta, Anupam},
  journal={arXiv preprint arXiv:1812.08999},
  year={2018}
}

@article{zhou2024unibias,
  title={Unibias: Unveiling and mitigating llm bias through internal attention and ffn manipulation},
  author={Zhou, Hanzhang and Feng, Zijian and Zhu, Zixiao and Qian, Junlang and Mao, Kezhi},
  journal={arXiv preprint arXiv:2405.20612},
  year={2024}
}

@article{ma2023fairness,
  title={Fairness-guided few-shot prompting for large language models},
  author={Ma, Huan and Zhang, Changqing and Bian, Yatao and Liu, Lemao and Zhang, Zhirui and Zhao, Peilin and Zhang, Shu and Fu, Huazhu and Hu, Qinghua and Wu, Bingzhe},
  journal={Advances in Neural Information Processing Systems},
  volume={36},
  pages={43136--43155},
  year={2023}
}

@article{tornqvist2024text,
  title={A text-to-tabular approach to generate synthetic patient data using LLMs},
  author={Tornqvist, Margaux and Zucker, Jean-Daniel and Fauvel, Tristan and Lambert, Nicolas and Berthelot, Mathilde and Movschin, Antoine},
  journal={arXiv preprint arXiv:2412.05153},
  year={2024}
}

@misc{angwin2016machine,
  author = {Angwin, Julia and Larson, Jeff and Kirchner, Lauren and Mattu, Surya},
  title = {Machine Bias},
  howpublished = {https://www.propublica.org/article/machine-bias-risk-assessments-in-criminal-sentencing},
  note = {ProPublica},
  year = {2016},
  month = may
}

@misc{asuncion2007uci,
  author = {Asuncion, Arthur and Newman, David},
  title = {UCI Machine Learning Repository},
  year  = {2007},
  howpublished = {http://archive.ics.uci.edu/ml},
  note = {University of California, Irvine, School of Information and Computer Sciences}
}

@article{kim2024epic,
  title={EPIC: Effective Prompting for Imbalanced-Class Data Synthesis in Tabular Data Classification via Large Language Models},
  author={Kim, Jinhee and Kim, Taesung and Choo, Jaegul},
  journal={Advances in Neural Information Processing Systems},
  volume={37},
  pages={31504--31542},
  year={2024}
}

@article{borisov2022language,
  title={Language models are realistic tabular data generators},
  author={Borisov, Vadim and Se{\ss}ler, Kathrin and Leemann, Tobias and Pawelczyk, Martin and Kasneci, Gjergji},
  journal={arXiv preprint arXiv:2210.06280},
  year={2022}
}

@article{cherepanova2024improving,
  title={Improving llm group fairness on tabular data via in-context learning},
  author={Cherepanova, Valeriia and Lee, Chia-Jung and Akpinar, Nil-Jana and Fogliato, Riccardo and Bertran, Martin Andres and Kearns, Michael and Zou, James},
  journal={arXiv preprint arXiv:2412.04642},
  year={2024}
}

@article{liu2023confronting,
  title={Confronting LLMs with traditional ML: Rethinking the fairness of large language models in tabular classifications},
  author={Liu, Yanchen and Gautam, Srishti and Ma, Jiaqi and Lakkaraju, Himabindu},
  journal={arXiv preprint arXiv:2310.14607},
  year={2023}
}

@inproceedings{baracaldo2022benchmarking,
  title={Benchmarking the effect of poisoning defenses on the security and bias of the final model},
  author={Baracaldo, Nathalie and Eykholt, Kevin and Ahmed, Farhan and Zhou, Yi and Priya, Shriti and Lee, Taesung and Kadhe, Swanand and Tan, Yusong and Polavaram, Sridevi and Suggs, Sterling and others},
  booktitle={Workshop on Trustworthy and Socially Responsible Machine Learning, NeurIPS 2022},
  year={2022}
}

\newpage
\appendix
\onecolumn

\section*{Appendix}
This appendix complements the main manuscript with additional experimental results and implementation details:
\begin{itemize}
    \item \ref{app:datasets} details the datasets.    
    \item \ref{app:a} presents extended results for all model families, validating the findings from Sections~\ref{sec:bias-model} and~\ref{sec:adv-bias}.
    \item \ref{app:c} provides complete prompt templates and demonstrates that bias propagation persists without explicit mirroring instructions.
\end{itemize}

\section{Datasets} \label{app:datasets}

We evaluate five tabular classification datasets covering socioeconomic, criminal-justice, healthcare, and financial domains. Adult, Compas, Diabetes, and Bank Marketing are binary classification datasets, whereas Thyroid is formulated as a multiclass classification task. The protected-group definitions reported below are used consistently for prompt construction, synthetic-data fairness evaluation, and downstream fairness evaluation. Attributes and tasks are detailed in Table \ref{tab:dataset_summary}.

\begin{table*}[t]
\scriptsize
\setlength{\tabcolsep}{3pt}
\renewcommand{\arraystretch}{1.08}
\centering
\caption{Summary of the datasets used in the evaluation. The protected group is the subgroup targeted in the adversarial experiments.}
\label{tab:dataset_summary}
\begin{tabular}{p{1.25cm} p{2.25cm} p{6.2cm} p{2.05cm} p{3.15cm}}
\toprule
\textbf{Dataset} & \textbf{Task} & \textbf{Attributes} & \textbf{Target} & \textbf{Protected group} \\
\midrule

Adult & Income prediction &
age, workclass, fnlwgt, education, educational-num, marital-status, occupation, relationship, race, gender, capital-gain, capital-loss, hours-per-week, and native-country. &
income: income $>50$K &
Female, married-civ-spouse, United-States. \\

COMPAS & Two-year recidivism prediction &
sex, age, juv\_fel\_count, juv\_misd\_count, juv\_other\_count, priors\_count, age-category indicators, race indicators, and charge-degree indicators. The prepared dataset also contains id as an identifier. &
y: recidivism &
African-American \\

Diabetes & Diabetes diagnosis &
Pregnancies, Glucose, BloodPressure, SkinThickness, Insulin, BMI, DiabetesPedigreeFunction, and Age. &
Outcome: diabetes diagnosis &
Age between 21 and 30 \\

Thyroid & Thyroid-risk classification &
Age, Gender, Pregnancy, Family\_History\_of\_Thyroid, Goiter, Fatigue, Weight\_Change, Hair\_Loss, Heart\_Rate\_Changes, Sensitivity\_to\_Cold\_or\_Heat, Increased\_Sweating, Muscle\_Weakness, Constipation\_or\_More\_Bowel\_Movements, Depression\_or\_Anxiety, Difficulty\_Concentrating\_or\_Memory\_Problems, and Dry\_or\_Itchy\_Skin. &
Thyroid\_Risk\_Level: no, moderate, or high risk &
Age between 20 and 35 \\

Bank & Term-deposit subscription prediction &
age, job, marital, education, default, balance, housing, loan, contact, day, month, duration, campaign, pdays, previous, and poutcome. &
y: subscribed to term deposit &
Age between 18 and 30 \\

\bottomrule
\end{tabular}
\end{table*}

\paragraph{Adult} The Adult dataset contains census records and is commonly used to predict whether an individual's annual income exceeds \$50,000. Its attributes combine numerical variables, such as age, capital gain, capital loss, and weekly working hours, with categorical variables describing education, occupation, marital status, relationship, race, gender, and country of origin. The favorable target outcome is income above \$50,000. Gender is used as the main protected attribute in the conditional-bias experiments, with Female defined as the targeted subgroup. We additionally use gender and race to construct the four intersectional groups studied in Section~\ref{sec:bias-model}. In the adversarial experiment, the targeted Adult subgroup combines Female with \textit{married-civ-spouse} and \textit{United-States}, while the remaining attributes are assigned plausible employment and demographic values. 

\paragraph{Compas} The Compas dataset contains criminal-justice records used to study two-year recidivism prediction. Its features include demographic information, such as age, gender, and race, together with criminal-history variables, including the number of prior convictions, juvenile offenses, and the degree of the current charge. The positive target denotes recidivism. Race is the protected attribute, and African-American individuals form the targeted subgroup. This dataset is particularly relevant to our study because disparities in protected-group outcome rates may be propagated into downstream risk assessments. The adversarial records use plausible ranges for age and prior convictions rather than extreme values, allowing us to study targeted bias injection under feature-aligned conditions. 

\paragraph{Diabetes} The Diabetes dataset is a numerical healthcare dataset for binary diabetes prediction. It contains diagnostic measurements such as pregnancies, glucose concentration, blood pressure, skin thickness, insulin, body-mass index, diabetes pedigree function, and age. The positive target indicates a diabetes diagnosis. Since the dataset does not contain a categorical protected attribute suitable for the attack considered in this work, we define the protected subgroup using age, with individuals between 21 and 30 years old forming the targeted group. The adversarial profile uses clinically plausible ranges for glucose, BMI, insulin, and diabetes pedigree values. 

\paragraph{Thyroid} The Thyroid dataset contains clinical and demographic variables used to predict thyroid-risk categories. Unlike the other datasets, its target is multiclass, with the high-risk category used as the favorable or targeted outcome when computing group disparities. The attributes include age, gender, and thyroid-related symptoms and risk factors such as goiter, family history, and fatigue. We define individuals between 20 and 35 years old as the protected subgroup. The adversarial examples associate this subgroup with the high-risk target while sampling binary clinical attributes from valid values.

\paragraph{Bank Marketing}
The Bank Marketing dataset contains demographic, financial, and campaign-related variables used to predict whether a customer subscribes to a term deposit. Its target is binary, with subscription (\textit{yes}) treated as the favorable outcome. The attributes include age, occupation, marital status, education, account balance, credit default, housing and personal loans, contact method, and current and previous campaign information. We define customers younger than 30 years as the protected subgroup. The adversarial examples associate this subgroup with the positive subscription outcome while assigning plausible values to the remaining financial and campaign attributes.

\section{Additional Experimental Results} \label{app:a}
This section extends the empirical evaluation from Sections~\ref{sec:bias-model} and~\ref{sec:adv-bias} to three additional model families: Granite-8B, Mistral-7B, and Qwen-70B. These results demonstrate that bias propagation is a phenomenon observed across model architectures, scales, and context window sizes, validating the generality of our main findings.

\subsection{Marginal Bias Propagation}

We first examine marginal bias propagation, systematically increasing the representation of the protected group in the in-context examples via the parameter $\pi$. Here, we analyze the relationship between the distributional drift of the prompt (relative to the anchor $\mathcal{D}_0$) and the resulting drift in the generated distribution. Furthermore, we evaluate how the protected group probability in the generated samples aligns with the probability in the in-context examples. For these experiments on the Adult dataset, the protected group is defined as ``Black''.

\begin{figure}[h]
  \centering
  \includegraphics[width=0.32\columnwidth]{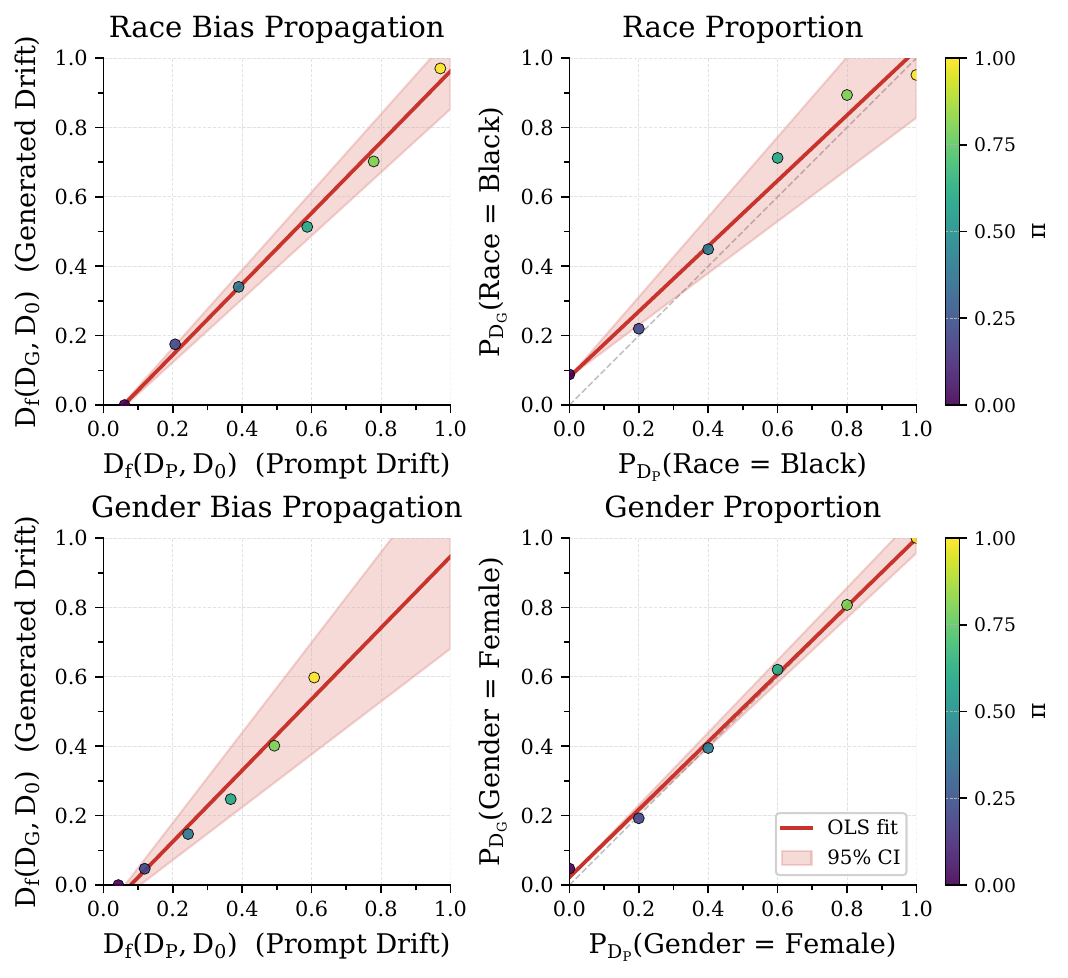}
  \includegraphics[width=0.32\columnwidth]{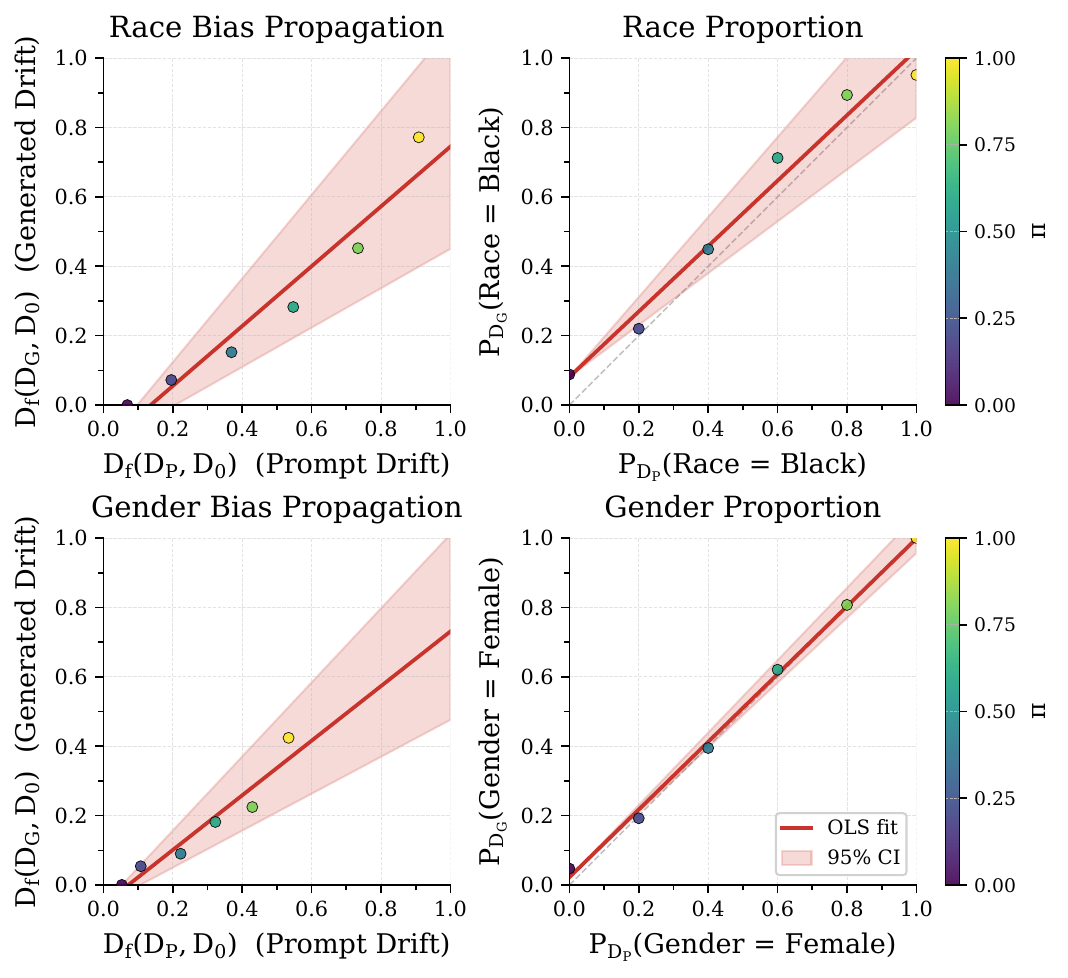}
  \includegraphics[width=0.32\columnwidth]{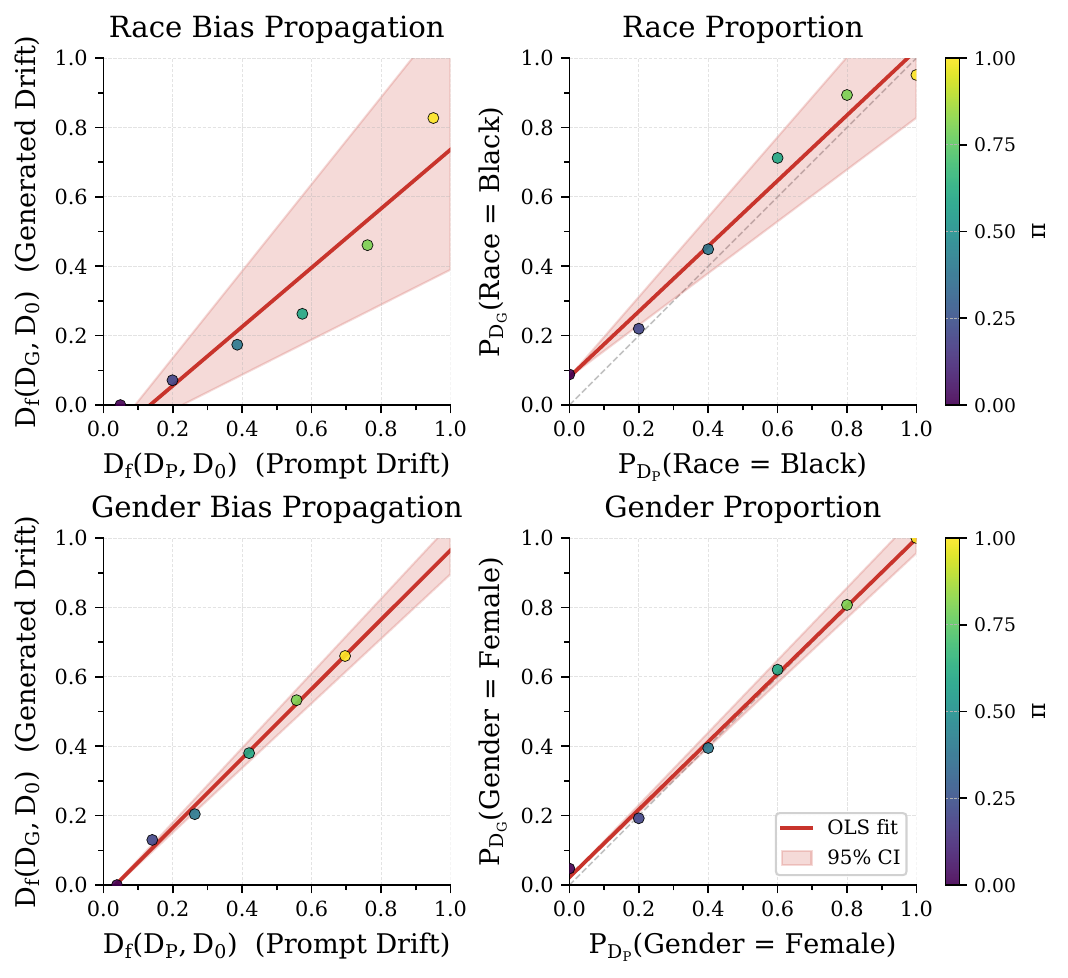}
    \caption{\textbf{Bias propagation across different models.} (\textit{Left}: Granite-8B; \textit{Center}: Mistral-7B; \textit{Right}: Qwen-70B). In all cases, as the bias in the prompt increases, the bias in the generated output scales up linearly.}
  \label{fig:univariate_drift_composite}
\end{figure}

Figure \ref{fig:univariate_drift_composite} illustrates the distributional drift across the three models. Consistent with the main text, we observe a relationship between the divergence in the prompt ($D_f(\mathcal{D}_P, \mathcal{D}_0)$) and the divergence in the generated data ($D_f(\mathcal{D}_G, \mathcal{D}_0)$).

Qwen-70B exhibits near-perfect replication of prompt statistics, behaving analogously to the Mixtral-8x7B model analyzed in the main text, with a low-variance regression line for gender bias propagation. However, for race bias propagation, the relationship deviates from strict linearity. In contrast, as shown in the Granite-8B and Mistral-7B plots (Figure \ref{fig:univariate_drift_composite}, Left and Center), while the general linear trend persists, these smaller models exhibit higher variance for gender bias, whereas Granite-8B specifically displays greater linearity for race bias. This suggests that the targeted attribute interacts with the model prior, influencing the magnitude of the drift. However, the right plot shows a strictly linear relationship between the race proportion in the prompt and the generated data across all models, which indicates that even if distributional drift fluctuates, the generated data faithfully preserves the marginal properties of the prompt.

\paragraph{Sensitivity to Context Size}

\begin{figure}[t]
  \centering
  \includegraphics[width=0.85\linewidth]{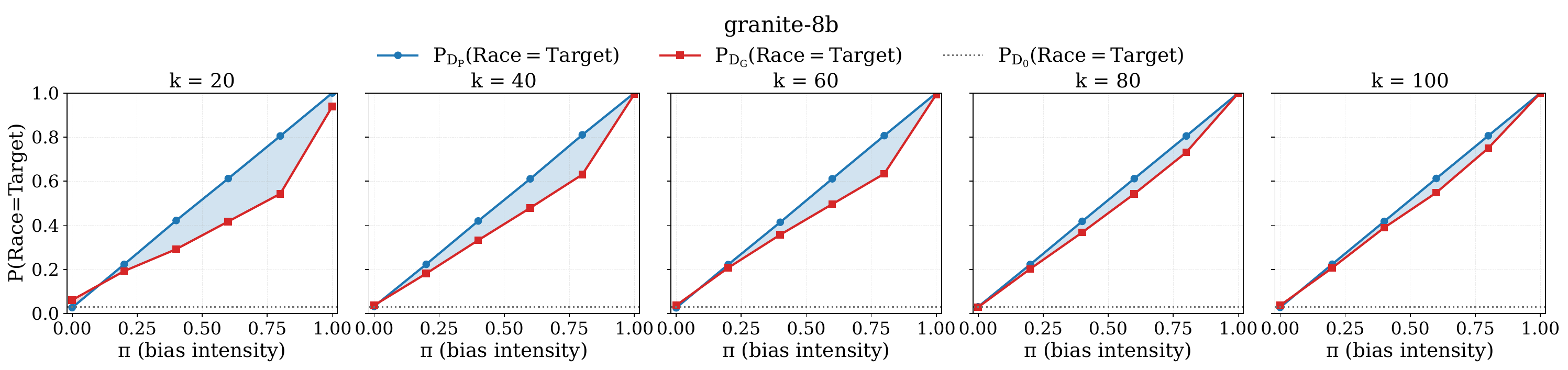}
  \includegraphics[width=0.85\linewidth]{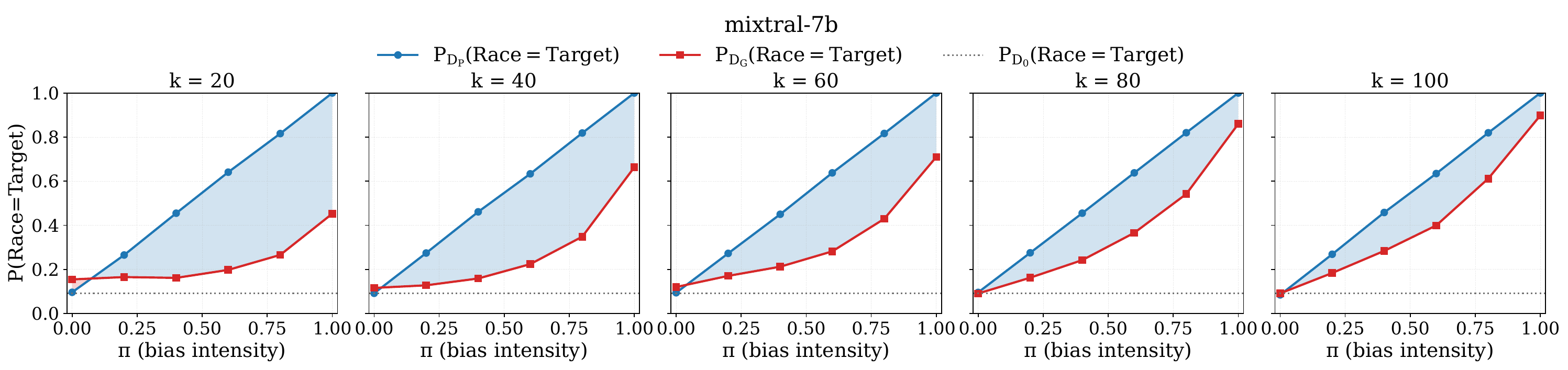}
  \includegraphics[width=0.85\linewidth]{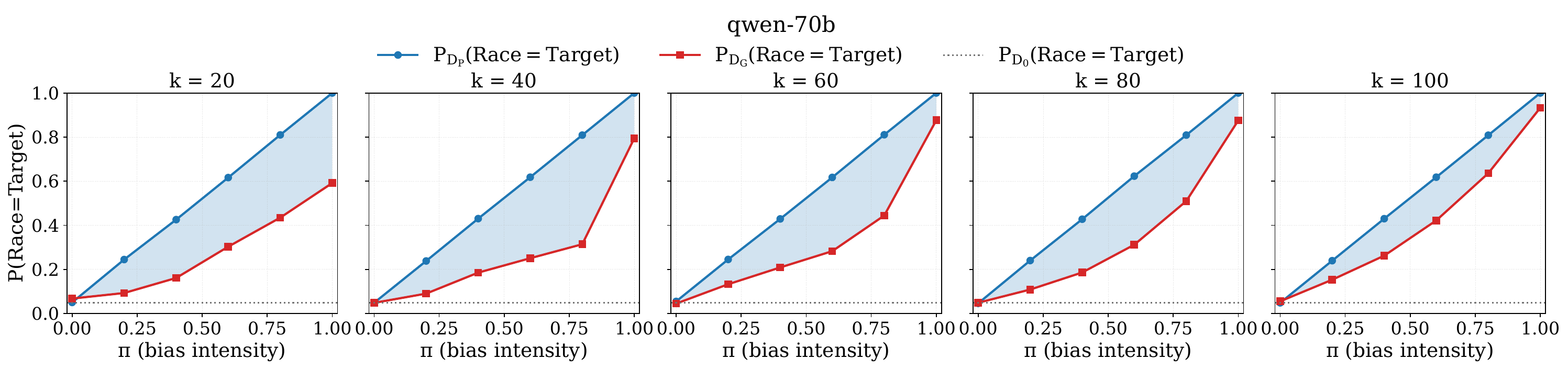}

    \caption{\textbf{Bias propagation strength vs. context size $k$.} (\textit{Top}: Granite-8B; \textit{Middle}: Mistral-7B; \textit{Bottom}: Qwen-70B). The red curves ($D_G$) align closer to the blue curves ($D_P$) as $k$ increases for all models.}
    \label{fig:beta_composite}
\end{figure}

Figure \ref{fig:beta_composite} illustrates the relationship between the protected group probability in the prompt and the generated data as a function of context size $k$. We observe across all models that as $k$ increases, the generated distribution converges toward the in-context statistics. This confirms that larger context windows improve the LLM fidelity to the prompt, thereby increasing the replication of demographic skews.

\subsection{Conditional Bias Propagation}

We further validate the capability of models to isolate bias within a specific subgroup and assess the extent to which this targeted bias leaks into non-targeted demographics. Using the Adult dataset, we systematically bias the Income probability for Female examples while maintaining a fixed probability of 0.5 for Male examples. However, we observe distinct behaviors regarding the non-targeted Male group across different model families. Qwen-70B demonstrates an advanced capability to disentangle conditional distributions $P(Y|X, A)$ from the context, exhibiting minimal leakage to the male subgroup despite the shifting female statistics. 

In contrast, smaller models like Mistral-7B display significant bias propagation into the non-targeted group, compared to Mixtral-8x7B within the same family, as observed in Figure \ref{fig:icl_samples}. This suggests that higher-capacity models achieve greater precision in bias propagation, compared to smaller models within the same architecture and pretraining. Paradoxically, this precision makes the feature-aligned attack described in Section \ref{sec:adv-bias} more effective on advanced models.

\begin{figure}[h]
  \centering
  \includegraphics[width=0.49\linewidth]{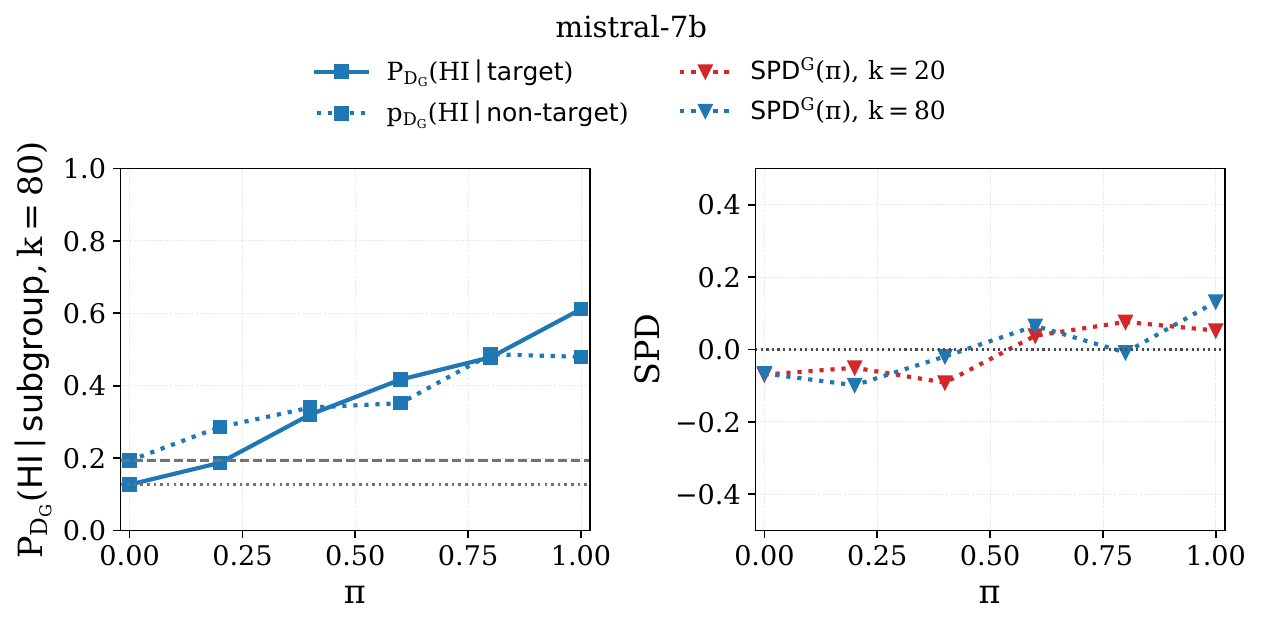}
  \includegraphics[width=0.49\linewidth]{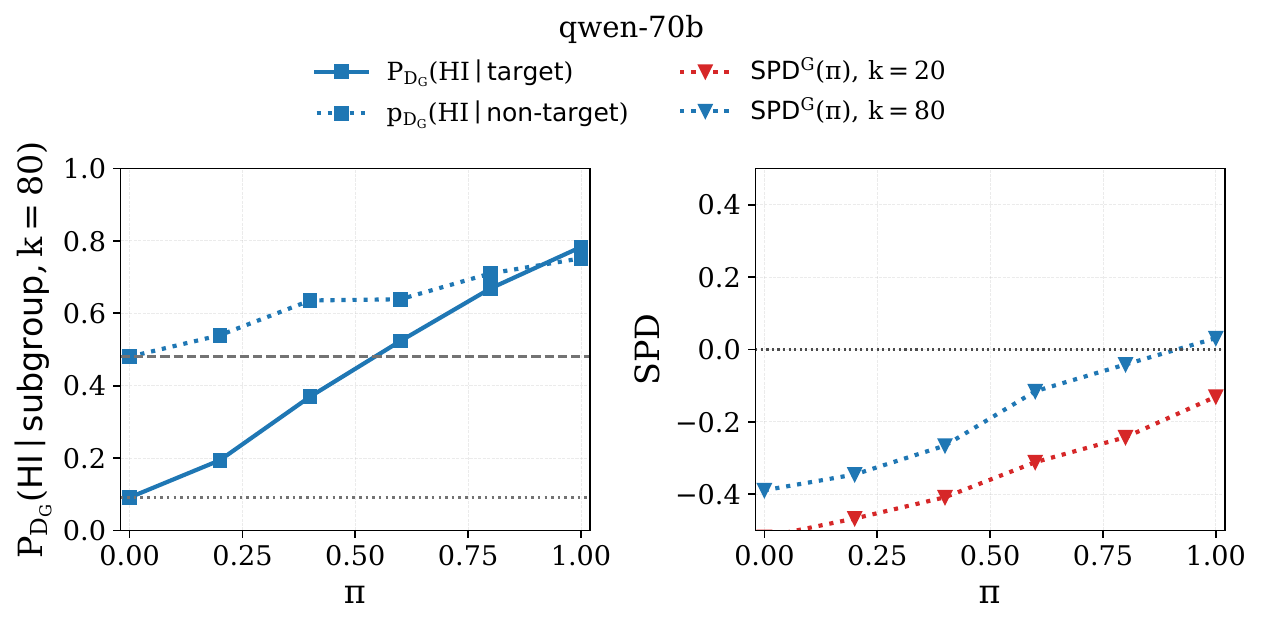}
  \vspace{-0.5em}
  \caption{\textbf{Conditional bias propagation on Adult (target: Female).} We systematically vary the in-context Income rate for the Female subgroup while fixing the non-targeted Male subgroup at $0.5$ ($k=80$). While all models correctly propagate the targeted shift, they differ in their ability to isolate this bias. Larger models within the same family (Mixtral-8x7B) demonstrate conditional propagation with minimal spillover to the other subgroup compared to smaller models (Mistral-7B).}
  \label{fig:cond_granite}
\end{figure}

\subsection{Adversarial In-Context Bias Injection}

\paragraph{Additional feature-aligned analysis}
Beyond shifting group-wise label rates, the feature-aligned attacker can also reshape the covariate profile of the targeted subgroup in the synthetic data. For Compas, the attacker anchors highly predictive recidivism features (e.g., priors\_count, juv\_misd\_count, juv\_other\_count) to plausible values that co-occur with the positive label for the protected group (Table \ref{tab:feature_alignment}). Figure~\ref{fig:feature_aligned} visualizes this effect by plotting the synthetic univariate densities of aligned predictors separately for African-American individuals and the complementary group as $\pi$ increases. As the fraction of adversarial examples grows, the protected-group distributions shift toward the attacker’s target region (e.g., mass concentrating around 5 prior counts and near-zero juvenile offenses), whereas the non-targeted group remains closer to its baseline distribution, with minor leak. This separation indicates that the attack is not merely changing marginal outcomes, it induces a subgroup-conditional covariate shift that can make the protected attribute more predictive for downstream models. 

\begin{figure*}[th!]
  \centering
  \includegraphics[width=0.49\linewidth]{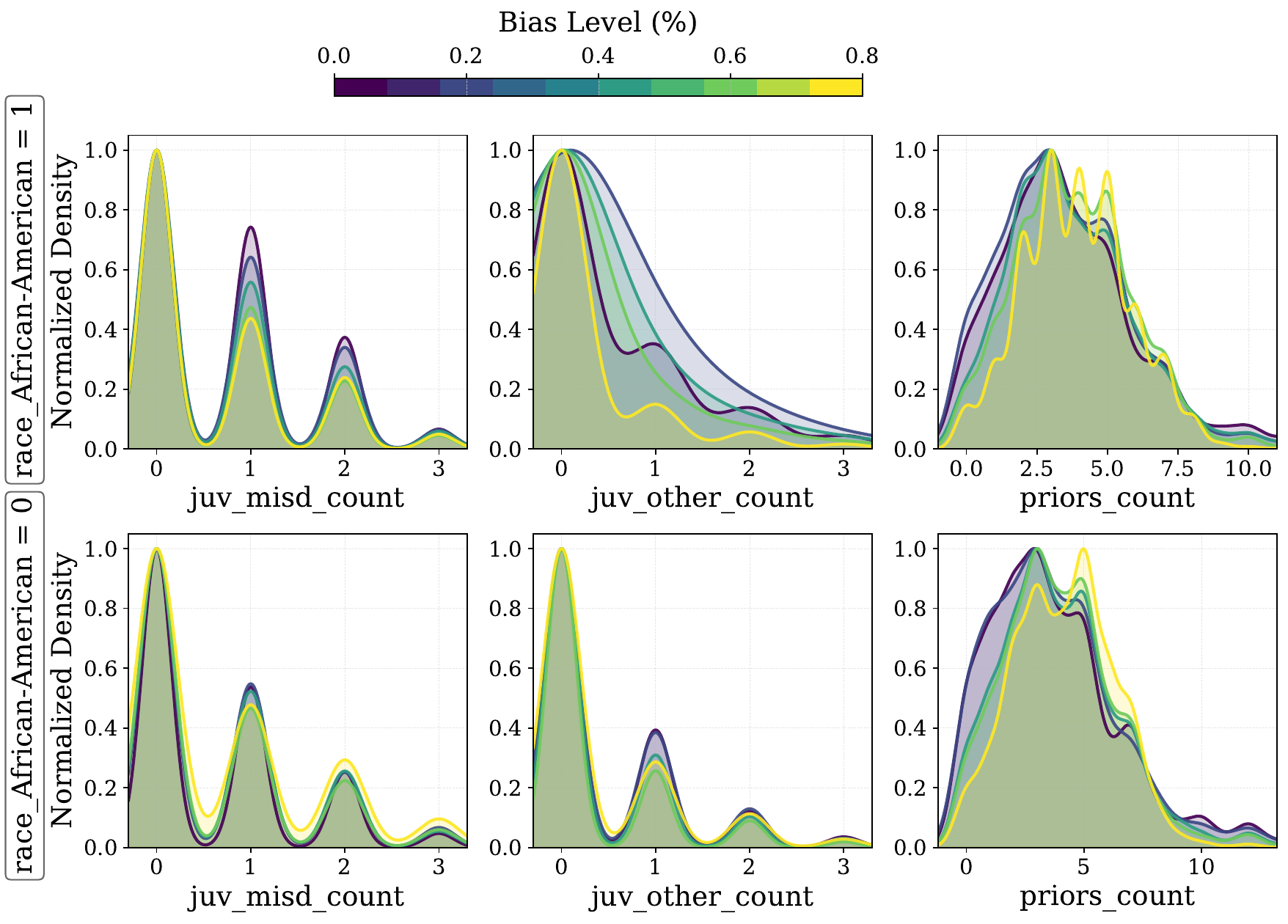}
  \includegraphics[width=0.49\linewidth]{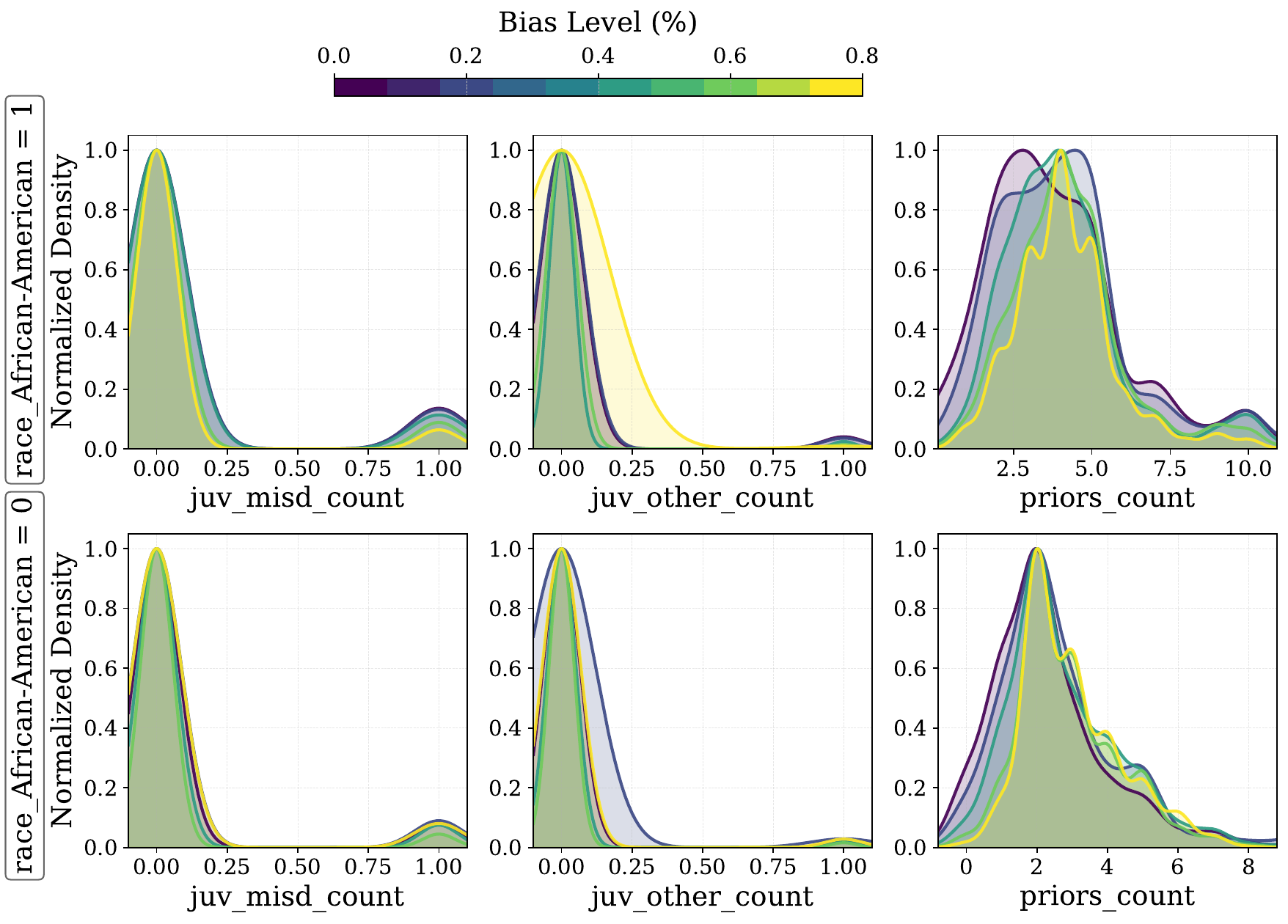}
  \caption{\textbf{Normalized density of aligned features under feature-aligned bias injection.} (\textit{Compas}, $k{=}80$) Synthetic univariate distributions of key recidivism predictors for the protected subgroup (African-American) and the complementary group as the fraction of adversarial in-context examples $\pi$ increases. Left: Mistral-7B; right: Qwen-70B. Increasing $\pi$ steers the protected-group feature profiles toward attacker-chosen plausible values (e.g., lower \texttt{priors\_count} and fewer juvenile offenses) while leaving the non-targeted group closer to baseline, illustrating the selective subgroup manipulation.}
  \label{fig:feature_aligned}
\end{figure*}

\begin{figure*}[th]
  \centering
    \includegraphics[width=0.85\textwidth]{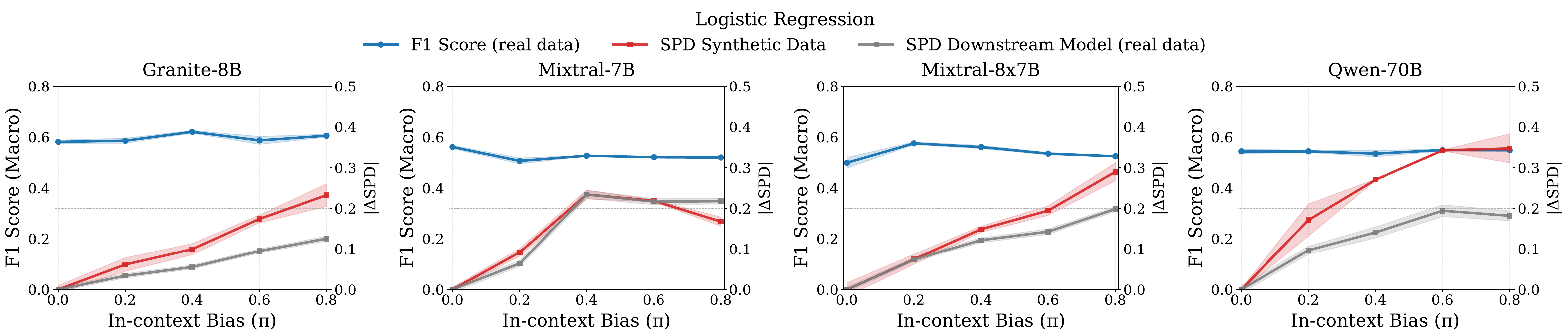}
    \includegraphics[width=0.85\textwidth]{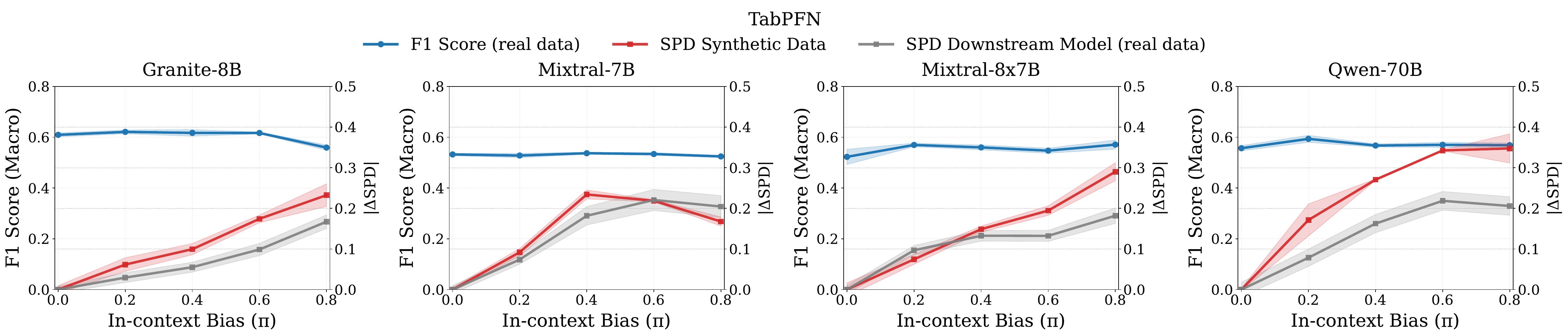}
\caption{(\textit{Compas}) Effect of the injected in-context bias rate $\pi$ on downstream utility and fairness for four LLM-based tabular generators. Top: Logistic Regression; bottom: TabPFN. Curves report macro F1 of the downstream classifier trained on synthetic data and evaluated on the real test set (blue), absolute statistical parity difference (SPD) in the synthetic data (red), and absolute SPD of the classifier's predictions on the real test set (grey). Across generators and both downstream models, macro F1 remains largely stable as $\pi$ increases, whereas both synthetic and downstream SPD increase substantially.}
\label{fig:Compas_log_attack}
\end{figure*}

\paragraph{Additional downstream classifiers}
Figure~\ref{fig:Compas_log_attack} complements the main Random-Forest results by evaluating whether adversarial in-context bias injection remains effective when the downstream learner is changed to Logistic Regression and TabPFN. Across all four LLM generators, increasing the injected bias rate $\pi$ produces a consistent rise in statistical disparity in the synthetic data (red) and in the downstream predictions evaluated on the real test set (grey), while the macro F1 on real data (blue) stays largely stable. This indicates that once the in-context examples distort the group-conditional label distribution of the synthetic data, both downstream models inherit\textendash and in some cases amplify \textendash these disparities despite similar utility. Overall, these plots reinforce the utility-fairness decoupling observed in the main experiments, where standard accuracy diagnostics alone are insufficient to detect prompt-level fairness attacks in few-shot LLM-based tabular generation.

\paragraph{Generalization to Bank Marketing}

We complement the main evaluation with the Bank Marketing dataset \cite{moro2014bank}, a real-world financial dataset containing approximately 42000 customer records and a mixture of numerical and categorical attributes. The binary prediction task is whether a customer subscribes to a term deposit. The features describe customer demographics and financial status, including age, occupation, marital status, education, account balance, credit default, housing loan, and personal loan, together with information about the current and previous marketing campaigns. We define individuals younger than 30 years as the targeted subgroup and associate this group with the positive subscription outcome in the adversarial demonstrations. 

\begin{figure}[h]
  \centering
  \includegraphics[width=0.65\linewidth]{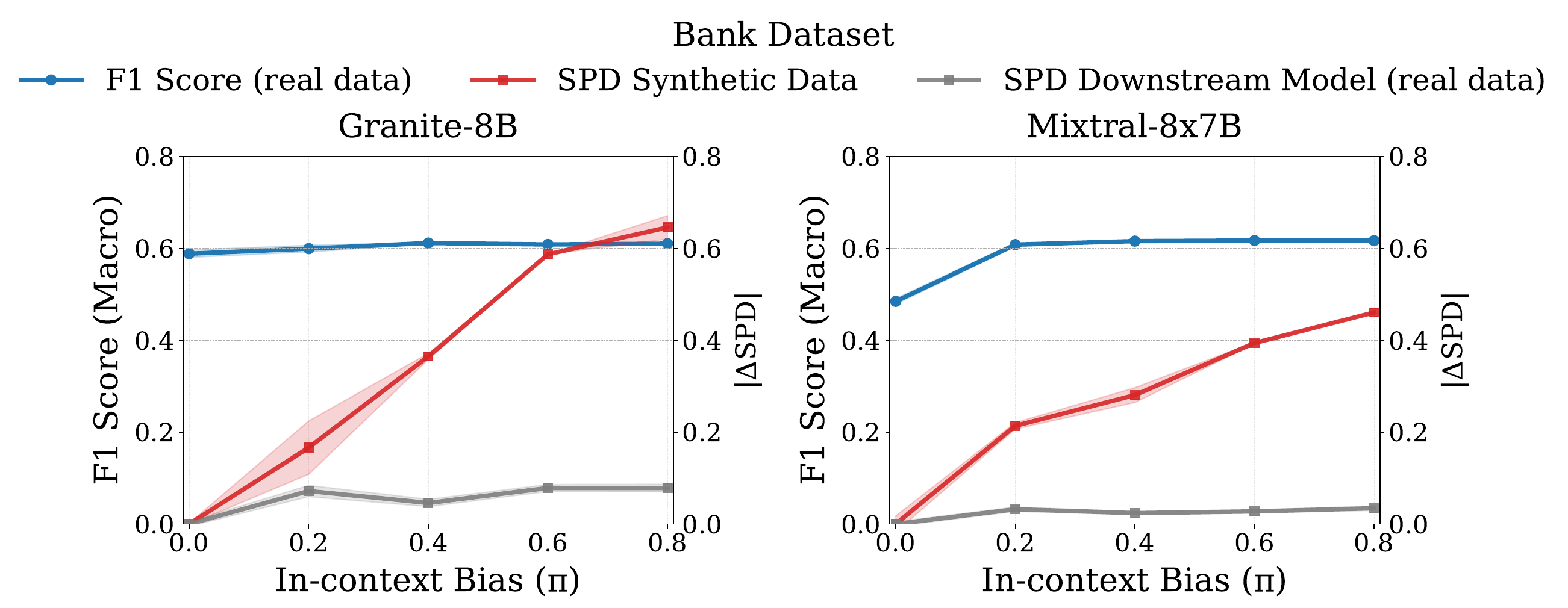}
  \vspace{-0.5em}
    \caption{\textbf{Generalization to Bank Marketing.} Effect of the injected in-context bias rate $\pi$ on downstream utility and fairness for Granite-8B and Mixtral-8x7B ($k=80$). The blue curve reports the macro-F1 of a Random Forest classifier trained on the synthetic data and evaluated on the real test set. The red and grey curves report the absolute SPD of the synthetic data and downstream predictions, respectively. Both generators exhibit a clear increase in synthetic-data disparity while maintaining broadly stable utility. The downstream effect is smaller and model-dependent, indicating that prompt-level bias reliably alters the generated distribution, while its transfer to downstream predictions depends on the generator and task.} 
  \label{fig:bank_results}
\end{figure}

Figure~\ref{fig:bank_results} reports utility and fairness as the adversarial fraction $\pi$ increases. For both Granite-8B and Mixtral-8x7B, the absolute synthetic-data disparity $|\mathrm{SPD}_S|$ increases consistently with the injected bias rate, confirming that prompt-conditioned bias propagation also occurs in this larger mixed-type dataset. At the same time, macro-F1 on the real test set remains comparatively stable, reproducing the utility--fairness decoupling observed in the primary experiments. 

The downstream effect is more attenuated and differs across generators. Granite-8B exhibits a modest increase in downstream disparity, whereas Mixtral-8x7B remains closer to its baseline despite the pronounced synthetic-data shift. These results extend the evidence for data-level bias propagation beyond the primary benchmarks, while also showing that the extent to which synthetic disparities are learned by a downstream classifier is model- and task-dependent.

\paragraph{Complementary Statistical Analysis}

Table~\ref{tab:synthetic_bias_validation} shows that biased in-context examples significantly propagate from the prompt to the generated synthetic data. Relative to each model's own $\pi=0$ baseline, nearly all bias conditions yield negative $\Delta\mathrm{SPD}_S$, indicating a systematic shift in the intended direction. The effect generally strengthens from $\pi=0.3$ to $\pi=0.6$, showing a larger prompt-induced shift at higher bias rate. Propagation is especially consistent for Diabetes, Thyroid, and Compas, and Qwen-70B exhibits significant shifts across all datasets. 

\begin{table*}[t]
\scriptsize
\setlength{\tabcolsep}{2pt}
\renewcommand{\arraystretch}{1.0}
\centering
\caption{Statistical validation of bias propagation in the generated synthetic data. $\Delta\mathrm{SPD}_S=\mathrm{SPD}_S(\pi)-\mathrm{SPD}_S(0)$ compares each biased condition against the same model and dataset at $\pi=0$. Confidence intervals are bootstrap intervals over generated rows with $B=1{,}000$ resamples, and $p_{\mathrm{perm}}$ is a two-sided permutation $p$-value.}
\label{tab:synthetic_bias_validation}
\begin{tabular}{l c r c c r c c r c c r c c}
\toprule
\textbf{Model} & $\boldsymbol{\pi}$
& \multicolumn{3}{c}{\textbf{Adult}}
& \multicolumn{3}{c}{\textbf{Compas}}
& \multicolumn{3}{c}{\textbf{Diabetes}}
& \multicolumn{3}{c}{\textbf{Thyroid}} \\
\cmidrule(lr){3-5}\cmidrule(lr){6-8}\cmidrule(lr){9-11}\cmidrule(lr){12-14}
& & $\boldsymbol{\Delta\mathrm{SPD}_S}$ & \textbf{95\% CI} & $\boldsymbol{p}$
& $\boldsymbol{\Delta\mathrm{SPD}_S}$ & \textbf{95\% CI} & $\boldsymbol{p}$
& $\boldsymbol{\Delta\mathrm{SPD}_S}$ & \textbf{95\% CI} & $\boldsymbol{p}$
& $\boldsymbol{\Delta\mathrm{SPD}_S}$ & \textbf{95\% CI} & $\boldsymbol{p}$ \\
\midrule
\multirow{2}{*}{Granite-8B}
& 0.3 & {-0.677} & [-0.770, -0.549] & 0.002 & {-0.038} & [-0.079, -0.001] & 0.060 & {-0.119} & [-0.158, -0.076] & 0.002 & {-0.278} & [-0.330, -0.221] & 0.002 \\
& 0.6 & {-0.728} & [-0.807, -0.600] & 0.002 & {-0.174} & [-0.213, -0.136] & 0.002 & {-0.244} & [-0.293, -0.202] & 0.002 & {-0.322} & [-0.374, -0.271] & 0.002 \\
\midrule
\multirow{2}{*}{Mistral-7B}
& 0.3 & {-0.276} & [-0.412, -0.117] & 0.005 & {-0.191} & [-0.226, -0.152] & 0.002 & {-0.122} & [-0.166, -0.075] & 0.002 & {-0.217} & [-0.278, -0.152] & 0.002 \\
& 0.6 & {-0.378} & [-0.491, -0.263] & 0.002 & {-0.219} & [-0.255, -0.181] & 0.002 & {-0.263} & [-0.307, -0.214] & 0.002 & {-0.303} & [-0.357, -0.250] & 0.002 \\
\midrule
\multirow{2}{*}{Mixtral-8x7B}
& 0.3 & {-0.195} & [-0.381, -0.000] & 0.017 & {-0.098} & [-0.136, -0.056] & 0.002 & {-0.130} & [-0.173, -0.085] & 0.002 & {-0.153} & [-0.207, -0.100] & 0.002 \\
& 0.6 & {-0.240} & [-0.413, -0.075] & 0.002 & {-0.195} & [-0.229, -0.158] & 0.002 & {-0.209} & [-0.256, -0.161] & 0.002 & {-0.152} & [-0.202, -0.105] & 0.002 \\
\midrule
\multirow{2}{*}{Qwen-70B}
& 0.3 & {-0.728} & [-0.841, -0.577] & 0.002 & {-0.205} & [-0.247, -0.154] & 0.002 & {-0.181} & [-0.233, -0.133] & 0.002 & {-0.168} & [-0.239, -0.097] & 0.002 \\
& 0.6 & {-0.723} & [-0.815, -0.594] & 0.002 & {-0.342} & [-0.385, -0.294] & 0.002 & {-0.491} & [-0.553, -0.434] & 0.002 & {-0.243} & [-0.298, -0.188] & 0.002 \\
\bottomrule
\end{tabular}
\end{table*}

\section{Prompt Structure} \label{app:c}

\subsection{Prompt Instruction Ablation}
\label{ablation:no_mirroring}

Our main experiments use prompts that explicitly instruct the LLM to "produce realistic yet diverse synthetic samples that mirror the causal structure and feature-label distributions of the provided examples." To test whether bias propagation depends on this explicit mirroring instruction, we conduct an ablation using a minimal prompt: "Using your knowledge of recidivism data, generate exactly two realistic samples." This prompt provides the same in-context examples and output format but offers no guidance about replicating prompt statistics.

\begin{figure*}[th]
  \centering
    \includegraphics[width=0.49\textwidth]{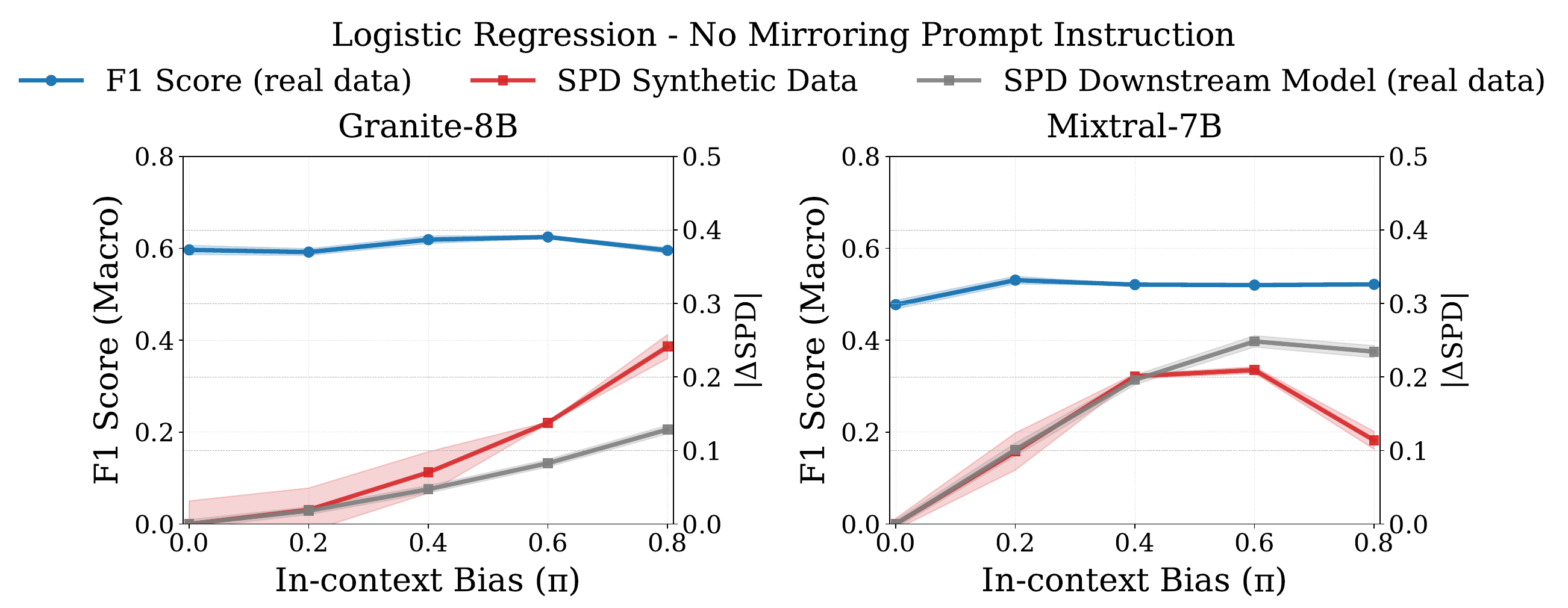}
    \includegraphics[width=0.49\textwidth]{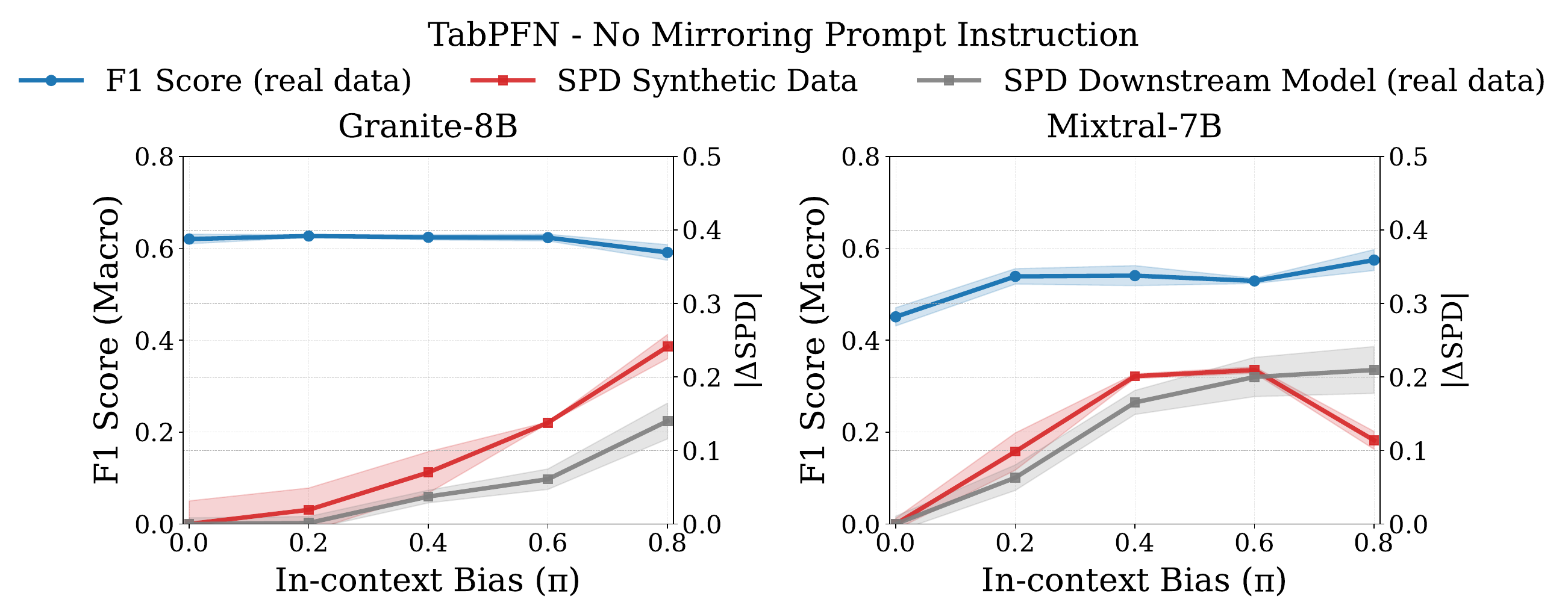}
\caption{\textbf{Bias propagation persists without explicit mirroring instructions.} (\textit{Compas, k=80}) Downstream utility and fairness on real test data for Logistic Regression (left) and TabPFN (right) classifiers trained on synthetic data. Mistral-7B and Granite-8B both show increasing SPD with bias rate $\pi$ despite stable F1, demonstrating that bias propagation occurs even when prompts contain no instruction to replicate in-context statistics.}
\label{fig:ablation_instruction}
\end{figure*}

We regenerate synthetic data using Mistral-7B and Granite-8B on Compas across $\pi$ with $k=80$. Figure~\ref{fig:ablation_instruction} shows that in-context bias propagation persists even without explicit instructions. For both models and both downstream classifiers (Logistic Regression and TabPFN), $\text{SPD}_S$ and $\text{SPD}_D$ increase substantially with $\pi$, while macro F1 remains stable.

These results show that bias propagation persists without explicit instructions to replicate in-context statistics. LLMs still transfer distributional patterns from demonstrations to generated outputs, indicating that removing the mirroring instruction alone does not prevent adversarial bias injection in the evaluated settings.

\subsection{Prompts Templates}

Here we add the prompt templates used for LLM-based tabular generation in our experiments. We include representative prompts for the Adult dataset to clarify (i) the common system role and JSON-format constraints shared across settings, and (ii) the task-specific instructions used in the marginal, conditional, and intersectional analyses. In all cases, the model receives $k$ in-context examples and is required to return a parseable JSON array with a fixed key set and key order. We provide only the changes made over the Marginal Analysis prompt for Conditional and Intersectional prompts.

\begin{tcolorbox}[
    promptbox,
    breakable,
    title={\;Unconstrained Prompt – Adult - Marginal Analysis\;},
    colframe=red!50!black,     
    colback =red!5!white       
]
\lstset{%
  basicstyle=\footnotesize\ttfamily,
  columns=fullflexible,
  keepspaces=true,
  breaklines=true}
\begin{lstlisting}
System role: 
You are a tabular synthetic data generator. You produce realistic yet diverse synthetic samples that mirror the causal structure and feature-label distributions of the provided examples.
User instructions:
    Using your knowledge of adult income data, generate exactly two realistic samples.
    You will first receive a block of real examples as in-context demonstrations:

In-context data:
{icl_examples}

Produce your output as a single JSON array containing two objects. Do not include any explanatory text, commentary, or extra fields. The JSON must be directly parseable (start with [ and end with ], no trailing commas).

Each object must have exactly these keys in this order:

{ 
"age": "int", 
"workclass": "string", 
"fnlwgt": "int", 
"education": "string",
"educational-num": "int", 
"marital-status": "string", 
"occupation": "string",
"relationship": "string", 
"race": "string",
"gender": "string", 
"capital-gain": "int", 
"capital-loss": "int", 
"hours-per-week": "int", 
"native-country": "string", 
"income": "string", 
}

Output example:
[
  {example 1},
  {example 2}
]

Strict requirement: Return only the JSON array of two objects with the exact key order and no extra keys or prose.

\end{lstlisting}
\end{tcolorbox}

\begin{tcolorbox}[
    promptbox,
    title={\;Balanced Prompt – Adult - Conditional Analysis\;},
    colframe=green!50!black,   
    colback =green!5!white     
]
\lstset{%
  basicstyle=\footnotesize\ttfamily,
  columns=fullflexible,
  keepspaces=true,
  breaklines=true}
\begin{lstlisting}

...

Output example:
[
  {example 1 - White},
  {example 2 - Black}
]

Strictly return only the JSON array of two objects corresponding to one White and one Black, with same format as the Output example - no prose, no reordered or extra keys.
\end{lstlisting}
\end{tcolorbox}

\begin{tcolorbox}[
    promptbox,
    title={\;Balanced Prompt – Adult - Intersectional Analysis\;},
    colframe=blue!50!black,     
    colback =blue!5!white       
]
\lstset{%
  basicstyle=\footnotesize\ttfamily,
  columns=fullflexible,
  keepspaces=true,
  breaklines=true}
\begin{lstlisting}
...

User instructions:
    Using your knowledge of adult income data, generate exactly four realistic samples-one for each combination of "gender" in {"Female", "Male"} and "race" in {"Black", "White"}:
    - ("gender": "Female", "race": "Black")
    - ("gender": "Female", "race": "White")
    - ("gender": "Male", "race": "Black")
    - ("gender": "Male", "race": "White")
    You will first receive a block of real examples as in-context demonstrations:

...

Output example:
[
  {example 1 - Female - >50K},
  {example 2 - Female - <=50K},
  {example 3 - Male - >50K},
  {example 4 - Male - <=50K}
]
Strictly return only the JSON array of four objects corresponding to the four (gender, race) combinations above, with the same format as the Output example - no prose, no reordered or extra keys.
\end{lstlisting}
\end{tcolorbox}

\begin{tcolorbox}[
    promptbox,
    title={\;No Mirroring Instruction – Compas\;},
    colframe=yellow!50!black,     
    colback =yellow!5!white       
]
\lstset{%
  basicstyle=\footnotesize\ttfamily,
  columns=fullflexible,
  keepspaces=true,
  breaklines=true}
\begin{lstlisting}
System role: 
You are a tabular synthetic data generator. You produce realistic yet diverse synthetic samples.

User instructions:
    
    Using your knowledge of recidivism data, generate exactly two realistic samples.
    You will first receive a block of real examples as in-context demonstrations. If this block is empty or missing, proceed anyway using general recidivism priors.

In-context data:
{icl_examples}
...
\end{lstlisting}
\end{tcolorbox}

\begin{tcolorbox}[
    promptbox,
    title={\;Instruction-based Mitigation – Compas\;},
    colframe=orange!50!black,     
    colback =orange!5!white       
]
\lstset{%
  basicstyle=\footnotesize\ttfamily,
  columns=fullflexible,
  keepspaces=true,
  breaklines=true}
\begin{lstlisting}
System role: 
You are a tabular synthetic data generator. You produce realistic yet diverse synthetic samples.

User instructions:
    
    Using your knowledge of recidivism data, generate exactly two realistic samples.
    You will first receive a block of real examples as in-context demonstrations. If this block is empty or missing, proceed anyway using general recidivism priors.

In-context data:
{icl_examples}

...

Assign positive label to all races equally often.

...
\end{lstlisting}
\end{tcolorbox}

\end{document}